\documentclass{article}

\usepackage[preprint,nonatbib]{neurips_2022}

\usepackage[utf8]{inputenc} 
\usepackage[T1]{fontenc}    
\usepackage{hyperref}       
\usepackage{url}            
\usepackage{booktabs}       
\usepackage{amsfonts}       
\usepackage{nicefrac}       
\usepackage{microtype}      
\usepackage{xcolor}         
\usepackage{amsmath,amssymb,mathtools,bm}
\usepackage{siunitx}
\usepackage{subcaption}
\usepackage[export]{adjustbox}
\usepackage{makecell}
\usepackage{arydshln}
\usepackage{multirow}
\usepackage{cleveref}
\usepackage{placeins}

\DeclareMathOperator*{\argmin}{argmin}

\title{Learning rich optical embeddings for privacy-preserving lensless image classification}

\author{%
	Eric Bezzam \\
  Audiovisual Communications Laboratory\\
  \'{E}cole Polytechnique F\'{e}d\'{e}rale de Lausanne\\
  \texttt{eric.bezzam@epfl.ch} \\
  \And
  Martin Vetterli\\
  Audiovisual Communications Laboratory\\
  \'{E}cole Polytechnique F\'{e}d\'{e}rale de Lausanne\\
  \texttt{martin.vetterli@epfl.ch} \\
  \And
  Matthieu Simeoni\\
  Center for Imaging\\
  \'{E}cole Polytechnique F\'{e}d\'{e}rale de Lausanne\\
  \texttt{matthieu.simeoni@epfl.ch} \\
}

\begin{document}

\maketitle

\begin{abstract}
By replacing the lens with a thin optical element, lensless imaging enables new applications and solutions beyond those supported by traditional camera design and post-processing, e.g.\ compact and lightweight form factors and visual privacy. The latter arises from the highly multiplexed measurements of lensless cameras, which require knowledge of the imaging system to recover a recognizable image. In this work, we exploit this unique multiplexing property: casting the optics as an encoder that produces learned embeddings directly at the camera sensor. We do so in the context of image classification, where we jointly optimize the encoder's parameters and those of an image classifier in an end-to-end fashion. Our experiments show that jointly learning the lensless optical encoder and the digital processing allows for lower resolution embeddings at the sensor, and hence better privacy as it is much harder to recover meaningful images from these measurements. Additional experiments show that such an optimization allows for lensless measurements that are more robust to typical real-world image transformations. While this work focuses on classification, the proposed programmable lensless camera and end-to-end optimization can be applied to other computational imaging tasks.
\end{abstract}

\section{Introduction}

Advances in imaging hardware, fabrication techniques, and computational methods have enabled novel camera design strategies that go beyond mimicking the human eye. \textit{Lensless imaging} is one of those approaches, 
replacing a lens (and necessary focusing distances) with a thinner and potentially inexpensive optical element and a computational image formation step~\cite{boominathan2022recent}.
A variety of applications in virtual/augmented reality, wearables, and robotics can benefit from the low-cost and compact form factor that the lensless imaging paradigm has to offer.

\begin{figure}[t!]
	\centering
	\begin{subfigure}{.24\textwidth}
		\centering
		\includegraphics[width=0.99\linewidth]{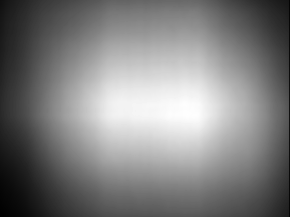}
		\caption{}
		\label{fig:celeba_0_diffuser_raw}
		
	\end{subfigure}
	\begin{subfigure}{.24\textwidth}
		\centering
		\includegraphics[width=0.99\linewidth]{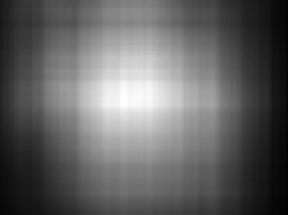} 
		\caption{}
		\label{fig:celeba_mls_raw_measurement}
	\end{subfigure}
	\begin{subfigure}{.24\textwidth}
		\centering
		\includegraphics[width=0.99\linewidth]{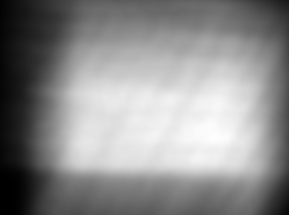} 
		\caption{}
		\label{fig:mnist_test_0_diffuser_raw}
	\end{subfigure}
	\begin{subfigure}{.24\textwidth}
		\centering
		\includegraphics[width=0.99\linewidth]{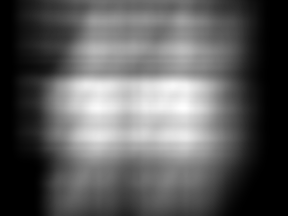} 
		\caption{}
		\label{fig:mls_digit_raw_measurement}
	\end{subfigure}
	\begin{subfigure}{.24\textwidth}
		\centering
		\includegraphics[width=0.99\linewidth]{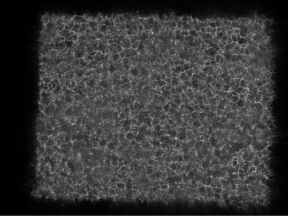}
		\caption{}
		\label{fig:diffuser_psf_down4}
	\end{subfigure}
	\begin{subfigure}{.24\textwidth}
		\centering
		\includegraphics[width=0.99\linewidth]{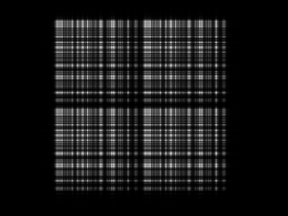} 
		\caption{}
		\label{fig:mls4mm_psf_diffracted}
	\end{subfigure}
	\begin{subfigure}{.24\textwidth}
		\centering
		\includegraphics[width=0.99\linewidth]{figs/diffuser_psf_down4.png}
		\caption{}
	\end{subfigure}
	\begin{subfigure}{.24\textwidth}
		\centering
		\includegraphics[width=0.99\linewidth]{figs/mls4mm_psf_diffracted.png} 
		\caption{}
	\end{subfigure}
	\\
	\begin{subfigure}{.24\textwidth}
		\centering
		\includegraphics[width=0.99\linewidth]{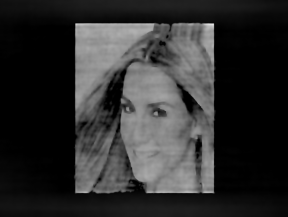}
		\caption{}
		\label{fig:celeba_0_diffuser_admm100}
	\end{subfigure}
	\begin{subfigure}{.24\textwidth}
		\centering
		\includegraphics[width=0.99\linewidth]{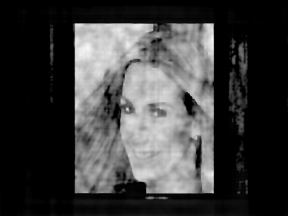} 
		\caption{}
		\label{fig:celeba_mls_down4_recon}
	\end{subfigure}
	\begin{subfigure}{.24\textwidth}
		\centering
		\includegraphics[width=0.99\linewidth]{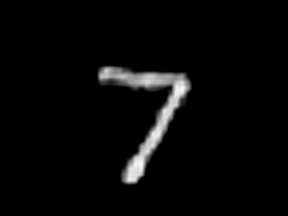} 
		\caption{}
		\label{fig:mnist_test_0_diffuser_admm50}
	\end{subfigure}
	\begin{subfigure}{.24\textwidth}
		\centering
		\includegraphics[width=0.99\linewidth]{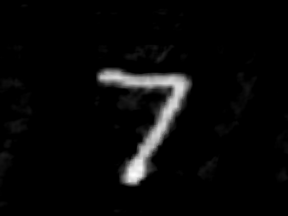} 
		\caption{}
		\label{fig:mls_admm_reconstruction}
	\end{subfigure}
	\\
	\begin{subfigure}{.24\textwidth}
		\centering
		\includegraphics[width=0.99\linewidth]{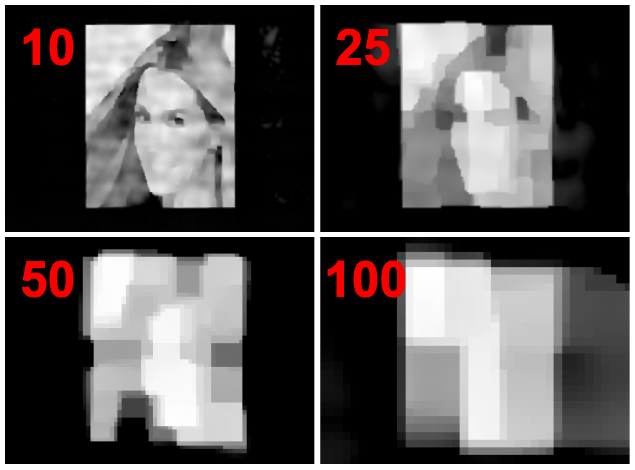}
		\caption{}
		\label{fig:celeba_diffuser_down_recon}
	\end{subfigure}
	\begin{subfigure}{.24\textwidth}
		\centering
		\includegraphics[width=0.99\linewidth]{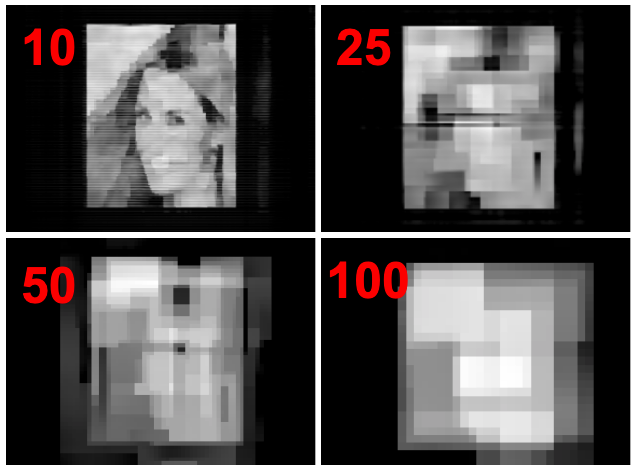} 
		
		\caption{}
		\label{fig:celeba_mls_down_recon}
	\end{subfigure}
	\begin{subfigure}{.24\textwidth}
		\centering
		\includegraphics[width=0.99\linewidth]{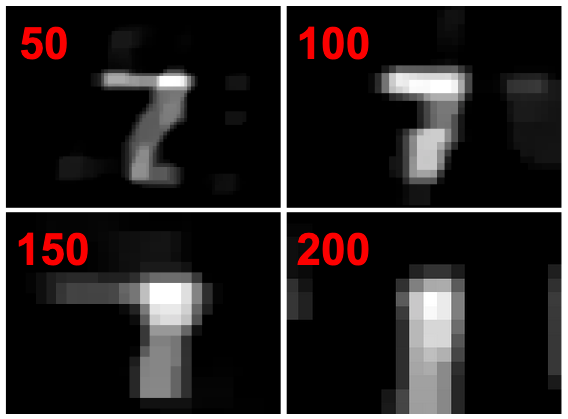} 
		\caption{}
		\label{fig:recon_mnist0_down}
	\end{subfigure}
	\begin{subfigure}{.24\textwidth}
		\centering
		\includegraphics[width=0.99\linewidth]{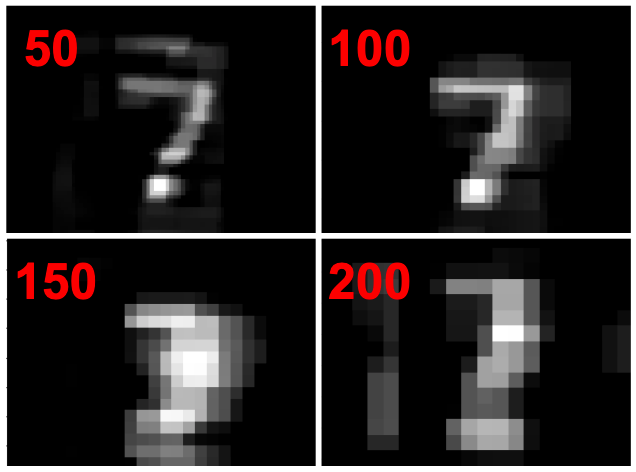} 
		\caption{}
		\label{fig:mnist_mls_down}
	\end{subfigure}
	\caption{Discerning content from lensless camera raw measurements (top row) is next to impossible, motivating privacy-preserving imaging with such cameras.
		However, with sufficient knowledge about the camera (e.g.\ a point spread function, second row) and an appropriate computational algorithm, one is able to recover an estimate of the underlying object (third row, using ADMM~\cite{admm} and a total variation prior).
		If the raw sensor measurement is under-sampled, it becomes increasingly difficult for classical recovery algorithms to recover an meaningful estimate of the underlying object. (Bottom row) PSF and raw measurements are downsampled by the factor in the top left corner, simulating a sensor of lower resolution prior to reconstruction.
	}
	\label{fig:raw2reconstruction}
\end{figure}

The optical element in such systems is typically a passive or programmable mask placed at a short distance from the sensor. The resulting measurements are highly multiplexed, as seen in \Cref{fig:celeba_0_diffuser_raw,fig:celeba_mls_raw_measurement,fig:mnist_test_0_diffuser_raw,fig:mls_digit_raw_measurement}, due to a system response, i.e.\ \emph{point spread function} (PSF), of large support unlike that of a lens. \Cref{fig:diffuser_psf_down4,fig:mls4mm_psf_diffracted} show the PSFs of typical lensless encoders, namely a caustic pattern of a height-varying phase mask~\cite{Antipa:18,phlatcam} and a diffracted coded aperture (CA) mask~\cite{Chi:11,10.1117/1.OE.54.2.023102,flatcam}.

The majority of contributions in lensless imaging have focused on improving the computational methods to go from raw measurements to demultiplexed images, e.g.\ from \Cref{fig:celeba_0_diffuser_raw,fig:celeba_mls_raw_measurement,fig:mnist_test_0_diffuser_raw,fig:mls_digit_raw_measurement} to \Cref{fig:celeba_0_diffuser_admm100,fig:celeba_mls_down4_recon,fig:mnist_test_0_diffuser_admm50,fig:mls_admm_reconstruction}. Data-driven techniques and deep learning have had an influential role in this progress, yielding faster reconstruction times and improved reconstruction quality~\cite{Khan_2019_ICCV,Monakhova:19,Pan:22} with respect to classical techniques based on  system inversion~\cite{10.1117/1.OE.54.2.023102,flatcam,sweepcam2020} and convex optimization~\cite{huang2013,Antipa:18,phlatcam}. 

While machine learning advances have been readily incorporated in lensless imaging reconstruction and classification tasks~\cite{8590781,Pan:21}, the design of the optical element itself remains rather heuristic-based. Criteria such as sparsity, a large number of directional filters, high contrast, a delta-like autocorrelation, or designs to simplify the computational recovery~\cite{Chi:11,10.1117/1.OE.54.2.023102,flatcam,Antipa:18,phlatcam} have been used to tackle the task of PSF engineering, independent of the down-stream task. The potential to \textit{jointly} optimize the optical encoding and a digital post-processing has been successfully demonstrated in other computational tasks, albeit with lenses, for extended depth-of-field~\cite{sitzmann2018,pinilla2022}, super-resolution~\cite{sitzmann2018}, classification~\cite{chang2018hybrid}, 3-D imaging~\cite{markley2021physicsbased,deb2022programmable}, and hyperspectral imaging~\cite{Vargas_2021_ICCV,9157577}. 

In this paper, we apply end-to-end optimization to a 
lensless camera to jointly learn (1) a programmable mask pattern prior to the sensor measurement \textit{and} (2) the subsequent digital processing. As well as exploiting edge components for compute and the compactness of lensless cameras, privacy-preserving classification is one of the key motivations for this approach. The multiplexed measurements of lensless cameras have been touted to maintain visual privacy~\cite{boominathan2022recent,9021989,DBLP:journals/corr/abs-2106-14577,shi2022loen} as they contain hardly any perceivable features. 
However, a malicious user with access to the camera can still recover an image of the underlying object through a couple measurements and clever post-processing.
The objective of this work is to jointly optimize the optical encoding and the digital classifier in order to significantly reduce the size of the sensor ``embedding'' by exploiting this multiplexing characteristic. As the sensor resolution decreases, it becomes increasingly difficult for lensless imaging reconstruction techniques to recover a meaningful image, as demonstrated in \Cref{fig:celeba_diffuser_down_recon,fig:celeba_mls_down_recon,fig:recon_mnist0_down,fig:mnist_mls_down}. 
Jointly optimizing this one-to-many mapping for a particular task, e.g.\ classification, has the potential to produce richer embeddings, much like digital encoders~\cite{doi:10.1126/science.1127647}, with a lower resolution sensor, all the while maintaining performance on the task at hand and enhancing visual privacy.

\paragraph{Contributions} In this work, we exploit multiplexing properties of lensless cameras in order to learn privacy-preserving embeddings by training the imaging system end-to-end. Concretely, we determine the optimal pattern for a programmable component prior to the sensor, i.e.\ an amplitude spatial light modulator (SLM), in order to perform image classification. 

To the best of our knowledge, one recent work has applied end-to-end optimization for lensless imaging with passive masks~\cite{shi2022loen}; however none with programmable components. Using such components can help reduce model mismatch through hardware-in-the-loop (HITL)~\cite{Peng:2020:NeuralHolography} or equivalently, physics-aware training~\cite{wright2022deep}. Moreover, the re-programmability of an SLM means the end-to-end optimized camera does not have to be relegated to a single application or setting. It can be updated after deployment or conveniently reconfigured for a different task or in the case of a malicious user.

Our experiments on handwritten digit classification demonstrate the potential of significantly reducing the embedding at the sensor, as our end-to-end approach consistently performs better than lensless cameras with a fixed encoder. Moreover, we show that jointly learning the SLM pattern with the classification task is more robust to typical image transformations: shifting, rescaling, rotating, perspective changes. We are unaware of any other work that has studied the consequences of such effects on lensless imaging.

Our end-to-end approach is based upon an imaging system that can be put together  from cheap and accessible components, totaling at around $100$~USD. As an SLM, we use a low-cost liquid crystal display (LCD), as in~\cite{zomet2006,huang2013}, which costs about $20$~USD.
To the best of our knowledge, we are the first to employ such a device in an end-to-end optimization for computational optics, as opposed to commercial SLMs which cost a few thousand USD. Our differentiable digital twin of the imaging system models incoherent, polychromatic propagation with the selected LCD component, using the bandlimited angular spectrum method (BLAS)~\cite{Matsushima:09} to account for diffraction.

In order to foster reproducibility, we open source the following under the GNU General Public License v3.0: wave propagation simulation\footnote{\url{https://github.com/ebezzam/waveprop}} and training software\footnote{\url{https://github.com/ebezzam/LenslessClassification}} Moreover, we have previously released a package to interface with the baseline and proposed cameras.\footnote{\url{https://github.com/LCAV/LenslessPiCam}}

\section{Problem statement}
\label{sec:problem}

\begin{figure}[t!]
	\centering
	\includegraphics[width=0.99\linewidth]{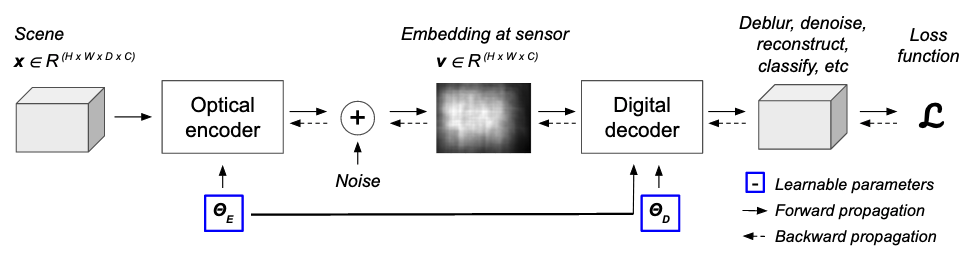} 
	\caption{Encoder-decoder perspective of cameras for end-to-end optimization. The scene could be four-dimensional (height, width, depth, and color) whereas the embedding measured at the sensor is at most three-dimensional (height, width, and color).}
	\label{fig:enc_dec}
\end{figure}

End-to-end approaches for optimizing optical components, also known as \emph{deep optics}~\cite{wetzstein2020inference} is a recent trend enabled by improved fabrication techniques and the continual development of more powerful and efficient hardware and libraries for machine learning. It is motivated by faster and cheaper inference for edge computing (taking advantage of the speed of light) and a desire to co-design the optics and the computational algorithm to obtain optimal performance for a particular application.

An encoder-decoder perspective is often used to frame such end-to-end approaches, casting the optics as the encoder and the subsequent computational algorithm as the decoder, as shown in \Cref{fig:enc_dec}, and can be formulated as the following optimization problem minimized for a labeled dataset $ \{\bm{x}_i, \bm{y}_i\}_{i = 1}^N $:
\begin{equation}
\hat{\bm{\theta}}_E, \hat{\bm{\theta}}_D = \argmin_{\bm{\theta}_E, \bm{\theta}_D} \sum_{i=1}^{N} \mathcal{L} \Big(\bm{y}_i, \underbrace{ D_{\bm{\theta}_E,\bm{\theta}_D} \big( \overbrace{O_{\bm{\theta}_E}  ( \bm{x}_i)}^{\text{embedding } \bm{v}_i} \big)}_{\text{decoder output }\bm{\hat{y}}_i}  \Big). \label{eq:optimization}
\end{equation}

$ O_{\bm{\theta}_E}(\cdot)  $ is the optical encoder, including additive noise, that outputs the sensor embedding  $ \bm{v}_i $ of an input $ \bm{x}_i $. The encoder encapsulates propagation in free space and through all optical components prior to the sensor. While this component can be simulated via a digital twin, the hardware itself can be used to produce physical realizations of $ \bm{v}_i $. Moreover, if the encoder parameters $\bm{\theta}_E$ of the physical system can be modified, 
the device itself can be used for forward propagation, and a differentiable digital model for backpropagating the error between the ground truth $ \bm{y}_i$ and the decoder output $ \bm{\hat{y}}_i $ that arose from $ \bm{v}_i $ that came directly from the device. This is the essence of HITL / physics-aware training. In some cases, the hardware can also be used for backpropagation~\cite{Zhou2020}.  

$ D_{\bm{\theta}_E,\bm{\theta}_D}(\cdot) $ is the digital decoder, which can perform a whole slew of tasks: deblurring, denoising, image reconstruction, classification, etc. It has its own set of parameters $ \bm{\theta}_D $ and can optionally make use of the optical encoder parameters, e.g.\ for physics-based learning~\cite{markley2021physicsbased}. Its output is fed to a loss function $ \mathcal{L}(\cdot) $ along with the ground-truth output $  \bm{y}_i $. 

In \Cref{sec:proposed} we present the hardware for our proposed lensless imaging system and how we model the digital twin for our optical encoder. In \Cref{sec:experiments}, as we explain our task, we present the architecture of our digital decoder, the loss function, and the labeled data $ \{\bm{x}_i, \bm{y}_i\}_{i = 1}^N $ for our experimental setup.

\section{Proposed solution for lensless classification}
\label{sec:proposed}

Our proposed camera design is motivated by the benefits of lensless cameras (compact, low-cost, privacy-preserving) and programmability.

To this end, a transmissive SLM serves as the only optical component in our encoder, specifically an off-the-shelf LCD driven by the ST7735R device which can be purchased for $\$20$.\footnote{\url{https://www.adafruit.com/product/358}} It can be wired to a Raspberry Pi ($\$35$) with the Raspberry Pi High Quality $12.3$ MP Camera ($\$50$) as a sensor, totaling our design to just $\$105$.

\begin{figure}[t!]
	\centering
	\includegraphics[width=0.7\linewidth]{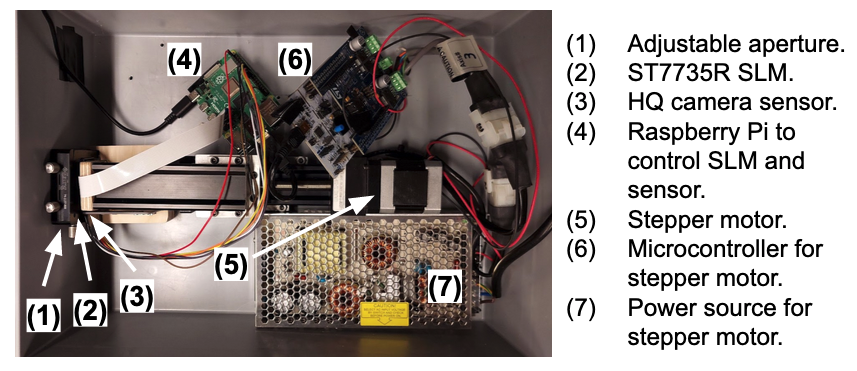} 
	\caption{Experimental prototype of programmable, amplitude SLM-based camera.}
	\label{fig:prototype_labeled}
\end{figure}

An experimental prototype of the proposed design with the aforementioned components can be seen in \Cref{fig:prototype_labeled}. The prototype includes an adjustable aperture and a stepper motor for programmatically setting the distance between the SLM and the sensor, both of which can be removed to produce a more compact design, similar to \Cref{fig:diffuser} of a lensless camera with a fixed diffuser.

\paragraph{Digital twin of optical encoder}

\label{sec:model}

End-to-end optimization requires a sufficiently accurate and differentiable simulation of the physical setup. 
Our digital twin of the imaging system shown in Figure~\ref{fig:prototype_labeled} accounts for wave-based image formation for spatially incoherent, polychromatic illumination, as is typical of natural scenes. A simulation based on wave-optics is necessary to account for diffraction due to the small SLM features and for wavelength-dependent propagation.

We adopt a common assumption from Fourier optics, namely that image formation is a linear shift-invariant (LSI) system between two parallel planes for a given wavelength~\cite{Goodman2005}. This implies that, there exists an impulse response, i.e.\ a PSF, that can be convolved with the \emph{scaled} scene in order to obtain its image at a given distance and for a specific wavelength. This convolution relationship is described in \Cref{sec:model_prop}.

Therefore, our digital twin modeling amounts to obtaining a PSF that encapsulates propagation from a given plane in the scene to the sensor plane. There are two ways to obtain this PSF: measuring it with a physical setup or simulating it.
For end-to-end approaches, a differentiable simulator is typically necessary in order to backpropagate the error to update the optical encoder parameters. In \Cref{sec:psf_modeling} we describe our modeling of this PSF for an SLM placed at a short distance in front of the sensor, as is the case for our imaging device.
The learnable parameters $\bm{\theta}_E$ of our optical encoder with the ST7735R component include: the SLM pixel amplitude values $ \{w_k\}_{k=1}^{K} $ and the distance between the SLM and the image plane $ d_2 $.
While both can be optimized in an end-to-end fashion, in this work we concentrate on optimizing  $ \{w_k\}_{k=1}^{K} $ jointly with the digital decoder parameters $\bm{\theta}_D$.

\section{Experiments}
\label{sec:experiments}

In this section, we apply our proposed camera and end-to-end optimization to handwritten digit classification (MNIST)~\cite{lecun1998mnist}.
\Cref{eq:optimization} can be slightly modified to
\begin{equation}
\label{eq:mnist}
\hat{\bm{\theta}}_E, \hat{\bm{\theta}}_D = \argmin_{\bm{\theta}_D, \bm{\theta}_E} \sum_{i=1}^{N} \mathcal{L} \Big( y_i, \underbrace{D_{\bm{\theta}_D} \overbrace{\big( O_{\bm{\theta}_E}  ( \bm{x}_i)}^{\text{embedding } \bm{v}_i} \big)}_{\text{decoder output }\bm{\hat{p}}_i}  \Big),
\end{equation}
as our decoder $ D_{\bm{\theta}_D}(\cdot) $ does not require information from the encoder in order to classify digits. The original $ \{\bm{x}_i\}_{i = 1}^N $ coming from MNIST are $ (28\times 28) $ images of handwritten digits and $ \{y_i\}_{i = 1}^N $ are labels from $ 0 $ to $ 9 $. $ \{\bm{x}_i\}_{i = 1}^N $ are simulated and resized to the dimensions of the PSF, as described in \Cref{sec:simulation}, and the decoder outputs $ \{\bm{\hat{p}}_i\}_{i = 1}^N $ are length-$ 10 $ vectors  of scores for each label.

We conduct two experiments for evaluating the effectiveness of jointly optimizing the optical encoder and the classification task:
\begin{enumerate}
	\item \Cref{sec:vary_dimension}: reduce the dimension of the embedding $ \bm{v}_i $ at the sensor and study its impact on classification performance. A lower resolution embedding at the sensor corresponds to enhanced visual privacy, as demonstrated in the introduction with \Cref{fig:celeba_diffuser_down_recon,fig:celeba_mls_down_recon,fig:recon_mnist0_down,fig:mnist_mls_down}.
	\item \Cref{sec:robustness}: apply common real-world image transformations (shifting, rescaling, rotating, perspective changes) to evaluate the robustness of the proposed camera and the end-to-end optimization to such deformations.
\end{enumerate}

$ D_{\bm{\theta}_D}(\cdot) $ takes on one of two architectures in our experiments: multi-class logistic regression or a two-layer fully-connected neural network (FCNN), which are detailed in \Cref{sec:logistic} and \Cref{sec:nn} respectively. In training both architectures, we use a cross entropy loss between the ground truth labels and the outputs of the decoder, and train for $ 50 $ epochs with a batch size of $ N = 200 $ and the Adam optimizer~\cite{adam}. More information on the training hyperparameters and compute hardware can be found in \Cref{sec:training_param}.

We use the provided train-test split of MNIST: $60'000$ training and $10'000$ test examples. Each example is simulated as per the approach described in \Cref{sec:simulation}  for an object-to-camera distance of $ \SI{40}{\centi\meter} $, a signal-to-noise ratio of $ \SI{40}{\decibel} $, and an object height of \SI{12}{\centi\meter} (unless specified otherwise). We compare six imaging systems in our experiments, and for each camera, a PSF is needed to perform this simulation. Below is a brief description of each camera and how we obtain its PSF for an object-to-camera distance of $ \SI{40}{\centi\meter} $:

\begin{itemize}
	\item \textit{Lens}: measured PSF for the camera shown in \Cref{fig:lensed_camera} with the lens focused at \SI{40}{\centi\meter}.
	\item \textit{CA} (coded aperture): a binary mask is generated by taking the outer product of a maximum length sequence (MLS), as is done in~\cite{flatcam}. A simulation of its diffraction pattern is used as the PSF.
	\item \textit{Diffuser}: measured PSF for the camera shown in \Cref{fig:diffuser}, where the diffuser is placed roughly \SI{4}{\milli\meter} from the sensor. The diffuser is double-sided tape as in the DiffuserCam tutorial~\cite{diffusercam_tut}. In~\cite{lenslesspicam}, the authors demonstrate the effectiveness of this simple diffuser for imaging when used with the Raspberry Pi High Quality Camera.
	\item \textit{Fixed SLM (m)}: measured PSF for the proposed camera shown in \Cref{fig:prototype_labeled} for a randomly programmed pattern. The mask-to-sensor distance is programmatically set to \SI{4}{\milli\meter} via the stepper motor to match the distance of the diffuser-based camera.
	\item \textit{Fixed SLM (s)}: simulated PSF for the proposed camera, using the approach described in \Cref{sec:psf_modeling} for a random set of SLM amplitude values and a mask-to-sensor distance of \SI{4}{\milli\meter}.
	\item \textit{Learned SLM}: simulated PSF for the proposed camera that is obtained by optimizing \Cref{eq:mnist} for the SLM weights and then simulating the corresponding PSF using  the approach described in \Cref{sec:psf_modeling} for a mask-to-sensor distance of \SI{4}{\milli\meter}. During training, the PSF changes at each batch as the SLM values are updated after backpropagation. 
\end{itemize}

More details such as the components for the measured PSFs and simulation details can be found in~\Cref{sec:baseline}.
For the fixed optical encoders, the embeddings $ \{ \bm{v}_i \}_{i=1}^{N} $ can be pre-computed with the approach described in \Cref{sec:simulation}. The resulting augmented dataset is normalized (according to the augmented training set statistics) prior to optimizing the classifier $ D_{\bm{\theta}_D}(\cdot) $. For \textit{Learned SLM}, we apply batch normalization~\cite{10.5555/3045118.3045167} and a ReLu activation to the sensor embedding prior to passing it to the classifier.
At inference, the parameters of batch normalization are fixed.

\subsection{Varying embedding dimension}
\label{sec:vary_dimension}

\Cref{tab:mnist_vary_embedding} reports the best test accuracy for each optical encoder, for a varying sensor embedding dimension and for two digital classification architectures: logistic regression and two-layer FCNN. The test accuracy curves can be found in \Cref{sec:test_acc_dim}.

While all approaches decrease in performance as the embedding dimension reduces, \textit{Learned SLM} is the most resilient as quantified by \Cref{tab:performance_drop}. 
The performance gap between \textit{Learned SLM} and fixed lensless encoders, as shown in \Cref{tab:mnist_vary_embedding}, decreases when a two layer FCNN is used. However, the benefits of learning this multiplexing are still evident for a very low embedding dimension of $ (3 \times 4)$.  

\begin{table}[b!]
	\caption{MNIST accuracy on test set, simulated accordingly.}
	\label{tab:mnist_vary_embedding}
	\centering
	\begin{tabular}{lcccc  cccc}
		\toprule
		\textit{Classifier} $\rightarrow $  & \multicolumn{4}{c}{Logistic regression} & \multicolumn{4}{c}{Single hidden layer, 800 units} \\
		\cmidrule(r){2-5} \cmidrule(r){6-9}
		\textit{Embedding}   $\rightarrow $  &  24$\times$32   &  12$\times$16 &  6$\times$8 &  3$\times$4 & 24$\times$32   &  12$\times$16 &  6$\times$8 &  3$\times$4\\
		\textit{Encoder}  $\downarrow $     &  =768    & =192 & =48 & =12 & =768    & =192 & =48  & =12\\
		\midrule
		Lens   & $92.3\%$ &  $74.8\%$ &  $42.8\%$  &  \multicolumn{1}{c|}{$18.4\%$}  & $97.7 \%$  & $83.0 \%$ & $41.8 \%$ & $18.8 \%$\\
		CA    &  $74.1 \%$ & $74.2 \%$  & $64.4 \%$ &   \multicolumn{1}{c|}{$59.1 \%$}  &  $97.3 \%$ & $96.3 \%$ & $91.0 \%$ & $69.9 \%$ \\
		Diffuser    & $91.0\%$  &  $81.6\%$ &  $72.6\%$ &   \multicolumn{1}{c|}{$48.5\%$}  &  $95.8 \% $&  $95.7 \%$ &  $92.8 \%$ &  $77.2\%$\\
		Fixed SLM (m)    &  $92.7\%$  & $91.5\%$ & $82.1\%$& \multicolumn{1}{c|}{$68.7\%$}  &   $97.2 \% $  & $97.1$ \% &  $95.9 \%$ &  $84.2\%$\\
		Fixed SLM (s)    & $92.6\%$   &  $91.5\%$  &   $84.9\%$  & \multicolumn{1}{c|}{$65.8\%$ }  &   $97.3\%$&  $97.4\%$ &  $95.9\%$ &  $86.4\%$\\
		Learned SLM    & $\bm{94.2\%}$  &  $\bm{92.9\%}$  &   $\bm{91.9\%}$  &  \multicolumn{1}{c|}{$\bm{83.0\%}$} & $\bm{97.9\%}$ & $\bm{97.7\%}$ & $\bm{96.6\%}$ & $\bm{90.3\%}$\\
		\bottomrule
	\end{tabular}
\end{table}

\begin{table}[t!]
	\caption{Relative drop in performance due to embedding compression from $ (24 \times 32) $ to $ (3 \times 4) $.}
	\label{tab:performance_drop}
	\centering
	\begin{tabular}{lcc}
		\toprule
		& \makecell{Logistic \\regression} & \makecell{Single hidden \\layer, 800 units} \\ 
		\midrule
		Lens & \SI{80.0}{\percent}  & \SI{80.8}{\percent}  \\ 
		CA & \SI{20.2}{\percent}  & \SI{28.2}{\percent}  \\ 
		Diffuser & \SI{46.7}{\percent} & \SI{19.4}{\percent}  \\ 
		Fixed SLM (m)& \SI{25.9}{\percent} & \SI{13.4}{\percent} \\ 
		Fixed SLM (s) & \SI{28.9}{\percent} & \SI{11.2}{\percent}  \\ 
		Learned SLM & $\bm{11.9\%}$ & $\bm{8.42\%}$ \\ 
		\bottomrule 
	\end{tabular} 
\end{table}

\begin{figure}[t!]
\begingroup
\renewcommand{\arraystretch}{1} 
\setlength{\tabcolsep}{0.2em} 
	\begin{tabular}{cccccc}
		\\
		&  PSF & 24$\times$32 & 12$\times$16  & 6$\times$8 & 3$\times$4  \\
		Lens
		& \includegraphics[width=0.16\linewidth,valign=m]{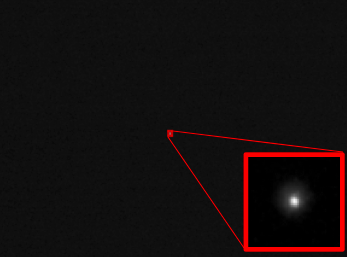} & \includegraphics[width=0.16\linewidth,valign=m]{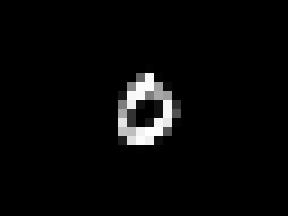}  &\includegraphics[width=0.16\linewidth,valign=m]{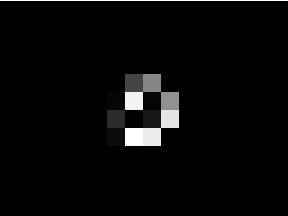} & \includegraphics[width=0.16\linewidth,valign=m]{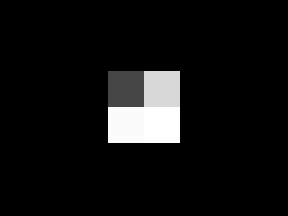} & \includegraphics[width=0.16\linewidth,valign=m]{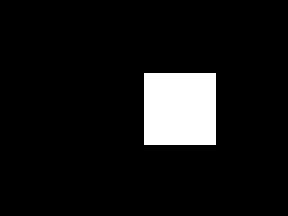}\\[30pt]
		\makecell{Coded \\aperture\\\cite{flatcam}}
		& \includegraphics[width=0.16\linewidth,valign=m]{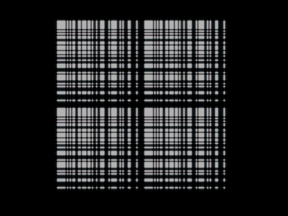} & \includegraphics[width=0.16\linewidth,valign=m]{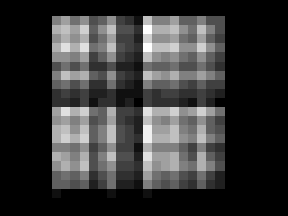}  &\includegraphics[width=0.16\linewidth,valign=m]{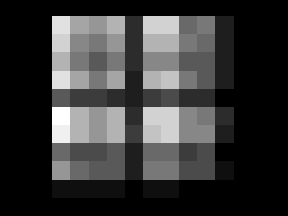} & \includegraphics[width=0.16\linewidth,valign=m]{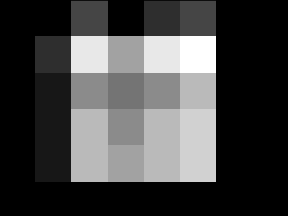} & \includegraphics[width=0.16\linewidth,valign=m]{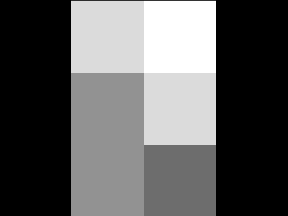}\\[30pt]
		\makecell{Diffuser~\cite{lenslesspicam}}
		& \includegraphics[width=0.16\linewidth,valign=m]{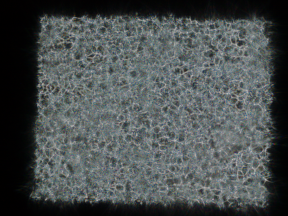} & \includegraphics[width=0.16\linewidth,valign=m]{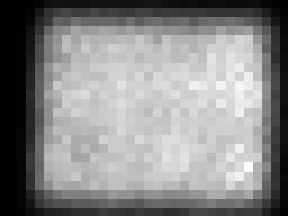}  &\includegraphics[width=0.16\linewidth,valign=m]{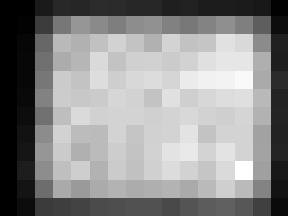} & \includegraphics[width=0.16\linewidth,valign=m]{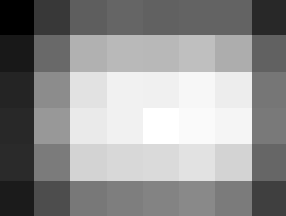} & \includegraphics[width=0.16\linewidth,valign=m]{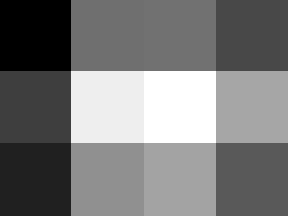}\\[30pt]
		\makecell{Fixed\\SLM\\(m)}
		& \includegraphics[width=0.16\linewidth,valign=m]{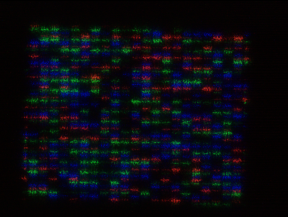} & \includegraphics[width=0.16\linewidth,valign=m]{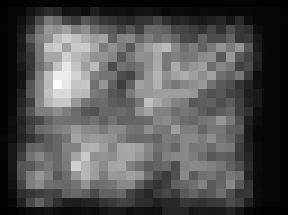}  &\includegraphics[width=0.16\linewidth,valign=m]{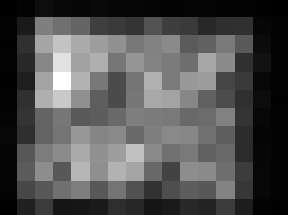} & \includegraphics[width=0.16\linewidth,valign=m]{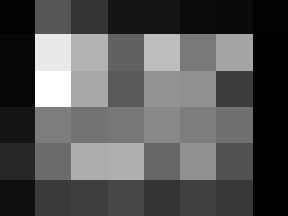} & \includegraphics[width=0.16\linewidth,valign=m]{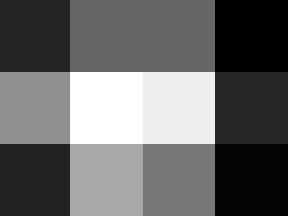}\\[30pt]
		\makecell{Learned\\SLM}
		& \includegraphics[width=0.16\linewidth,valign=m]{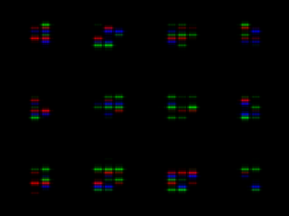} & \includegraphics[width=0.16\linewidth,valign=m]{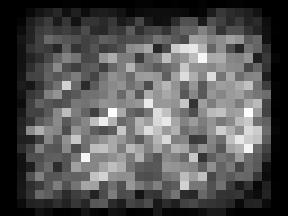}  &\includegraphics[width=0.16\linewidth,valign=m]{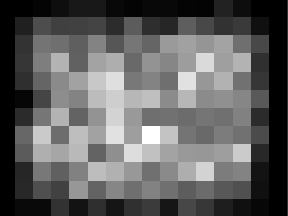} & \includegraphics[width=0.16\linewidth,valign=m]{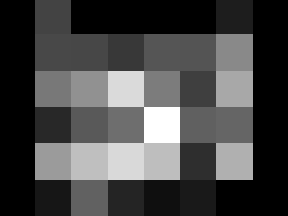} & \includegraphics[width=0.16\linewidth,valign=m]{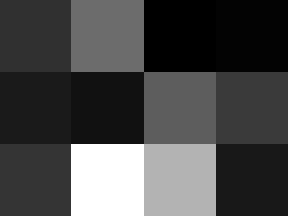}\\[30pt]
	\end{tabular}
\endgroup
\caption{PSFs of the baseline and proposed camera systems, and example sensor embeddings that are simulated with the approach described in \Cref{sec:simulation}. \textit{Lens}, \textit{Diffuser}, and \textit{Fixed SLM (m)} used measured PSFs while \textit{Coded aperture} and \textit{Learned SLM} used simulated ones. \textit{Fixed SLM (s)} is not shown as it is similar to \textit{Fixed SLM (m)}. The PSF for \textit{Learned SLM} is unique per model and embedding dimension. The one shown above was optimized for an embedding dimension of  $ (3\times 4) $ and a two-layer fully connected neural network. All of the learned PSFs can be seen \Cref{sec:learned_psfs}.}
\label{tab:mnist_examples}
\end{figure}

Moreover, the benefits of lensless multiplexing (for both fixed and learned encoders) can be clearly observed as the sensor dimension decreases and the \textit{Lens}' performance deteriorates. \Cref{tab:mnist_examples} provides some insight into this as downsampled measurements of \textit{Lens} can consist of a single pixel. On the other hand, the multiplexing property of lensless cameras leads to much richer measurements, for both \textit{Learned SLM} and the fixed lensless encoders. In \Cref{sec:low_res_recon}, we show example reconstructions for the varying embedding dimensions to show how visual privacy is enhanced by these lower resolution sensor embeddings even with knowledge of the PSF.

The PSFs corresponding to \textit{Learned SLM} for the different embedding dimensions and the two classifiers can be found in \Cref{sec:learned_psfs_vary}. It is interesting to note the PSFs for the embedding dimension of $ (3 \times 4)$ (also visible in the last row of \Cref{tab:mnist_examples}). They resemble the small kernels of convolutional neural networks (CNNs) which has motivated the design of amplitude masks in other end-to-end optimization tasks~\cite{chang2018hybrid,shi2022loen}. In our optimization, these small kernels were not set as an explicit constraint, but the resizing to a $ (3 \times 4)$ sensor embedding may explain why we observe $ 12 $ equally-spaced sub-masks.

\subsection{Robustness to common image transformations}
\label{sec:robustness}

The MNIST dataset is size-normalized and centered, making it ideal for training and testing image classification systems but is not representative of how images may be taken in-the-wild. In this experiment, we evaluate the robustness of lensless encoders to common image transformation, namely we study the effects of:
\begin{itemize}
	\item \textit{Shift}: while maintaining an object height of \SI{12}{\centi\meter} and an object-to-camera distance of \SI{40}{\centi\meter}, shift the image in any direction along the object plane such that it is still fully captured by the sensor.
	\item \textit{Rescale}: while maintaining an object-to-camera distance of \SI{40}{\centi\meter}, set a random height uniformly drawn from $[ \SI{2}{\centi\meter}, \SI{20}{\centi\meter} ]$.
	\item \textit{Rotate}:  while maintaining an object height of \SI{12}{\centi\meter} and an object-to-camera distance of \SI{40}{\centi\meter}, uniformly draw a rotation angle from $ [\SI{-90}{\degree}, \SI{90}{\degree}]$.
	\item \textit{Perspective}:  while maintaining an object height of \SI{12}{\centi\meter} and an object-to-camera distance of \SI{40}{\centi\meter}, perform a random perspective transformation via PyTorch's \texttt{RandomPerspective} with  \SI{100}{\percent} probability and a distortion factor of $ 0.5 $.\footnote{\texttt{RandomPerspective} documentation: \url{https://pytorch.org/vision/main/generated/torchvision.transforms.RandomPerspective.html}}
\end{itemize}

Both the train and test set of MNIST are augmented with the approach described in \Cref{sec:simulation} along with each of the above image transformations, i.e.\ one new dataset per transformation. The same image transformation distribution is used in both training and testing. An illustration of the various image transformations for each camera can be found in \Cref{sec:viz_transformation}.

\begin{table}[t!]
	\caption{MNIST accuracy on randomly transformed test set, simulated accordingly.}
	\label{tab:robustness}
	\centering
	\begin{tabular}{clccccc}
		\toprule
		\makecell{\textit{Embedding} \\ \textit{dimension}  $\downarrow $} &\textit{Encoder}  $\downarrow $& Original & Shift  & Rescale & Rotate & Perspective \\
		\midrule
		
		\multirow{6}{*}{$ (24 \times 32) = 768 $} & Lens&\SI{97.7}{\percent} & $\bm{84.4\%}$ & \SI{84.8}{\percent} & \SI{94.4}{\percent} & \SI{83.0}{\percent} \\
		& CA &   \SI{97.3}{\percent} & \SI{22.8}{\percent} &\SI{82.1}{\percent}  &  \SI{90.9}{\percent} & \SI{31.6}{\percent} \\
		& Diffuser &\SI{95.8}{\percent}& \SI{42.8}{\percent} &\SI{77.8}{\percent}   & \SI{89.7}{\percent} &  \SI{63.2}{\percent}\\
		& Fixed SLM (m) &\SI{97.2}{\percent}& \SI{50.9}{\percent} & \SI{83.0}{\percent} &  \SI{93.2}{\percent} &  \SI{77.4}{\percent} \\
		&Fixed SLM (s) & \SI{97.3}{\percent}&\SI{48.8}{\percent}  & \SI{84.3}{\percent} &  \SI{93.7}{\percent} & \SI{78.0}{\percent} \\
		& Learned SLM &$\bm{97.9\%}$ & $71.7\%$ & $\bm{92.0\%}$ & $\bm{95.0\%}$& $\bm{83.7\%}$\\
		\midrule
		\multirow{5}{*}{$ (6 \times 8) = 48 $} & CA &  \SI{91.0}{\percent} & \SI{14.9}{\percent}  & \SI{71.8}{\percent} & \SI{76.8}{\percent} & \SI{20.1}{\percent}\\
		& Diffuser & \SI{92.8}{\percent} &\SI{26.7}{\percent} & \SI{69.7}{\percent}  & \SI{84.9}{\percent}  & \SI{43.9}{\percent} \\
		& Fixed SLM &\SI{95.9}{\percent} &\SI{29.5}{\percent} & \SI{78.3}{\percent} & \SI{89.3}{\percent}  & \SI{58.7}{\percent} \\
		&Fixed SLM (sim.) &\SI{95.9}{\percent} & \SI{29.0}{\percent} &
		\SI{77.9}{\percent} 
		& \SI{89.7}{\percent} & \SI{57.8}{\percent} \\
		& Learned SLM & $\bm{96.6}\%$ & $\bm{59.3}\%$ & $\bm{88.4\%}$ & $\bm{93.1}\%$ & $\bm{73.4}\%$ \\
		\bottomrule
	\end{tabular}
\end{table}

\begin{table}[t!]
	\caption{Relative drop in performance due to image transformations.}
	\label{tab:performance_drop_trans}
	\centering
	\begin{tabular}{clcccc}
		\toprule
		\makecell{\textit{Embedding} \\ \textit{dimension}  $\downarrow $} &\textit{Encoder}  $\downarrow $ & Shift  & Rescale & Rotate & Perspective \\
		\midrule
		
		\multirow{6}{*}{$ (24 \times 32) = 768 $} & Lens& $\bm{13.6\%}$  & \SI{13.2}{\percent} & \SI{3.38}{\percent} & \SI{15.0}{\percent} \\
		& CA & \SI{76.6}{\percent}& \SI{15.6}{\percent} &\SI{6.58}{\percent} &\SI{67.5}{\percent}\\
		& Diffuser & \SI{55.3}{\percent}&  \SI{18.8}{\percent}&\SI{6.37}{\percent} &\SI{34.0}{\percent} \\
		& Fixed SLM (m)&\SI{47.6}{\percent} & \SI{14.6}{\percent} &\SI{4.12}{\percent} & \SI{20.4}{\percent}\\
		&Fixed SLM (s)& \SI{49.8}{\percent}& \SI{13.4}{\percent} &\SI{3.70}{\percent} &\SI{19.8}{\percent}\\
		& Learned SLM &\SI{26.8}{\percent} & $\bm{6.03\%}$ & $\bm{2.96\%}$& $\bm{14.5\%}$\\
		\midrule
		\multirow{5}{*}{$ (6 \times 8) = 48 $} &  CA & \SI{83.6}{\percent}  &\SI{21.1}{\percent} &\SI{15.6}{\percent} & \SI{77.9}{\percent}\\
		& Diffuser &\SI{71.2}{\percent} & \SI{24.9}{\percent}  & \SI{8.51}{\percent}  & \SI{52.7}{\percent} \\
		& Fixed SLM (m)& \SI{69.2}{\percent}  & \SI{18.4}{\percent}   & \SI{6.88}{\percent}  & \SI{38.8}{\percent} \\
		&Fixed SLM (s)  & \SI{69.8}{\percent}  & \SI{18.8}{\percent} 
		&\SI{6.47}{\percent} & \SI{39.7}{\percent} \\
		& Learned SLM &$\bm{38.6\%}$  & $\bm{8.49\%}$& $\bm{3.62\%}$ & $\bm{24.0\%}$ \\
		\bottomrule
	\end{tabular}
\end{table}

\Cref{tab:robustness} reports the best test accuracy for each optical encoder and for each image transformation when using a two-layer FCNN classifier, see \Cref{sec:nn} for architecture. In the top half of the table, we evaluate the impact of each image transformation for an embedding dimension of $ (24 \times 32) $ to see how lensless imaging techniques fare against a lensed camera. The main difficulty for lensless cameras is shifting as the whole sensor no longer captures multiplexed information, as shown in \Cref{sec:shift_effect}. This leads to a significant reduction in classification accuracy for all lensless approaches. \textit{Learned SLM} is able to cope with shifting much better than the fixed encoding strategies for lensing imaging, most likely because it is able to adapt its multiplexing for such perturbations. For the remaining image transformations, \textit{Learned SLM} is able to outperform the lensed camera, while all of the fixed lensless encodings exhibit worse performance than the lensed camera. 

In the bottom half of \Cref{tab:robustness}, we evaluate the impact of each image transformations for an embedding dimension of $ (6 \times 8) $, for which all lensless approaches exhibited satisfactory performance (above $ 90\% $) on the original dataset and more ``protection'' against post-processing recovery, as shown in \Cref{sec:low_res_recon}. We do not consider \textit{Lens} for this embedding dimension as it performed poorly for the simulated dataset without any transformations. Once again, \textit{Learned SLM} is more robust to image transformations, in particular shifts and perspective changes. \Cref{tab:performance_drop_trans} quantifies the reduction in classification performance due to each of the image transformations, with \textit{Learned SLM} being the least affected among the lensless imaging approaches.
 
\Cref{fig:learned_masks_robust} shows the PSFs corresponding to the \textit{Learned SLM} masks for the two embedding dimensions and the various image transformations. It is worth noting that the masks for the embedding dimension of $ (6 \times 8) $ trained with image transformations are denser than the mask that was obtained without image transformations (\Cref{fig:learned_psf_in48_singlehidden}). This may be a result of a need for more degrees-of-freedom to account for the higher complexity in the input space due to these distortions.

\section{Conclusion}
\label{sec:conclusion}

We have introduced a low-cost and programmable lensless imaging system that can produce robust, privacy-preserving classification results. This is achieved through an end-to-end training that jointly optimizes (1) an optical encoding that produces highly multiplexed embeddings directly at the sensor and (2) the architecture that classifies these privacy-preserving measurements. Our experiments on handwritten digit classification show that the proposed design and training strategy outperforms lensless systems that employ a fixed optical encoding, and is more resilient to common real-world image transformations. Moreover, jointly training the optical encoder and the  digital decoder allows one to reduce the sensor resolution, further enhancing the visual privacy of the measurements.
Adding to the security of the camera is the ability to re-configure the optical encoder if a malicious user obtains information that can be used to decode the sensor embeddings.

For future work, we plan exploit the programmability aspect of our proposed camera with real world data, namely employ hardware-in-the-loop (HITL) techniques as this has been shown to reduce model mismatch~\cite{Peng:2020:NeuralHolography,wright2022deep}. Moreover, a programmable mask allows for time-multiplexed measurements that can be used as additional features for an imaging or classification task~\cite{sweepcam2020,Vargas_2021_ICCV}. 

\paragraph{Limitations} Training end-to-end is expensive due to optical wave propagation simulation.
This could be alleviated by using the hardware itself to perform the forward propagation, as is done in HITL, but this approach comes with its own limitations as forward propagation cannot be parallelized for a batch of training examples. 

Relying on physical devices for computation can also have drawbacks. They are more susceptible to degradation (due to usage and over time) than purely digital computations. Moreover, device tolerances can lead to unwanted differences between two seemingly identical setups. Such differences may be more prominent for low-cost components such as the cheap LCD used in this paper, as opposed to commercial SLMs.

\section*{Acknowledgments and disclosure of funding}
	We thank Sepand Kashani for his input and insight at the initial stages of the project, Julien Fageot and Karen Adam for their feedback and discussions, and Arnaud Latty and Adrien Hoffet for their help in building the experimental prototype.

	This work was in part funded by the Swiss National Science Foundation (SNSF) 
	under grants CRSII5 193826 ``AstroSignals - A New Window on the Universe, with 
	the New Generation of Large Radio-Astronomy Facilities'' (M.~Simeoni) and
	200 021 181 978/1 ``SESAM - Sensing and Sampling: Theory and Algorithms''
	(E.~Bezzam).


{
	\small
	\bibliographystyle{plain}
	\bibliography{references}

\begin{thebibliography}{10}

\bibitem{Antipa:18}
Nick Antipa, Grace Kuo, Reinhard Heckel, Ben Mildenhall, Emrah Bostan, Ren Ng,
  and Laura Waller.
\newblock Diffusercam: lensless single-exposure 3d imaging.
\newblock {\em Optica}, 5(1):1--9, Jan 2018.

\bibitem{flatcam}
M.~Salman Asif, Ali Ayremlou, Aswin Sankaranarayanan, Ashok Veeraraghavan, and
  Richard~G. Baraniuk.
\newblock Flatcam: Thin, lensless cameras using coded aperture and computation.
\newblock {\em IEEE Transactions on Computational Imaging}, 3(3):384--397,
  2017.

\bibitem{lenslesspicam}
Eric Bezzam, Sepand Kashani, Martin Vetterli, and Matthieu Simeoni.
\newblock Lensless{P}i{C}am: A hardware and software platform for lensless
  computational imaging with a {R}aspberry {P}i, 2022.

\bibitem{diffusercam_tut}
C.~Biscarrat, S.~Parthasarathy, G.~Kuo, and N.~Antipa.
\newblock Build your own diffusercam: {T}utorial, 2018.

\bibitem{phlatcam}
Vivek Boominathan, Jesse~K. Adams, Jacob~T. Robinson, and Ashok Veeraraghavan.
\newblock Phlatcam: Designed phase-mask based thin lensless camera.
\newblock {\em IEEE Transactions on Pattern Analysis and Machine Intelligence},
  42(7):1618--1629, 2020.

\bibitem{boominathan2022recent}
Vivek Boominathan, Jacob~T Robinson, Laura Waller, and Ashok Veeraraghavan.
\newblock Recent advances in lensless imaging.
\newblock {\em Optica}, 9(1):1--16, 2022.

\bibitem{admm}
Stephen Boyd, Neal Parikh, Eric Chu, Borja Peleato, and Jonathan Eckstein.
\newblock Distributed optimization and statistical learning via the alternating
  direction method of multipliers.
\newblock {\em Found. Trends Mach. Learn.}, 3(1):1–122, jan 2011.

\bibitem{chang2018hybrid}
Julie Chang, Vincent Sitzmann, Xiong Dun, Wolfgang Heidrich, and Gordon
  Wetzstein.
\newblock Hybrid optical-electronic convolutional neural networks with
  optimized diffractive optics for image classification.
\newblock {\em Scientific reports}, 8(1):1--10, 2018.

\bibitem{Chi:11}
Wanli Chi and Nicholas George.
\newblock Optical imaging with phase-coded aperture.
\newblock {\em Opt. Express}, 19(5):4294--4300, Feb 2011.

\bibitem{deb2022programmable}
Diptodip Deb, Zhenfei Jiao, Alex Bo-Yuan Chen, Misha Ahrens, Kaspar Podgorski,
  and Srinivas~C Turaga.
\newblock Programmable 3d snapshot microscopy with fourier convolutional
  networks, 2022.

\bibitem{10.1117/1.OE.54.2.023102}
Michael~J. DeWeert and Brian~P. Farm.
\newblock {Lensless coded-aperture imaging with separable Doubly-Toeplitz
  masks}.
\newblock {\em Optical Engineering}, 54(2):1 -- 9, 2015.

\bibitem{Goodman2005}
J.W. Goodman.
\newblock {Introduction to Fourier optics}, 2005.

\bibitem{doi:10.1126/science.1127647}
G.~E. Hinton and R.~R. Salakhutdinov.
\newblock Reducing the dimensionality of data with neural networks.
\newblock {\em Science}, 313(5786):504--507, 2006.

\bibitem{sweepcam2020}
Yi~Hua, Shigeki Nakamura, M.~Salman Asif, and Aswin~C. Sankaranarayanan.
\newblock Sweepcam — depth-aware lensless imaging using programmable masks.
\newblock {\em IEEE Transactions on Pattern Analysis and Machine Intelligence},
  42(7):1606--1617, 2020.

\bibitem{huang2013}
Gang Huang, Hong Jiang, Kim Matthews, and Paul Wilford.
\newblock Lensless imaging by compressive sensing.
\newblock In {\em 2013 IEEE International Conference on Image Processing},
  pages 2101--2105, 2013.

\bibitem{10.5555/3045118.3045167}
Sergey Ioffe and Christian Szegedy.
\newblock Batch normalization: Accelerating deep network training by reducing
  internal covariate shift.
\newblock In {\em Proceedings of the 32nd International Conference on
  International Conference on Machine Learning - Volume 37}, ICML'15, page
  448–456. JMLR.org, 2015.

\bibitem{Khan_2019_ICCV}
Salman~S. Khan, Adarsh~V. R., Vivek Boominathan, Jasper Tan, Ashok
  Veeraraghavan, and Kaushik Mitra.
\newblock Towards photorealistic reconstruction of highly multiplexed lensless
  images.
\newblock In {\em Proceedings of the IEEE/CVF International Conference on
  Computer Vision (ICCV)}, October 2019.

\bibitem{adam}
Diederik~P. Kingma and Jimmy Ba.
\newblock Adam: {A} method for stochastic optimization.
\newblock In Yoshua Bengio and Yann LeCun, editors, {\em 3rd International
  Conference on Learning Representations, {ICLR} 2015, San Diego, CA, USA, May
  7-9, 2015, Conference Track Proceedings}, 2015.

\bibitem{lecun1998mnist}
Yann LeCun.
\newblock The mnist database of handwritten digits.
\newblock {\em http://yann. lecun. com/exdb/mnist/}, 1998.

\bibitem{markley2021physicsbased}
Eric Markley, Fanglin~Linda Liu, Michael Kellman, Nick Antipa, and Laura
  Waller.
\newblock Physics-based learned diffuser for single-shot 3d imaging.
\newblock In {\em NeurIPS 2021 Workshop on Deep Learning and Inverse Problems},
  2021.

\bibitem{Matsushima:09}
Kyoji Matsushima and Tomoyoshi Shimobaba.
\newblock Band-limited angular spectrum method for numerical simulation of
  free-space propagation in far and near fields.
\newblock {\em Opt. Express}, 17(22):19662--19673, Oct 2009.

\bibitem{Monakhova:19}
Kristina Monakhova, Joshua Yurtsever, Grace Kuo, Nick Antipa, Kyrollos Yanny,
  and Laura Waller.
\newblock Learned reconstructions for practical mask-based lensless imaging.
\newblock {\em Opt. Express}, 27(20):28075--28090, Sep 2019.

\bibitem{9021989}
Thuong Nguyen~Canh and Hajime Nagahara.
\newblock Deep compressive sensing for visual privacy protection in flatcam
  imaging.
\newblock In {\em 2019 IEEE/CVF International Conference on Computer Vision
  Workshop (ICCVW)}, pages 3978--3986, 2019.

\bibitem{9157577}
Pedram Pad, Simon Narduzzi, Clément Kündig, Engin Türetken, Siavash~A.
  Bigdeli, and L.~Andrea Dunbar.
\newblock Efficient neural vision systems based on convolutional image
  acquisition.
\newblock In {\em 2020 IEEE/CVF Conference on Computer Vision and Pattern
  Recognition (CVPR)}, pages 12282--12291, 2020.

\bibitem{Pan:21}
Xiuxi Pan, Xiao Chen, Tomoya Nakamura, and Masahiro Yamaguchi.
\newblock Incoherent reconstruction-free object recognition with mask-based
  lensless optics and the transformer.
\newblock {\em Opt. Express}, 29(23):37962--37978, Nov 2021.

\bibitem{Pan:22}
Xiuxi Pan, Xiao Chen, Saori Takeyama, and Masahiro Yamaguchi.
\newblock Image reconstruction with transformer for mask-based lensless
  imaging.
\newblock {\em Opt. Lett.}, 47(7):1843--1846, Apr 2022.

\bibitem{paszke2017automatic}
Adam Paszke, Sam Gross, Soumith Chintala, Gregory Chanan, Edward Yang, Zachary
  DeVito, Zeming Lin, Alban Desmaison, Luca Antiga, and Adam Lerer.
\newblock Automatic differentiation in pytorch.
\newblock 2017.

\bibitem{Peng:2020:NeuralHolography}
Y.~Peng, S.~Choi, N.~Padmanaban, and G.~Wetzstein.
\newblock {Neural Holography with Camera-in-the-loop Training}.
\newblock {\em ACM Trans. Graph. (SIGGRAPH Asia)}, 2020.

\bibitem{pinilla2022}
Samuel Pinilla, Seyyed Reza~Miri Rostami, Igor Shevkunov, Vladimir Katkovnik,
  and Karen Eguiazarian.
\newblock Hybrid diffractive optics design via hardware-in-the-loop methodology
  for achromatic extended-depth-of-field imaging, 2022.

\bibitem{DBLP:journals/corr/abs-2106-14577}
Yamin Sepehri, Pedram Pad, Pascal Frossard, and L.~Andrea Dunbar.
\newblock Privacy-preserving image acquisition using trainable optical kernel.
\newblock {\em CoRR}, abs/2106.14577, 2021.

\bibitem{shi2022loen}
Wanxin Shi, Zheng Huang, Honghao Huang, Chengyang Hu, Minghua Chen, Sigang
  Yang, and Hongwei Chen.
\newblock Loen: Lensless opto-electronic neural network empowered machine
  vision.
\newblock {\em Light: Science \& Applications}, 11(1):1--12, 2022.

\bibitem{1227801}
P.Y. Simard, D.~Steinkraus, and J.C. Platt.
\newblock Best practices for convolutional neural networks applied to visual
  document analysis.
\newblock In {\em Seventh International Conference on Document Analysis and
  Recognition, 2003. Proceedings.}, pages 958--963, 2003.

\bibitem{sitzmann2018}
Vincent Sitzmann, Steven Diamond, Yifan Peng, Xiong Dun, Stephen Boyd, Wolfgang
  Heidrich, Felix Heide, and Gordon Wetzstein.
\newblock End-to-end optimization of optics and image processing for achromatic
  extended depth of field and super-resolution imaging.
\newblock {\em ACM Trans. Graph.}, 37(4), jul 2018.

\bibitem{8590781}
Jasper Tan, Li~Niu, Jesse~K. Adams, Vivek Boominathan, Jacob~T. Robinson,
  Richard~G. Baraniuk, and Ashok Veeraraghavan.
\newblock Face detection and verification using lensless cameras.
\newblock {\em IEEE Transactions on Computational Imaging}, 5(2):180--194,
  2019.

\bibitem{Vargas_2021_ICCV}
Edwin Vargas, Julien N.~P. Martel, Gordon Wetzstein, and Henry Arguello.
\newblock Time-multiplexed coded aperture imaging: Learned coded aperture and
  pixel exposures for compressive imaging systems.
\newblock In {\em Proceedings of the IEEE/CVF International Conference on
  Computer Vision (ICCV)}, pages 2692--2702, October 2021.

\bibitem{wetzstein2020inference}
Gordon Wetzstein, Aydogan Ozcan, Sylvain Gigan, Shanhui Fan, Dirk Englund,
  Marin Solja{\v{c}}i{\'c}, Cornelia Denz, David~AB Miller, and Demetri
  Psaltis.
\newblock Inference in artificial intelligence with deep optics and photonics.
\newblock {\em Nature}, 588(7836):39--47, 2020.

\bibitem{wright2022deep}
Logan~G Wright, Tatsuhiro Onodera, Martin~M Stein, Tianyu Wang, Darren~T
  Schachter, Zoey Hu, and Peter~L McMahon.
\newblock Deep physical neural networks trained with backpropagation.
\newblock {\em Nature}, 601(7894):549--555, 2022.

\bibitem{Zhou2020}
Tiankuang Zhou, Lu~Fang, Tao Yan, Jiamin Wu, Yipeng Li, Jingtao Fan, Huaqiang
  Wu, Xing Lin, and Qionghai Dai.
\newblock In situ optical backpropagation training of diffractive optical
  neural networks.
\newblock {\em Photon. Res.}, 8(6):940--953, Jun 2020.

\bibitem{zomet2006}
A.~Zomet and S.K. Nayar.
\newblock Lensless imaging with a controllable aperture.
\newblock In {\em 2006 IEEE Computer Society Conference on Computer Vision and
  Pattern Recognition (CVPR'06)}, volume~1, pages 339--346, 2006.

\end{thebibliography}
}

\newpage

\appendix

\renewcommand\thefigure{\thesection.\arabic{figure}}  
\setcounter{figure}{0}   

\section{Appendix}

\subsection{Modeling incoherent polychromatic wave propagation between two planes} 
\label{sec:model_prop}

\begin{figure}[b]
	\centering
	\begin{subfigure}[b]{.5\textwidth}
		\centering
		\includegraphics[width=0.99\linewidth]{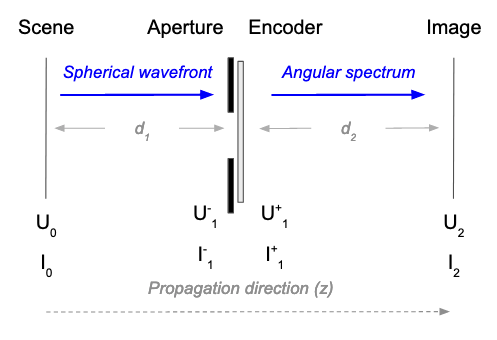} 
		\caption{}
		\label{fig:propagation}
	\end{subfigure}
	\hfill
	\begin{subfigure}[b]{.4\textwidth}
		\centering
		\includegraphics[width=0.99\linewidth]{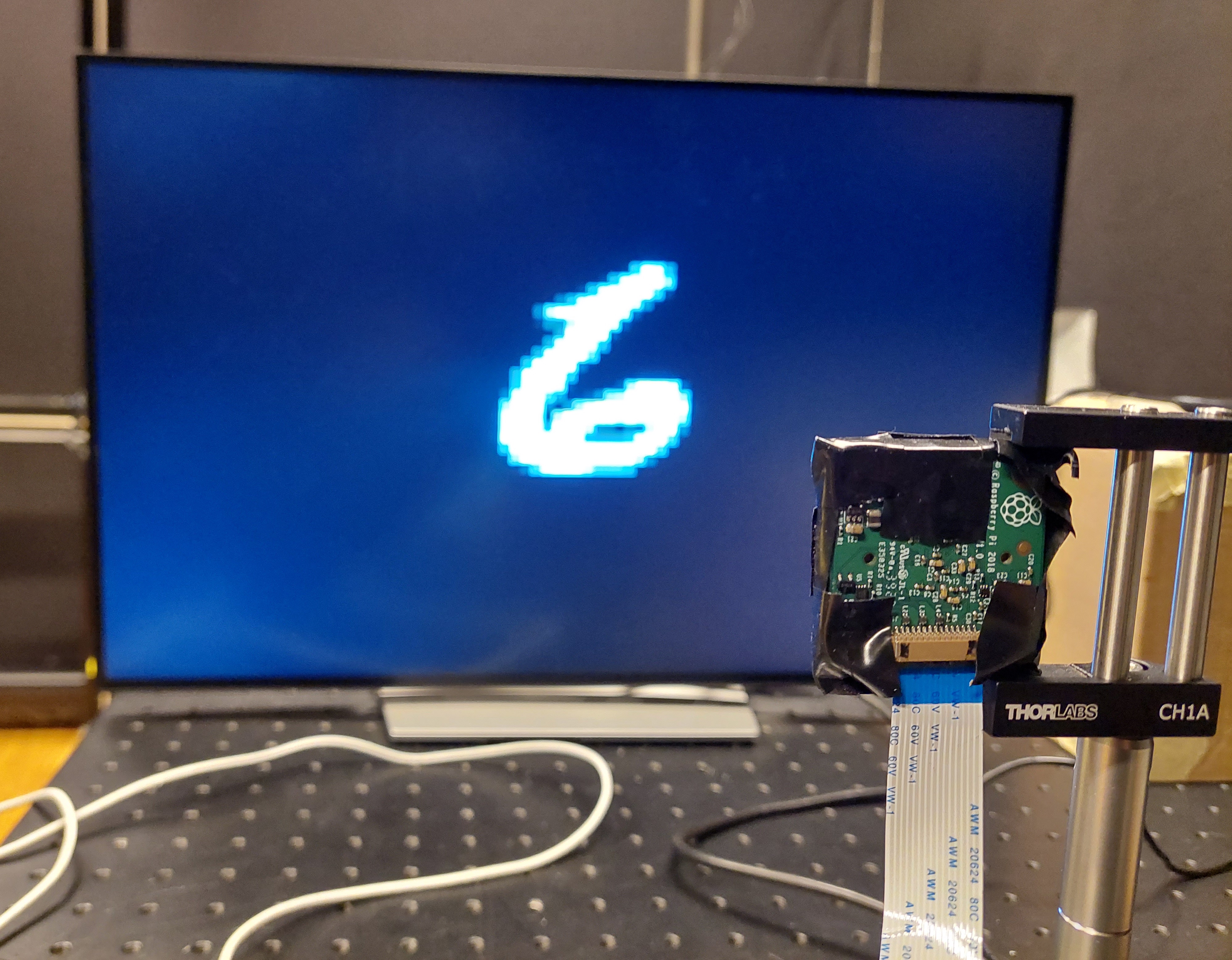}
		\caption{}
		\label{fig:measurement_setup}
	\end{subfigure}
	\caption{(a) Propagation setup. Not drawn to scale for visualization purposes. (b) Example physical measurement setup.}
\end{figure}

\paragraph{Convolutional relationship} \Cref{fig:propagation} illustrates the physical setup assumed by our simulation: the scene of interest is at a fixed distance $d_1$ from the optical encoder, which itself is at a distance $d_2$ from the image plane. We adopt a common assumption from 
\textit{scalar diffraction theory}, namely that image formation is a linear shift-invariant (LSI) system~\cite{Goodman2005}. This implies that there exists an impulse response, i.e.\ a point spread function (PSF), that can be convolved with the input scene to obtain the output image. In optics, this convolutional relationship is between the  \emph{scaled} scene and the image, namely 
\begin{equation}
\label{eq:coherent}
U_2(x, y; \lambda) = \int_{\mathbb{R}}dr\int_{\mathbb{R}} ds\, h(x - r, y - s; \lambda) \Bigg[ \frac{1}{|M|} U_0 \Big( \frac{r}{M}, \frac{s}{M} ; \lambda\Big)\Bigg],
\end{equation}
where $U_0$ and $U_2$ are the wave fields, i.e.\ complex amplitudes, at the scene and the image planes respectively,  $h$ is the PSF, and $M = - d_2 / d_1$ is a magnification factor that also accounts for inversion~\cite{Goodman2005}. Note that this convolution is dependent on the wavelength $ \lambda $.

This LSI assumption significantly reduces the computational load for simulating optical wave propagation, as the convolution theorem and the fast Fourier transform (FFT) algorithm can be used to efficiently evaluate the wave field at the image plane via the spatial frequency domain. An aperture, or some form of cropping, helps to enforce the LSI assumption in order to avoid new patterns from emerging at the sensor for lateral shifts at the scene plane. This is particularly necessary for encoders with a large support, e.g. those of lensless cameras.

For lenses, the PSF in \Cref{eq:coherent} can be approximated by the Fraunhofer diffraction pattern (scaled Fourier transform) of the aperture function. 
For an arbitrary mask, the simplifications resulting from a lens are not possible and a more exact diffraction model is needed to predict the image pattern, e.g.\ Fresnel propagation or the angular spectrum method~\cite{Goodman2005}.

\paragraph{Incoherent, polychromatic illumination} The convolutional relationship in \Cref{eq:coherent} is for coherent illumination, e.g.\ coming from a laser. However, illumination from natural scenes typically consists of diffuse or extended sources which are considered to be \textit{incoherent}. In such cases, impulses at the image plane vary in a statistically independent fashion, thus requiring them to be added on an intensity basis~\cite{Goodman2005}. In order words, for incoherent illumination, the convolution of \Cref{eq:coherent} should be expressed with respect to intensity:
\begin{equation}
\label{eq:incoherent}
I_2(x, y; \lambda) = \int_{\mathbb{R}}dr\int_{\mathbb{R}} ds\, p(x - r, y - s; \lambda) \Bigg[ \frac{1}{|M|^2} I_0 \Big( \frac{r}{M}, \frac{s}{M} ; \lambda\Big)\Bigg],
\end{equation}
where $I_0$ and $I_2$ are the intensities at the scene and image planes respectively, with image intensity defined as the average instantaneous intensity:
\begin{equation}
I(x, y; \lambda) = \langle  | U(x, y; \lambda, t) |^2 \rangle,
\end{equation}
and the intensity PSF is proportional to the squared modulus of the coherent illumination PSF, i.e.\ of \Cref{eq:coherent}:
\begin{equation}
\label{eq:intensity_psf}
p(x, y) \propto | h(x, y)|^2.
\end{equation}

For polychromatic simulation, each wavelength has to be simulated independently. Converting this multispectral data to RGB is typically done in two steps: (1) mapping each wavelength to the XYZ coordinates defined by the International Commission on Illumination\footnote{\url{https://en.wikipedia.org/wiki/CIE_1931_color_space}} and (2)  converting to red-green-blue (RGB) values based on a reference white.\footnote{\url{http://www.brucelindbloom.com/index.html?Eqn_RGB_XYZ_Matrix.html}}

\subsection{Simulating propagation with an image of the desired scene}
\label{sec:simulation}

When simulating the propagation between two planes, as shown in \Cref{fig:propagation}, one may wish that the scene in the $ I_0 $ plane corresponds to the content of a digital image. This section describes how to process an image such that its output corresponds to (1) a  scene at a distance $ d_1 $ from the camera, (2) content from the original image having a height $h_{\text{obj}}$, and (3) a measurement taken by a camera of a known PSF.

An RGB image can be interpreted as image intensities at three wavelengths: red, green, and blue~\cite{sitzmann2018}. We use the following wavelengths for red, green, and blue respectively: \SI{640}{\nano\meter}, \SI{550}{\nano\meter}, and \SI{460}{\nano\meter}. For a grayscale image, such as images from MNIST~\cite{lecun1998mnist}, the same data can be used across channels, or it can be convolved with a grayscale version of the PSF.

Concretely, given an image $\bm{x} \in \mathbb{R}^{H\times W \times C}$ with $ C $ channels and an intensity PSF $\bm{p} \in \mathbb{R}^{H_{\text{PSF}}\times W_{\text{PSF}} \times C}$, the simulation of a sensor measurement $\bm{v} \in \mathbb{R}^{DH_{\text{PSF}}\times DW_{\text{PSF}} \times C}$ (with an optional downsampling factor $ D \geq 1 $) can be summarized by the following steps:
\begin{enumerate}
	\item Resize $\bm{x} $ to the PSF's dimension to obtain $\bm{x}_r \in \mathbb{R}^{H_{\text{PSF}}\times W_{\text{PSF}} \times C}$, while preserving $\bm{x} $'s original aspect ratio and scaling it to correspond to a desired object height (or width). The details of this rescaling are explained in \Cref{sec:rescale}.
	\item Convolve each channel of $\bm{x}_r $ with the corresponding PSF channel to obtain $\bm{y} \in \mathbb{R}^{H_{\text{PSF}}\times W_{\text{PSF}} \times C}$. Due to large convolution kernels, this is typically best to do in the spatial frequency domain, where convolution corresponds to an element-wise multiplication, and the FFT algorithm can be used to efficiently move between domains.
	\item If $D \neq 1$, downsample the convolution output to the sensor resolution.  We apply bilinear interpolation for this resizing.
	\item Add noise at a desired signal-to-noise ratio (SNR). More on this in \Cref{sec:noise}. 
\end{enumerate}

Having a faithful estimate of the intensity PSF $\bm{p} $ is the most vital part of the above simulation. For a fixed optical encoder, if a physical setup is available, the best approach may be to simply measure the PSF by placing a point source (e.g.\ a white LED behind a pinhole as shown in \Cref{fig:measuring_psf}) at the desired distance and taking the resulting measurement as the intensity PSF. Some post-processing may be necessary to remove sensor noise and balance color channels. For encoders that have no parametric function, e.g.\ pseudo-random diffusers~\cite{Antipa:18}, measuring the PSF may be the only viable option.

For encoders that have a parametric function, e.g.\ with lenses and/or SLMs, it is possible to simulate the intensity PSF. This is in fact necessary for most end-to-end optimization techniques, unless forward-/back-propagation are done directly with hardware~\cite{Zhou2020}. Even for a parametric encoder, it can be useful to measure the PSF (if a physical setup is available) in order to reduce mismatch due to model assumptions / simplifications~\cite{phlatcam}. In \Cref{sec:psf_modeling}, we describe our modeling of the PSF of an SLM at a particular wavelength, and explain how we account for the specifications of the ST7735R component and the Raspberry Pi High Quality Camera. 

\paragraph{Using physical measurements} The above simulation of \Cref{fig:propagation} seeks to replicate the physical measurement setup shown in \Cref{fig:measurement_setup}, namely projecting the image of the desired scene on a display at a distance $ d_1 $ from the camera. While such a measurement would produce more realistic results, it can be very time-consuming for an \textit{entire dataset} of images. If this dataset is to be used for a task with a fixed optical encoder, e.g.\ a lens or diffuser, it  may be worth the time and effort as the measurement only has to be done once. For a task that seeks to optimize the optical encoder in an end-to-end fashion, new measurements would have to be performed during training whenever updates are made to the optical encoder. This is highly impracticable for optical encoders that require precise fabrication~\cite{sitzmann2018,phlatcam,markley2021physicsbased}. In the case of programmable optical encoders, alternating between physical measurements and updating the optical encoder lends itself to hardware-in-the-loop / physics-aware training~\cite{Peng:2020:NeuralHolography,wright2022deep}. This has the potential to reduce model-mismatch but at the cost of longer training, due to acquisition time and a lack of parallelization. This technique is outside the scope of this work.

\subsection{Rescaling image to PSF resolution for a desired object height}
\label{sec:rescale}

The goal of this step in the simulation of \Cref{sec:simulation} is to rescale a digital image $\bm{x} \in \mathbb{R}^{H\times W \times C}$ such that its convolution with a digital PSF $\bm{p} \in \mathbb{R}^{H_{\text{PSF}}\times W_{\text{PSF}} \times C}$ corresponds to the setup in \Cref{fig:propagation} for an object of height $h_{\text{obj}}$. For such a configuration,
namely a scene-to-encoder distance of $d_1$ and an encoder-to-image distance of $d_2$,
the object height \emph{at the sensor} is given by:
\begin{equation}
h_{\text{sensor}} = h_{\text{obj}} (d_2 / d_1) = |M|h_{\text{obj}}.
\end{equation}
For our simulation we are interested in the number of pixels that this height corresponds to. If our PSF was measured for the above distances with a sensor resolution of $ (H_{\text{sensor}} \times W_{\text{sensor}} ) $ pixels and a pixel pitch of $ \Delta $,\footnote{The sensor may have a different resolution than the PSF used in simulation as we may wish to downsample the measured PSF for a lighter computational load, especially if we are simulating a large dataset during training. Note that this simplification is possible for PSF's with a broad support such as those of diffusers and SLMs, but doing so with a PSF with a very small support, e.g.\ a lens, can significantly hurt the simulation quality.} the sensor will have captured a PSF for a scene of the following physical dimensions: 
\begin{equation}
(h_{\text{scene}} \times w_{\text{scene}} ) = \Big(\dfrac{\Delta H_{\text{sensor}} }{|M|} \times \dfrac{\Delta W_{\text{sensor}}  }{|M|}  \Big).
\end{equation}
Consequently, the object height \textit{in pixels} is approximately given by:
\begin{equation}
H_{\text{pixel}} = \text{round}\Big( \dfrac{h_{\text{obj}} H_{\text{PSF}}}{h_{\text{scene}}} \Big).
\end{equation}
Therefore, to rescale the original input image $\bm{x} \in \mathbb{R}^{H\times W \times C}$ to the PSF resolution, while preserving its aspect ratio and scaling it such that it corresponds to the desired object height, we need to perform the following steps:
\begin{enumerate}
	\item Resize  $\bm{x} $ to $ \big(  \text{round}(SH) \times \text{round}(SW) \times C \big) $ where $S = (H_{\text{pixel}} / H)$ .
	\item Pad above to $(H_{\text{PSF}}\times W_{\text{PSF}} \times C)$.
\end{enumerate}
The resulting image $\bm{x}_r \in \mathbb{R}^{H_{\text{PSF}}\times W_{\text{PSF}} \times C}$ can then be convolved with the PSF to simulate a propagation as in \Cref{fig:propagation}.

\subsection{Adding noise at a desired signal-to-noise ratio}
\label{sec:noise}

Different types of noise can be added during simulation. In practice, read noise at a sensor follows a Poisson distribution with respect to the input. However, as this distribution is typically not differentiable with respect to the optical encoder parameters, Gaussian noise is used instead~\cite{sitzmann2018,markley2021physicsbased}.

In order to add generated noise to a signal and obtain a desired signal-to-noise ratio (SNR), the generated noise must be scaled appropriately. SNR (in dB) is defined as:
\begin{equation}
\text{SNR} = 10 \log_{10} (\sigma_S^2 / \sigma_N^2),
\end{equation}
where $\sigma_S^2$ is the clean image variance and $\sigma_N^2$ is the generated noise variance. For a target SNR $T$, the generated noise can be scaled with the following factor:
\begin{equation}
k = \sqrt{ \dfrac{\sigma_S^2 }{\sigma_N^2 10 ^{(T/10)}}}.
\end{equation}

In our simulation, we generate noise following a Poisson distribution, as we do not backpropagate through noise generation to the optical encoder parameters.

\subsection{Point spread function modeling for a spatial light modulator}
\label{sec:psf_modeling}

Our modeling of the PSF for propagation through a spatial light modulator (SLM) is similar to that of~\cite{sitzmann2018}, namely for each wavelength $ \lambda $, we simulate the propagation in \Cref{fig:propagation}:
\begin{enumerate}
	\item \textit{From the scene to the optical element}: propagation is modeled by spherical wavefronts. Assuming a point source at the scene plane $ U_0 $,  we have the following wave field at the aperture plane:
	\begin{equation}
	U_1^-(x, y; z=d_1, \lambda) = \exp \Big(j \frac{2\pi}{\lambda} \sqrt{x^2 + y^2 +  d_1^2} \Big),
	\end{equation}
	where $ d_1 $ is the distance between the scene and the camera aperture.
	\item\textit{At the optical element}: the wave field is multiplied with a potentially complex-valued mask pattern $M(x, y)$ corresponding to the SLM:
	\begin{equation}
	\label{eq:after_mask}
	U_1^+(x, y; z=d_1, \lambda) = U_1^-(x, y; z=d_1,\lambda)  M(x, y).
	\end{equation}
	Note that an infinitesimally small distance is assumed between the opening of the aperture $ U_1^- $ and the exit of the SLM $ U_1^+ $.
	\item \textit{To the sensor}: free-space propagation according to scalar diffraction theory~\cite{Goodman2005} as light is diffracted by the optical element. We employ the bandlimited angular spectrum method (BLAS) which produces accurate simulations for both near- and far-field~\cite{Matsushima:09}. This yields the following wave field PSF at the sensor plane:
	\begin{equation}
	\label{eq:output}
	U_2(x, y; z=d_1 + d_2, \lambda) = \mathcal{F}^{-1}  \Big(  \mathcal{F} \big( U_1^+(x, y; z=d_1, \lambda)  \big) H(u, v; z=d_2, \lambda) \Big),
	\end{equation}
	where $ \mathcal{F}$ and $\mathcal{F}^{-1} $ denote the spatial Fourier transform and its inverse,  $u, v$ are spatial frequencies for $x, y$, and the free-space frequency response $ H(u, v; z=d_2, \lambda) $ according BLAS is given by:
	\begin{align}
	H(u, v; z=d_2, \lambda ) = e^ {j \frac{2 \pi}{\lambda} d_2 \sqrt{1 - (\lambda u)^2 - (\lambda v)^2} } \text{rect}\Big(\frac{u}{2u_{\text{limit}}}\Big)  \text{rect}\Big(\frac{v}{2v_{\text{limit}}}\Big),
	\end{align}
	where the bandlimiting frequencies are given by
	\begin{align}
	u_{\text{limit}} = \dfrac{\sqrt{(d_2 / S_x )^2  + 1} }{\lambda},  \quad v_{\text{limit}} = \dfrac{\sqrt{( d_2 / S_y )^2  + 1} }{\lambda},
	\end{align}
	and $ (S_x \times S_y) $ are the physical dimensions of the propagation region, in our case the physical dimensions of the sensor.
	\item As we are simulating incoherent light, we require the squared modulus of the wave field PSF:
	\begin{equation}
	P(x, y; z = d_1 + d_2, \lambda) = \big| U_2(x, y; z=d_1 + d_2, \lambda)  \big|^2.
	\end{equation}
	
	Note that we assume the intensity PSF to be equal to the squared modulus of the wave field PSF, as is done in~\cite{sitzmann2018}. In theory, they are simply proportional and multiple realizations would be needed to estimate this statistical quantity~\cite{Goodman2005}.

\end{enumerate}

\subsubsection{Modeling the spatial light modulator} 
A key component in the above PSF simulation is modeling the complex-valued mask $M(x, y)$ associated with the SLM. Two assumptions are commonly made in its modeling:
\begin{itemize}
	\item The mask is assumed to be either a phase transformation, i.e.\ $|M(x, y)| = 1$, or an amplitude transformation, i.e.\ $M(x, y) \in \mathbb{R}$. 
	\item The mask is discretized according to the SLM resolution. This approximation neglects \emph{deadspace} (or equivalently the fill factor) of individual pixels.
\end{itemize}

We model the SLM as a superposition of apertures for each adjustable pixel:
\begin{align}
\label{eq:mask_gen}
M(x, y) = \sum_{k}^K w_{k} A(x - x_k, y-y_k),
\end{align}
where the complex-valued weights $\{w_{k}\}_{k=1}^K$ satisfy $|w_{k}| \leq 1$, the coordinates $ \{(x_k, y_k)\}_{k=1}^{K} $ are the centers of the SLM pixels, and the aperture function $A(\cdot)$ is assumed to be identical for each SLM pixel. This model takes into account deadspace but assumes that no stray light passes between the pixels. 

While numerical discretization may not be able to perfectly sample $M(x, y)$ to account for arbitrary shifts of $A(\cdot)$ in \Cref{eq:mask_gen},
these shifts can be accounted for in the spatial frequency domain $(u, v)$:
\begin{align}
\mathcal{F} (M(x, y)) = M(u, v) = A(u, v) \sum_{k}^K w_{k} e^{j u x_k} e^{j v y_k}.
\end{align}
\Cref{eq:after_mask} for the wave field at the exit of the SLM then becomes
\begin{align}
U_1^+(x, y; z=d_1, \lambda) = U_1^-(x, y; z=d_1,\lambda)  \mathcal{F}^{-1} (M(u, v)).
\end{align}
While this allows for arbitrary shifts, it requires an additional FFT and can be expensive when tracking gradients in order to optimize the SLM weights $w_{k}$. 

A cheaper way to account for deadspace is to discretize $M(x, y)$ at a finer resolution than that of the SLM, and only modulate those pixels which fall within the individual SLM pixel apertures (by setting the appropriate $ w_k $ value). While optimizing the SLM weights in our end-to-end approach (for the experiments in \Cref{sec:experiments}), we adopt this latter simplification which is much more tractable when tracking gradients.

Other parameters that have been modeled for SLMs in holography and that are applicable to the context of imaging include: non-linear mapping between voltage-to-phase/amplitude of the individual SLM pixels, Zernike coefficients to model deviations from theoretical propagation models, and content-dependent undiffracted (stray) light~\cite{Peng:2020:NeuralHolography}.

\subsubsection{Specifics for the ST7735R component and the Raspberry Pi High Quality Camera}
\label{sec:ST7735R}

\begin{figure}[t!]
	\centering
	\begin{subfigure}{.46\textwidth}
		\centering
		\includegraphics[width=0.99\linewidth]{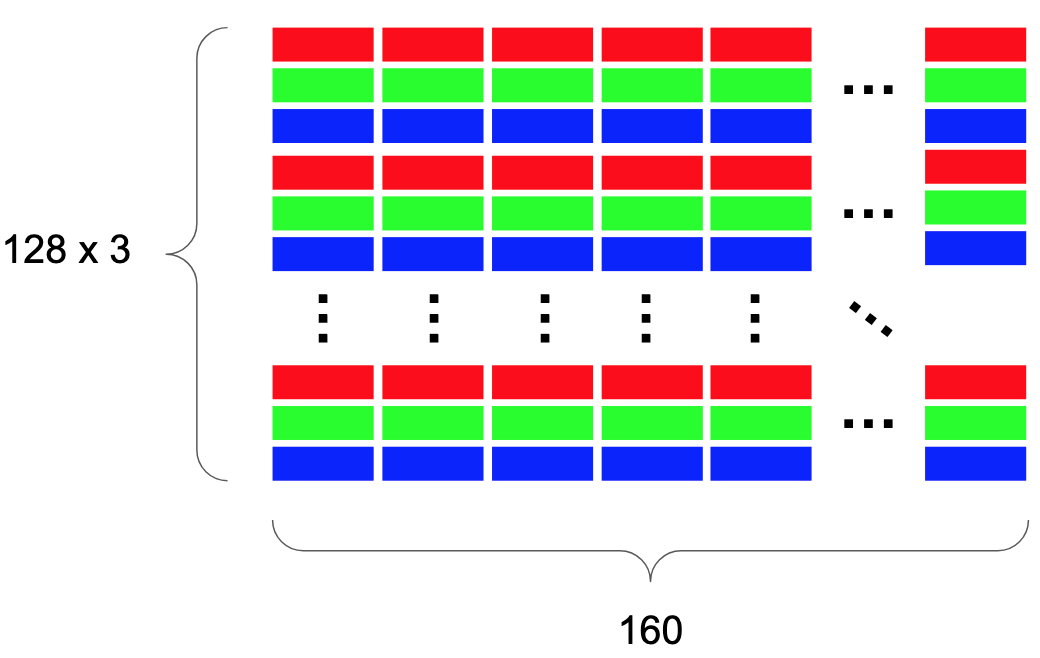}
		\caption{}
		\label{fig:slm_pixel_layout}
	\end{subfigure}
	\hfill
	\begin{subfigure}{.38\textwidth}
		\centering
		\includegraphics[width=0.99\linewidth]{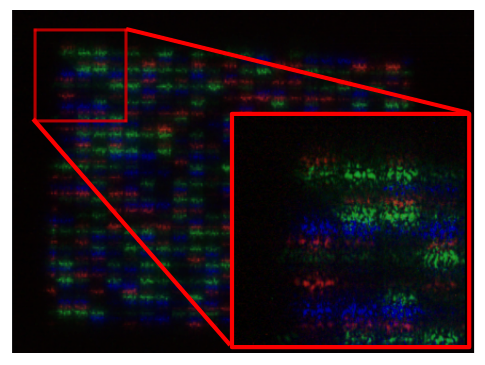}
		\caption{}
		\label{fig:adafruit_section}
	\end{subfigure}	
	\caption{Visualizing pixel layout of the ST7735R component. (a) Red, green, blue color filter arrangement. (b) Zooming into section of a measured point spread function for a random pattern.}
\end{figure}

As the ST7735R component is originally intended to serve as a color display, it has an interleaved pattern of red, green, and blue filters as shown in \Cref{fig:slm_pixel_layout},\footnote{More information can be found on the device driver datasheet: \url{https://cdn-shop.adafruit.com/datasheets/ST7735R_V0.2.pdf}} which can be modeled as a wavelength-dependent version of \Cref{eq:mask_gen}
\begin{align}
\label{eq:adafruit_mask}
M(x, y; \lambda) &= \sum_{c\in\{R,G,B\}} \!\!\!\!\!\! F_c(\lambda)  \, \sum_{k_c}^{K_c}
w_{k_c}
A(x - x_{k_c}, y-y_{k_c}),
\end{align}
where:
\begin{itemize}
	\item $ \{ F_c(\cdot)\}_{c\in\{R,G,B\}} $ is the wavelength-response of each color filter,
	\item $ \Big\{  \{w_{k_c}\}_{k_c=1}^{K_c} , c\in\{R,G,B\}  \Big\} $ are \emph{real}-valued weights for the red, green, and blue sub-pixels, and $ \Big\{  \{(x_{k_c}, y_{k_c})\}_{{k_c}=1}^{K_c}, c\in\{R,G,B\}  $ are their respective centers.
\end{itemize}

The display has a resolution of $128\times 160$ color pixels, with three sub-pixels per color pixel as shown in \Cref{fig:slm_pixel_layout}. The dimension of each sub-pixel is $ (\SI{0.06}{\milli\meter} \times \SI{0.18}{\milli\meter}) $ and the dimension of the entire screen is $ (\SI{28.03}{\milli\meter} \times \SI{35.04}{\milli\meter}) $.\footnote{ST7735R breakout board datasheet: \url{https://cdn-shop.adafruit.com/datasheets/JD-T1800.pdf}} If we assume a uniform spacing of sub-pixels, this corresponds to a pixel pitch of roughly $ (\SI{0.073}{\milli\meter} \times \SI{0.22}{\milli\meter}) $ 
and a fill-factor of $82\%$, namely deadspace of $18\%$ around each sub-pixel. 

The Raspberry Pi High Quality Camera\footnote{Raspberry Pi High Quality Camera datasheet: \url{https://cdn-shop.adafruit.com/product-files/4561/4561+Raspberry+Pi+HQ+Camera+Product+Brief.pdf}} uses the Sony IMX477R back-illuminated sensor which has the following specifications: $3040 \times 4056$ pixel resolution, $7.9\si{\milli\meter}$ sensor diagonal, and a pixel size of $ (1.55\si{\micro\meter} \times 1.55\si{\micro\meter} ) $, which corresponds to full sensor dimensions of $ (\SI{4.71}{\milli\meter} \times \SI{6.29}{\milli\meter}) $. As the display of the ST7735R component is larger than the sensor, we use a subset of its pixels that covers the sensor area. From the pixel pitch of the ST7735R component determined above $ (\SI{0.073}{\milli\meter} \times \SI{0.22}{\milli\meter}) $, it can be concluded that the number of SLM sub-pixels that overlap the sensor is around $ 64 \times 22 $. Moreover, for enforcing the LSI assumption described in \Cref{sec:model_prop}, we crop the SLM such that $ 80\% $ of sensor surface is exposed, which corresponds to $ 51 \times 22  = 1122 $ SLM sub-pixels. This is the number of SLM sub-pixels that we optimize in \Cref{sec:experiments}

\subsection{Detailed description about baseline and proposed point spread functions}
\label{sec:baseline}

\begin{figure}[t!]
	\centering
	\begin{subfigure}[b]{.22\textwidth}
		\centering
		\includegraphics[width=0.99\linewidth]{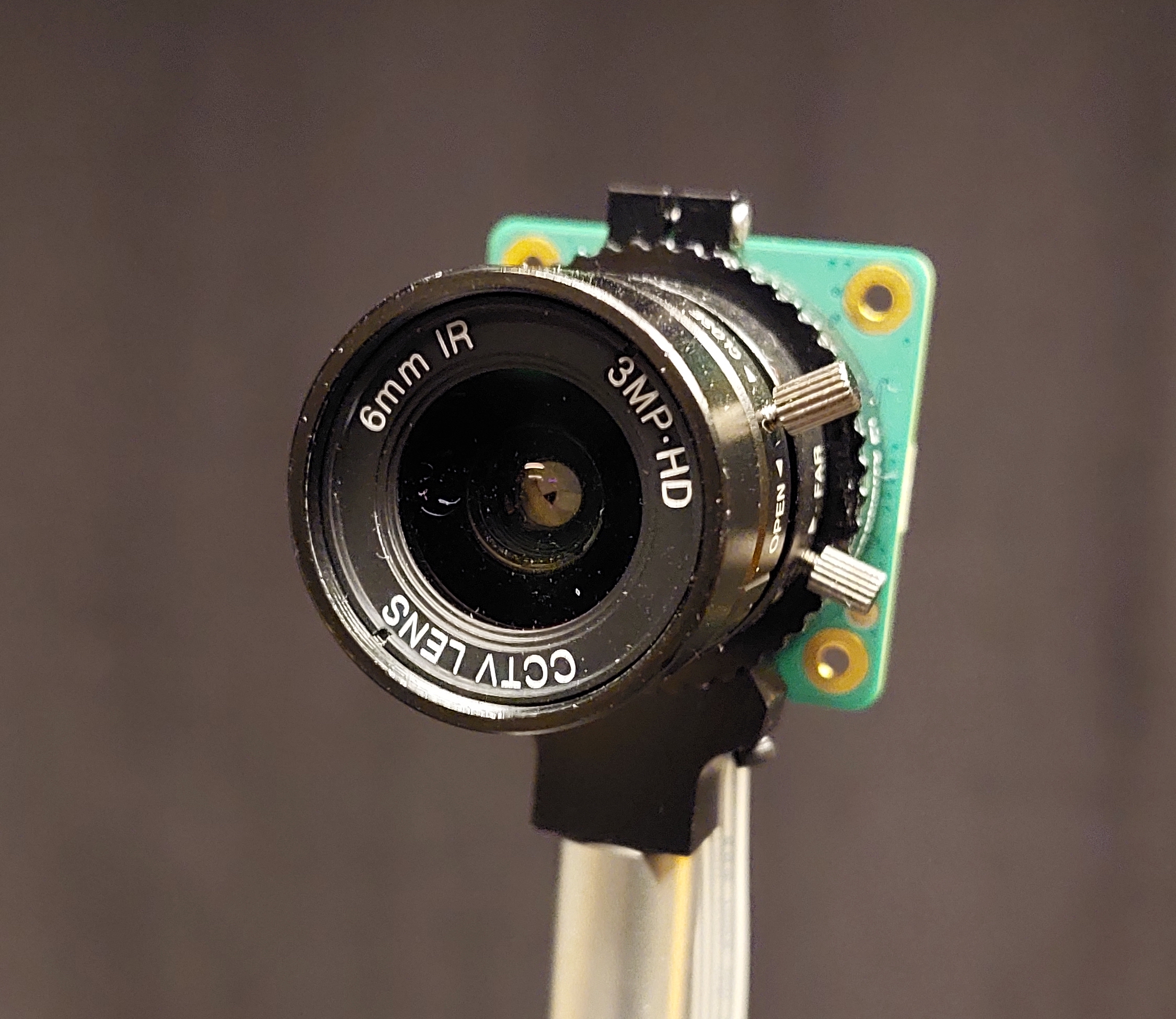}
		\caption{}
		\label{fig:lensed_camera}
	\end{subfigure}
	\begin{subfigure}[b]{.22\textwidth}
		\centering
		\includegraphics[width=0.99\linewidth]{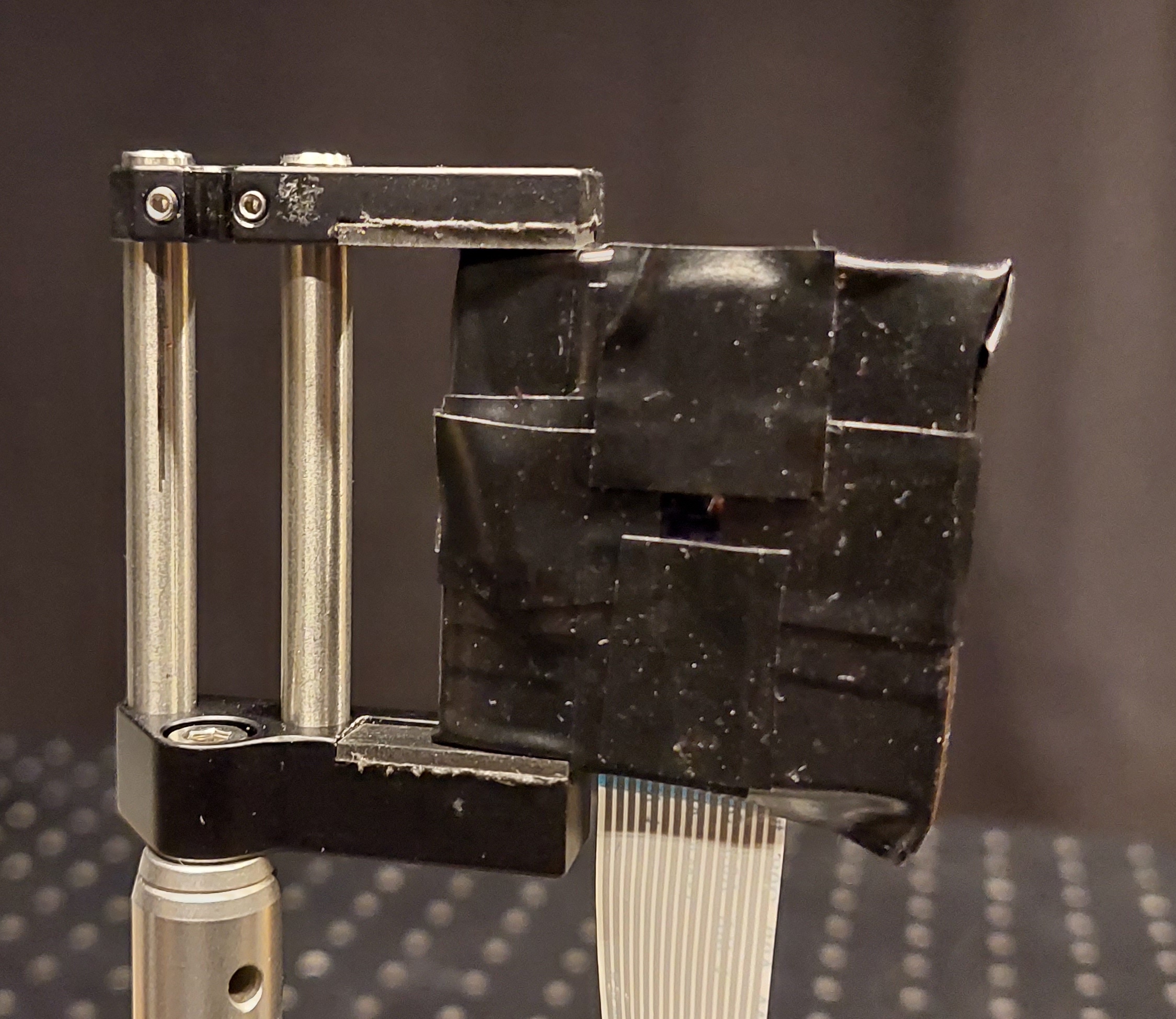}
		\caption{}
		\label{fig:diffuser}
	\end{subfigure}
	\begin{subfigure}[b]{.25\textwidth}
		\centering
		\includegraphics[width=0.99\linewidth]{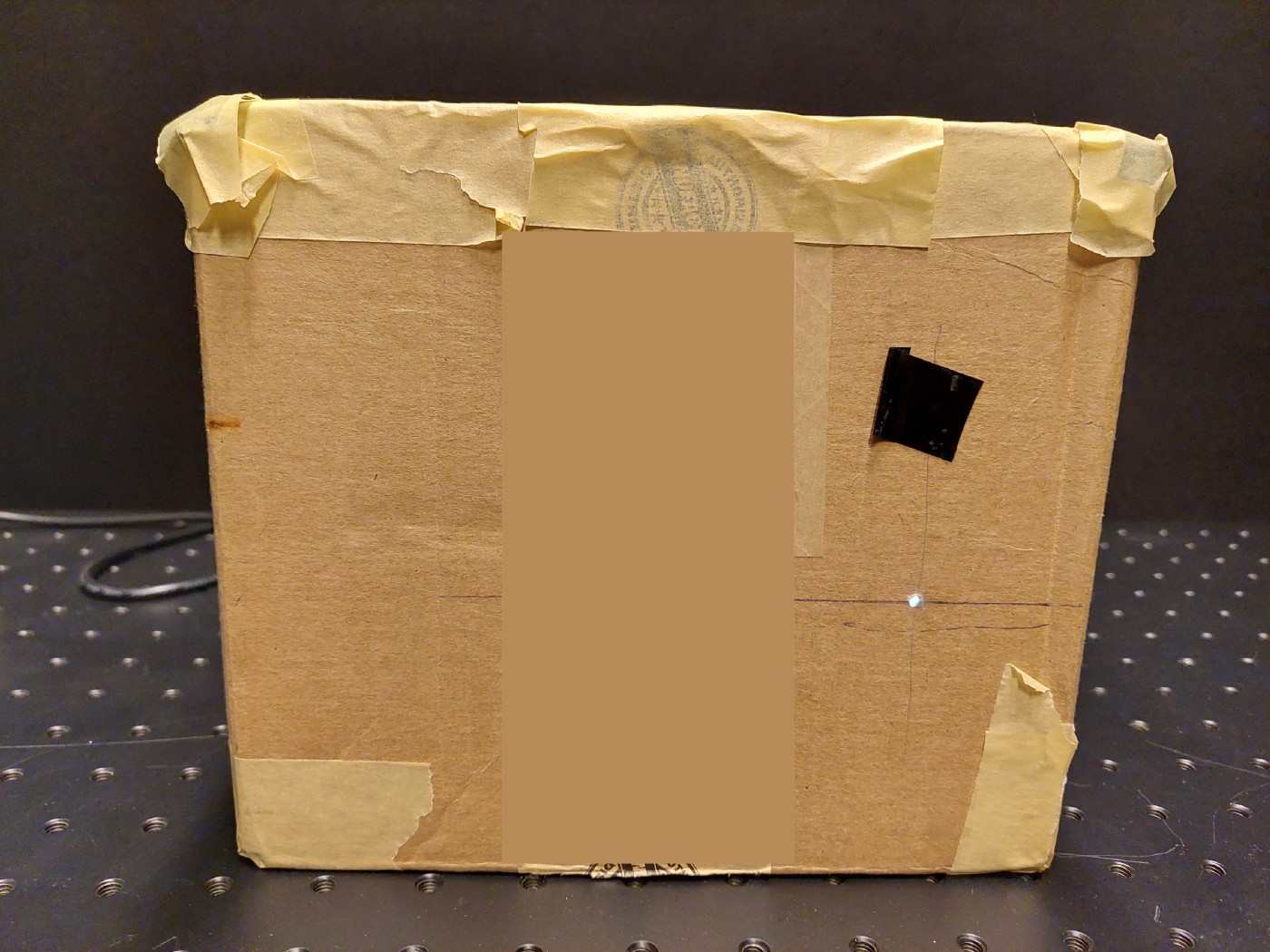}
		\caption{}
		\label{fig:measuring_psf}
	\end{subfigure}
	\caption{Baseline cameras: (a) lensed and (b) diffuser-based. (c) Point source for measuring PSF.}
\end{figure}

For the experiments in \Cref{sec:experiments}, our baseline and proposed imaging systems use the Raspberry Pi High Quality Camera, either for the PSF measurement or in simulating the PSF.
Measured PSFs are obtained by placing a white LED behind a pinhole aperture, as shown in \Cref{fig:measuring_psf}, at the target distance (\SI{40}{\centi\meter}), and measuring the response in an environment with no external light.

Simulated PSFs are obtained by using the approach described in \Cref{sec:psf_modeling}. Unless noted otherwise, the scene, encoder, and images planes ($ U_0, U_1^+, U_2 $ respectively in \Cref{fig:propagation}) for simulating the PSFs take on the size and resolution of the Raspberry Pi High Quality Camera: $3040 \times 4056$ pixel resolution, and a pixel size of $ (1.55\si{\micro\meter} \times 1.55\si{\micro\meter} ) $.

Below are technical details regarding each PSF:
\begin{itemize}
		\item \textit{Lens}: measured PSF for the camera shown in \Cref{fig:lensed_camera}, which has a $6\si{\milli\meter}$ wide angle lens\footnote{6mm Wide Angle Lens for Raspberry Pi HQ Camera datasheet: \url{https://cdn-shop.adafruit.com/product-files/4563/4563-datasheet.pdf}} focused at \SI{40}{\centi\meter}. The lens and its objective have a thickness of \SI{34}{\milli\meter}, and the lens is \SI{7.53}{\milli\meter} from the sensor.
		\item \textit{CA} (coded aperture): a binary mask is generated by (1) generating a length-$63$ maximum length sequence (MLS) binary array,\footnote{Using the SciPy function \texttt{max\textunderscore len\textunderscore seq}: \url{https://docs.scipy.org/doc/scipy/reference/generated/scipy.signal.max_len_seq.html}} (2) repeating the sequence to create a $ 126 $-length sequence, and (3) computing the outer product with itself to create a $ 126\times126 $ matrix. 
		These are the same steps for generating a coded aperture mask, as in~\cite{flatcam}, except that we use a shorter MLS sequence ($ 63 $ instead of $ 255 $) to obtain a feature size of $ \SI{30}{\micro\meter} $ (as in~\cite{flatcam}). The mask covers $ 80\% $ of the sensor height (as \textit{Fixed SLM (s)} and \textit{Learned SLM} below). For the PSF, we simulate the mask's diffraction pattern for a distance of $ d_2 = \SI{0.5}{\milli\meter} $, matching the distance in~\cite{flatcam}. 
		\item \textit{Diffuser}: measured PSF for the camera shown in \Cref{fig:diffuser}, where the diffuser is placed roughly \SI{4}{\milli\meter} from the sensor. The diffuser is double-sided tape as in the DiffuserCam tutorial~\cite{diffusercam_tut}. In~\cite{lenslesspicam}, the authors demonstrate the effectiveness of this simple diffuser when used with the Raspberry Pi High Quality Camera. It is less than \SI{1}{\milli\meter} thick and is placed roughly \SI{4}{\milli\meter} from the sensor.
		\item \textit{Fixed SLM (m)}: measured PSF for the proposed camera shown in \Cref{fig:prototype_labeled} for a random pattern. With a stepper motor, the mask-to-sensor distance is programmatically set to \SI{4}{\milli\meter} to match the distance of the diffuser-based camera.
		\item \textit{Fixed SLM (s)}: simulated PSF for the proposed camera, using the approach described in \Cref{sec:psf_modeling} for a random set of SLM amplitude values and a mask-to-sensor distance of \SI{4}{\milli\meter}. The aperture is set such that SLM pixels covering $ 80\% $ of the sensor surface area are exposed. This corresponds to  $ 51 \times 22 = 1122 $ SLM sub-pixels as described in \Cref{sec:ST7735R}.
		\item \textit{Learned SLM}: simulated PSF for the proposed camera that is obtained by optimizing \Cref{eq:mnist} for the SLM weights, and simulating the corresponding PSF with  the approach described in \Cref{sec:psf_modeling} for a mask-to-sensor distance of \SI{4}{\milli\meter}. Like \textit{Fixed SLM (s)}, SLM pixels that cover $ 80\% $ of the sensor surface area are used, corresponding to  $ 51 \times 22 = 1122 $ SLM pixels. As the SLM values are updated after backpropagation during training, the resulting PSF is different for each batch. Moreover, when simulating the PSF with the approach described in \Cref{sec:psf_modeling}, the downsampling factor is set to $ D = 8 $ (resolution of $380 \times 507$), as computing a full-scale PSF at each batch leads to much longer training times. With the compute hardware described in \Cref{sec:training_param}, it takes around \SI{3}{\minute} to simulate the entire MNIST dataset ($ 70'000 $ examples) for a downsampling factor of  $ D = 8 $, whereas it takes around \SI{250}{\minute} to simulate the same dataset at full resolution $ (3040 \times 4056) $.
\end{itemize}

As the PSF for \textit{Learned SLM} is of a lower dimension than the rest of the PSFs (downsampled by a factor of $ 8 $), when simulating each example in the dataset with the approach described in \Cref{sec:simulation}, we first downsample the other PSFs (except \textit{Lens}) by a factor of $ 8 $, such that the intensity PSF also has a resolution of $380 \times 507$. Note that this cannot be done for the \textit{Lens} PSF due to its very compact support; so we retain an intensity PSF $\bm{p} \in \mathbb{R}^{3040\times 4056 \times 3}$ when simulating \textit{Lens}' examples.

\subsection{Training details, hyperparameters, hardware, and network architectures}
\label{sec:training_param}

Experiments and classifiers in \Cref{sec:experiments} were run on a Dell Precision 5820 Tower X-Series (08B1) machine  with an Intel i9-10900X \SI{3.70}{\giga\hertz} CPU and two NVIDIA RTX A5000 GPUs. PyTorch~\cite{paszke2017automatic} was used for dataset preparation and training.

As the task consists of multi-label classification ($ 10 $ digits from $ 0 $ to $ 9 $), we use a cross entropy loss in optimizing \Cref{eq:mnist}:
\begin{equation}\label{eq:opt}
\mathcal{L} (y, \bm{\hat{p}}) = -\log \dfrac{\exp(\bm{\hat{p}}_{c=y})}{\sum_{c=0}^{9} \exp(\bm{\hat{p}}_c)},
\end{equation}
where $ y \in [0, 10)$ are the ground truth labels, and $ \bm{\hat{p}} =  D_{\bm{\theta}_D} \big( O_{\bm{\theta}_E}  ( \bm{x}) \big) \in \mathbb{R}^{10}$ are the predicted scores for a given input $ \bm{x} $ that passes through the optical encoder $ O_{\bm{\theta}_E} (\cdot) $ and the digital decoder $ D_{\bm{\theta}_D}  (\cdot) $.

All classifiers are trained for $ 50 $ epochs and with a batch size of $ N = 200 $. A large batch size is used to accelerate the training of \textit{Learned SLM}, i.e.\ to minimize the number of PSF updates per epoch and to parallelize FFT convolutions with the PSF. As the other approaches have a fixed optical encoder (and therefore fixed PSF), the FFT convolutions only need to be done once prior to training. When the two GPUs are used, it takes approximately $ \SI{15}{\minute} $ for the fixed-encoder classifiers to train and $ \SI{7.5}{\hour} $ for the end-to-end optical encoder digital classifier architecture to train. This significantly larger  training time for the end-to-end approach is because PSF simulation and dataset augmentation has to be done during training, while this can be pre-computed for the fixed encoders.

In the following sub-sections, we describe the two classifier architectures used in \Cref{sec:experiments}, namely the $ D_{\bm{\theta}_D} (\cdot) $ in \Cref{eq:mnist}. For the fixed optical encoders, the embeddings $ \{ \bm{v}_i \}_{i=1}^{N} $ that are inputted to the classifiers are pre-computed with the approach described in \Cref{sec:simulation}. The resulting augmented dataset is normalized (according to the augmented training set statistics). For \textit{Learned SLM}, we apply batch normalization~\cite{10.5555/3045118.3045167} and a ReLu activation to the sensor embedding prior to passing it to the classifier.
At inference, the parameters of batch normalization are fixed.

\subsubsection{Logistic regression for \Cref{sec:vary_dimension}}
\label{sec:logistic}

The classifier performs the following steps:
\begin{enumerate}
	\item Flatten sensor embedding. 
	\item Fully connected linear layer to $ 10 $ classes.
	\item Softmax decision layer.
\end{enumerate}

\subsubsection{Two-layer fully connected neural network for \Cref{sec:vary_dimension,sec:robustness}}
\label{sec:nn}

The classifier performs the following steps:
\begin{enumerate}
	\item Flatten sensor embedding. 
	\item Fully connected linear layer to hidden layer of $ 800 $ units, as in~\cite{1227801}
	\item Batch normalization.
	\item ReLu activation.
	\item Fully connected linear layer to $ 10 $ classes.
	\item Softmax decision layer.
\end{enumerate}

\subsection{Test accuracy curves for experiment on varying embedding dimension - \Cref{sec:vary_dimension}}
\label{sec:test_acc_dim}

\begin{figure}[h!]
	\centering
	\begin{subfigure}{.24\textwidth}
		\centering
		\includegraphics[width=0.99\linewidth]{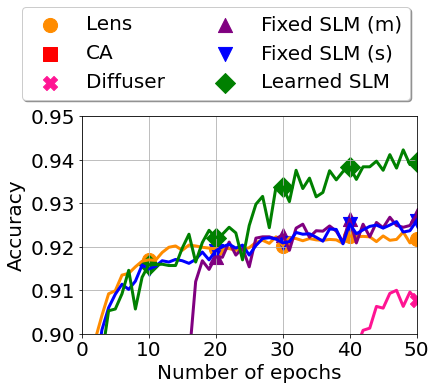}
		\caption{LR, 24$\times$32 = 768.}
		\label{fig:mnist_784}
	\end{subfigure}
	\hfill
	\begin{subfigure}{.24\textwidth}
		\centering
		\includegraphics[width=0.99\linewidth]{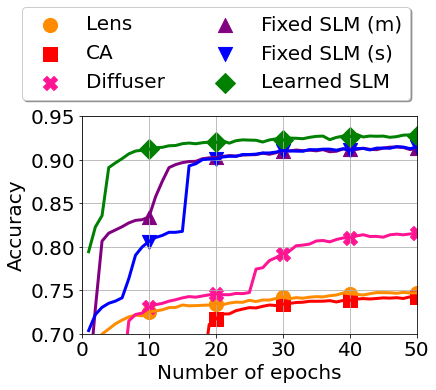}
		\caption{LR, 12$\times$16 = 192.}
		\label{fig:mnist_192}
	\end{subfigure}
	\hfill
	\begin{subfigure}{.24\textwidth}
		\centering
		\includegraphics[width=0.99\linewidth]{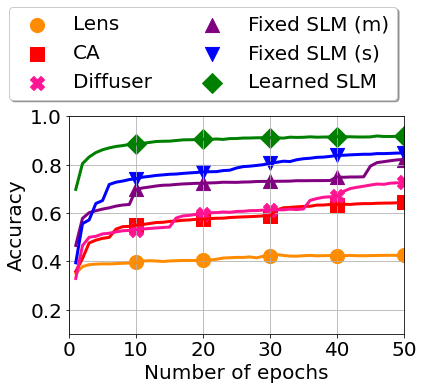}
		\caption{LR, 6$\times$8 = 48.}
		\label{fig:mnist_49}
	\end{subfigure}
	\hfill
	\begin{subfigure}{.24\textwidth}
		\centering
		\includegraphics[width=0.99\linewidth]{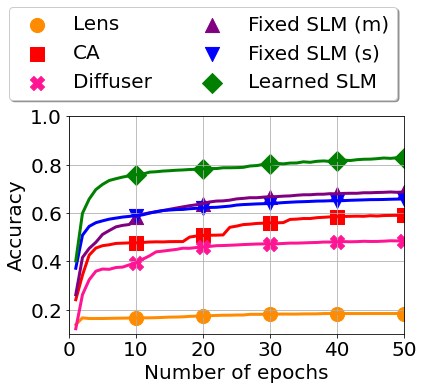}
		\caption{LR, 3$\times$4 = 12.}
		\label{fig:mnist_12}
	\end{subfigure}
	\\
	\begin{subfigure}{.24\textwidth}
		\centering
		\includegraphics[width=0.99\linewidth]{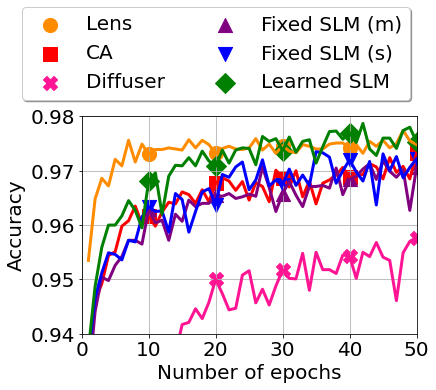}
		\caption{NN, 24$\times$32 = 768.}
		\label{fig:mnist_784_800}
	\end{subfigure}
	\hfill
	\begin{subfigure}{.24\textwidth}
		\centering
		\includegraphics[width=0.99\linewidth]{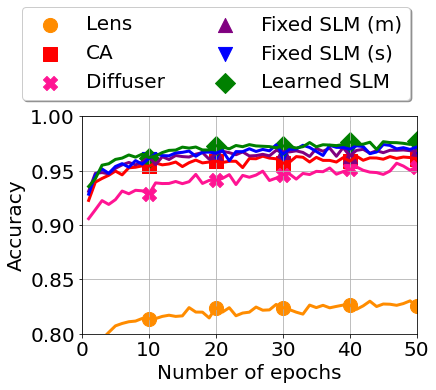}
		\caption{NN, 12$\times$16 = 192.}
		\label{fig:mnist_192_800}
	\end{subfigure}
	\hfill
	\begin{subfigure}{.24\textwidth}
		\centering
		\includegraphics[width=0.99\linewidth]{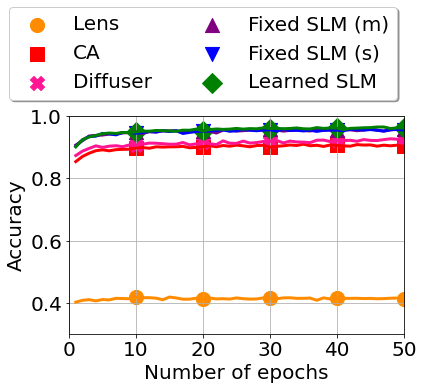}
		\caption{NN, 6$\times$8 = 48.}
		\label{fig:mnist_49_800}
	\end{subfigure}
	\hfill
	\begin{subfigure}{.24\textwidth}
		\centering
		\includegraphics[width=0.99\linewidth]{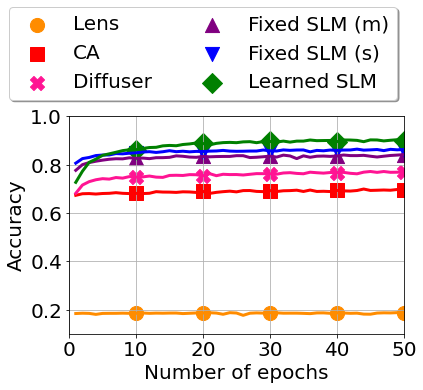}
		\caption{NN, 3$\times$4 = 12.}
		\label{fig:mnist_12_800}
	\end{subfigure}
	\caption{MNIST test accuracy curves while varying the sensor embedding dimension. Top row is for a logistic regression (LR) classifier, bottom row for a fully connected neural network (NN) with a single hidden layer of 800 units. The input dimension is indicated in the sub-figure caption.}
	\label{fig:mnist_vary_embedding}
\end{figure}

\newpage

\subsection{Example reconstructions of low-dimensional embeddings}
\label{sec:low_res_recon}

Although lensless measurements exhibit visual privacy in their raw measurements, with sufficient knowledge about the camera (e.g.\ a point spread function) and an appropriate computational algorithm, one is able to recover an estimate of the underlying object. In this section, we apply a common approach for recovering an image estimate from the raw measurements, namely solving the following inverse problem~\cite{Antipa:18}:
\begin{equation}\label{eq:lensless_inverse}
	\hat{\bm{x}} = \argmin_{\bm{x}\geq 0} \dfrac{1}{2}   \|  \bm{y} - \bm{H}\bm{x}\|_2^2 + \tau \| \Psi \bm{x}\|_1,
\end{equation}
where $ \hat{\bm{x}} $ is the image estimate, $ \bm{y} $ is the raw measurement, $ \bm{H} $ models the cropped convolution (aperture followed by optical encoder), $\bm{x}$ is the underlying image, and $ \bm{\Psi} $ maps $\bm{x}$ into a domain in which it is sparse. As MNIST data is sparse in pixels, $ \bm{\Psi} $ could be the identity matrix. However, our experiments found that the finite difference operator obtains better results.

\begin{figure}[h]
	\centering
	\begin{subfigure}{.19\textwidth}
		\centering
		\includegraphics[width=0.99\linewidth]{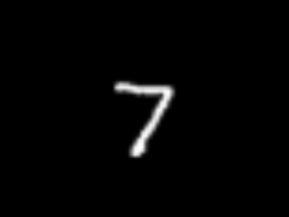}
		\caption{Lens.}
		\label{fig:lens_0_down4}
	\end{subfigure}
	\hfill
	\begin{subfigure}{.19\textwidth}
		\centering
		\includegraphics[width=0.99\linewidth]{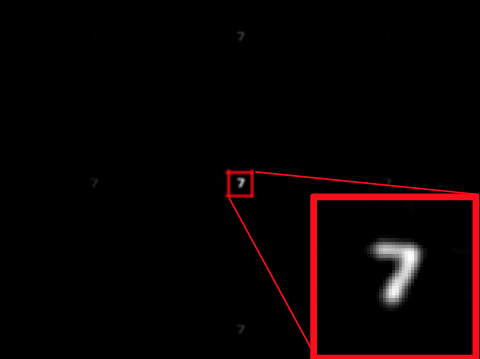}
		\caption{Coded aperture.}
		\label{fig:ca_0_down4}
	\end{subfigure}
	\hfill
	\begin{subfigure}{.19\textwidth}
		\centering
		\includegraphics[width=0.99\linewidth]{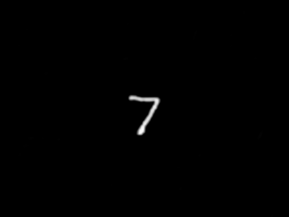}
		\caption{Diffuser.}
		\label{fig:diffuser_0_down4}
	\end{subfigure}
	\hfill
	\begin{subfigure}{.19\textwidth}
		\centering
		\includegraphics[width=0.99\linewidth]{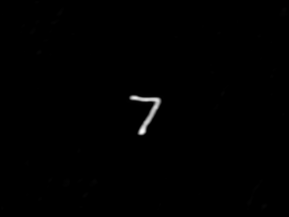}
		\caption{Fixed SLM (m).}
		\label{fig:adafruit_0_down4}
	\end{subfigure}
	\hfill
	\begin{subfigure}{.19\textwidth}
		\centering
		\includegraphics[width=0.99\linewidth]{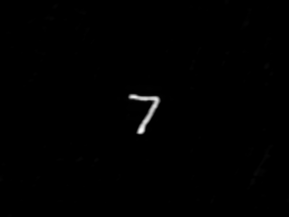}
		\caption{Fixed SLM (s).}
		\label{fig:adasim_0_down4}
	\end{subfigure}
	\caption{Example reconstruction for an embedding dimension of $ 760 \times 1014 $.}
	\label{fig:higher_res_reconstruction}
\end{figure}

Example reconstructions solving \Cref{eq:lensless_inverse} with 10 iterations of ADMM~\cite{admm} for the setup in \Cref{sec:vary_dimension} (object height of \SI{12}{\centi\meter}, \SI{40}{\centi\meter} from the camera) can be seen in \Cref{fig:higher_res_reconstruction} for an embedding dimension of $ 760 \times 1014 $ (downsampling Raspberry Pi HQ Camera resolution by $ 4 $). This represents a typical resolution for lensless imaging, where such imaging systems are able to recover an accurate estimate of the underlying image. Note that the height of the digit in \textit{Lens} is larger than the rest due to a larger mask-to-sensor. Conversely, \textit{Coded aperture} has a very small mask-to-sensor distance and therefore a very small object height at the sensor.

\begin{figure}[!htb]
	\begingroup
	\renewcommand{\arraystretch}{1} 
	\setlength{\tabcolsep}{0.2em} 
	\begin{tabular}{cccccc}
		\\
		& \includegraphics[width=0.16\linewidth,valign=m]{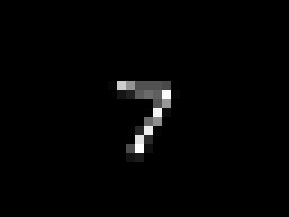} & \includegraphics[width=0.16\linewidth,valign=m]{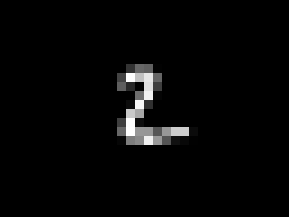}  &\includegraphics[width=0.16\linewidth,valign=m]{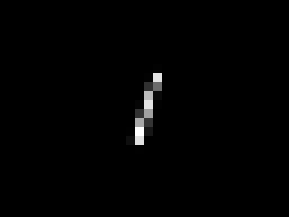} & \includegraphics[width=0.16\linewidth,valign=m]{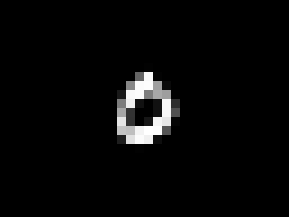} & \includegraphics[width=0.16\linewidth,valign=m]{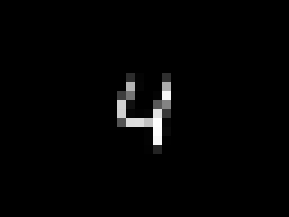}\\[30pt]
		\makecell{Coded \\aperture\\\cite{flatcam}}
		& \includegraphics[width=0.16\linewidth,valign=m]{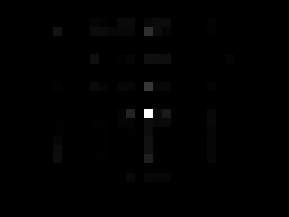} & \includegraphics[width=0.16\linewidth,valign=m]{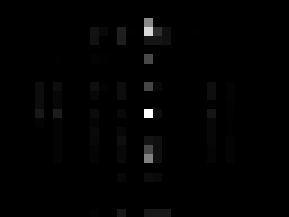}  &\includegraphics[width=0.16\linewidth,valign=m]{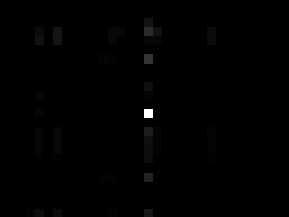} & \includegraphics[width=0.16\linewidth,valign=m]{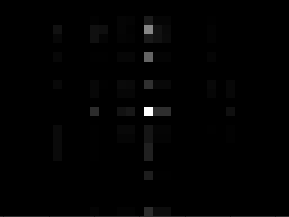} & \includegraphics[width=0.16\linewidth,valign=m]{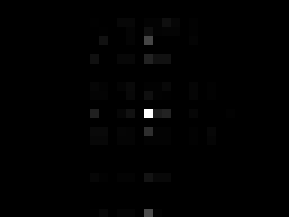}\\[30pt]
		\makecell{Diffuser~\cite{lenslesspicam}}
		& \includegraphics[width=0.16\linewidth,valign=m]{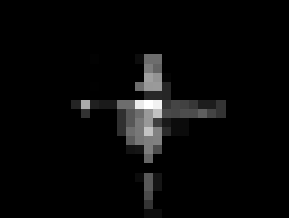} & \includegraphics[width=0.16\linewidth,valign=m]{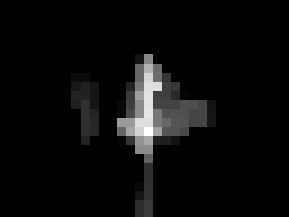}  &\includegraphics[width=0.16\linewidth,valign=m]{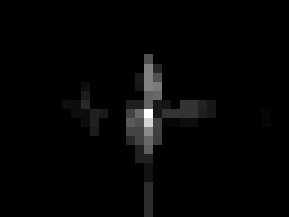} & \includegraphics[width=0.16\linewidth,valign=m]{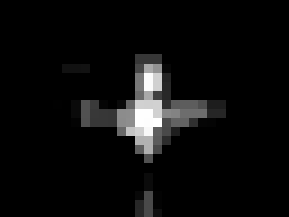} & \includegraphics[width=0.16\linewidth,valign=m]{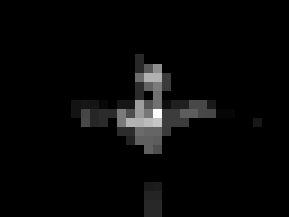}\\[30pt]
		\makecell{Fixed\\SLM\\(m)}
		& \includegraphics[width=0.16\linewidth,valign=m]{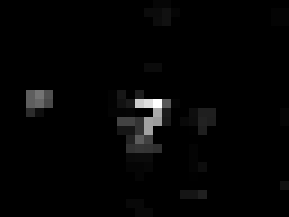}  
		& \includegraphics[width=0.16\linewidth,valign=m]{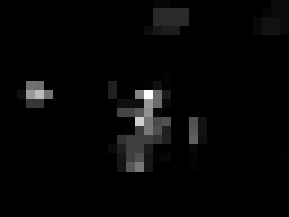}  &\includegraphics[width=0.16\linewidth,valign=m]{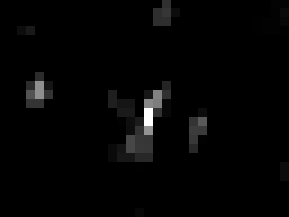} & \includegraphics[width=0.16\linewidth,valign=m]{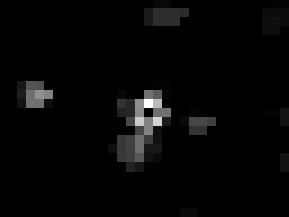} & \includegraphics[width=0.16\linewidth,valign=m]{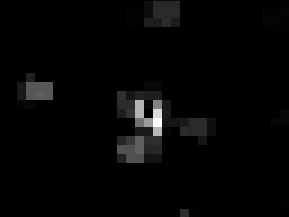}\\[30pt]
		\makecell{Fixed\\SLM\\(s)}
		& \includegraphics[width=0.16\linewidth,valign=m]{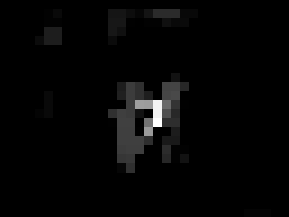}  
		& \includegraphics[width=0.16\linewidth,valign=m]{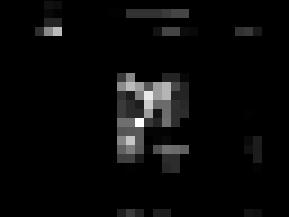}  &\includegraphics[width=0.16\linewidth,valign=m]{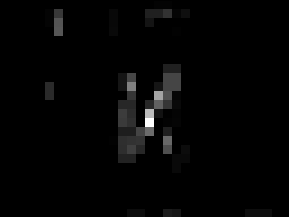} & \includegraphics[width=0.16\linewidth,valign=m]{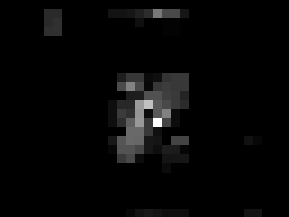} & \includegraphics[width=0.16\linewidth,valign=m]{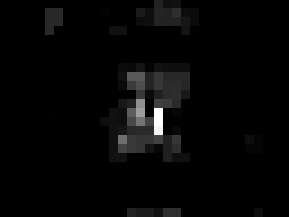}\\[30pt]
		\makecell{Learned\\SLM}
		& \includegraphics[width=0.16\linewidth,valign=m]{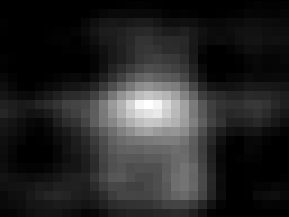}  
		& \includegraphics[width=0.16\linewidth,valign=m]{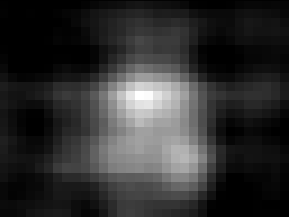}  &\includegraphics[width=0.16\linewidth,valign=m]{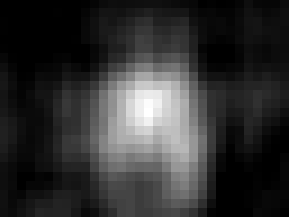} & \includegraphics[width=0.16\linewidth,valign=m]{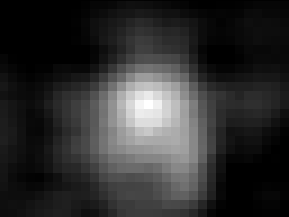} & \includegraphics[width=0.16\linewidth,valign=m]{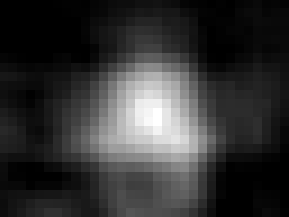}\\[30pt]
	\end{tabular}
	\endgroup
	\caption{Example reconstructions (except for \textit{Lens}) for an embedding dimension of $ 24\times 32  = 768$, which corresponds to a downsampling of around $ 126 $ along each dimension. Reconstructions are obtained using ADMM to solve \Cref{eq:lensless_inverse} for $ 100 $ iterations.
	}
	\label{fig:recon}
\end{figure}

For the reconstructions in \Cref{fig:recon} (except \textit{Lens} as the image is directly form on the sensor), we use a much lower dimensional sensor resolution, namely $ 24 \times 32 = 768 $ as in the experiments of \Cref{sec:vary_dimension} (downsampling Raspberry Pi HQ Camera resolution by around $ 126 $). \Cref{eq:lensless_inverse} is solved with 100 iterations of ADMM~\cite{admm}. \textit{Coded aperture} produces poor results as its mask-to-sensor distance is just \SI{0.5}{\milli\meter} as in the proposed design of~\cite{flatcam}. As a result, the object height at the sensor is $ \SI{12}{\centi\meter} \times  |M| = \SI{12}{\centi\meter}  \times (\SI{0.5}{\milli\meter} / \SI{40}{\centi\meter})  = \SI{0.15}{\milli\meter} $, which is about $ 100 $ pixels on the Raspberry Pi HQ sensor. As the sensor resolution is downsampled by around $ 126 $, the corresponding reconstruction (and underlying image) is contained within a single pixel. For the remaining lensless approaches, the reconstructed digit is of much poorer quality than downsampling by a factor of $ 4 $ (\Cref{fig:higher_res_reconstruction}).
Nonetheless, some features of digits can be distinguished, e.g.\ $ 7 $, $ 1 $, $ 0 $, and $ 4 $ for \textit{Fixed SLM (m)}.

\begin{figure}[!htb]
	\begingroup
	\renewcommand{\arraystretch}{1} 
	\setlength{\tabcolsep}{0.2em} 
	\begin{tabular}{cccccc}
		\\
		& \includegraphics[width=0.16\linewidth,valign=m]{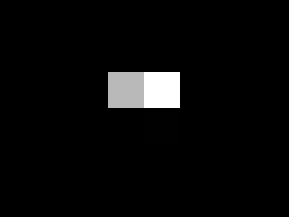} & \includegraphics[width=0.16\linewidth,valign=m]{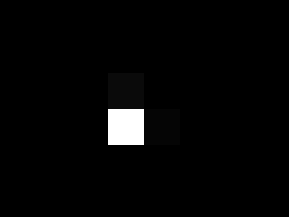}  &\includegraphics[width=0.16\linewidth,valign=m]{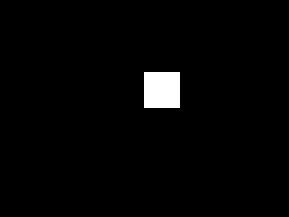} & \includegraphics[width=0.16\linewidth,valign=m]{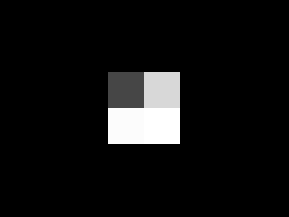} & \includegraphics[width=0.16\linewidth,valign=m]{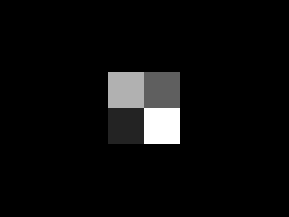}\\[30pt]
		\makecell{Coded \\aperture\\\cite{flatcam}}
		& \includegraphics[width=0.16\linewidth,valign=m]{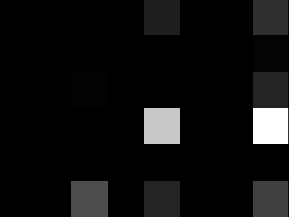} & \includegraphics[width=0.16\linewidth,valign=m]{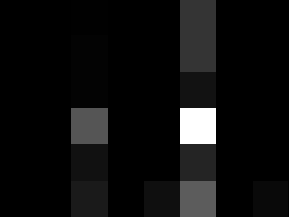}  &\includegraphics[width=0.16\linewidth,valign=m]{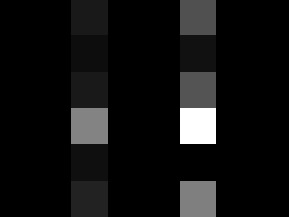} & \includegraphics[width=0.16\linewidth,valign=m]{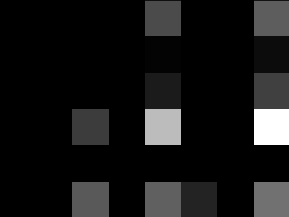} & \includegraphics[width=0.16\linewidth,valign=m]{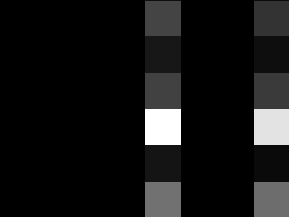}\\[30pt]
		\makecell{Diffuser~\cite{lenslesspicam}}
		& \includegraphics[width=0.16\linewidth,valign=m]{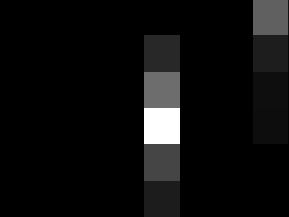} & \includegraphics[width=0.16\linewidth,valign=m]{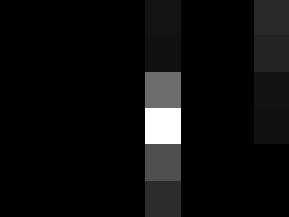}  &\includegraphics[width=0.16\linewidth,valign=m]{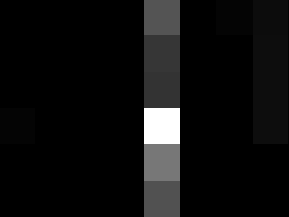} & \includegraphics[width=0.16\linewidth,valign=m]{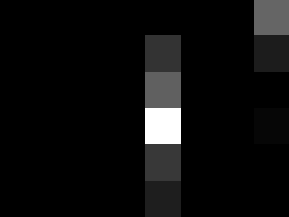} & \includegraphics[width=0.16\linewidth,valign=m]{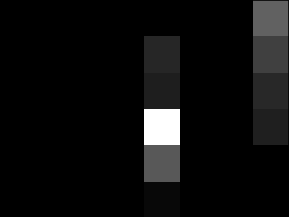}\\[30pt]
		\makecell{Fixed\\SLM\\(m)}
		& \includegraphics[width=0.16\linewidth,valign=m]{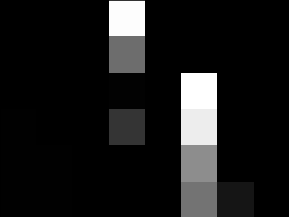}  
		& \includegraphics[width=0.16\linewidth,valign=m]{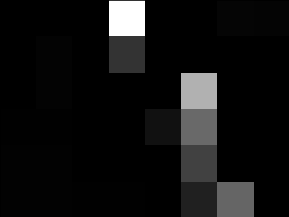}  &\includegraphics[width=0.16\linewidth,valign=m]{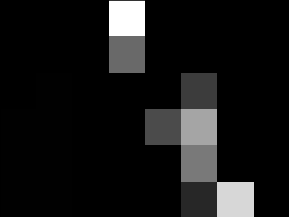} & \includegraphics[width=0.16\linewidth,valign=m]{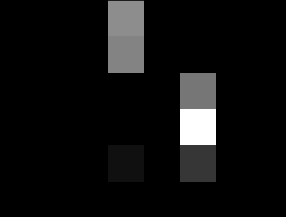} & \includegraphics[width=0.16\linewidth,valign=m]{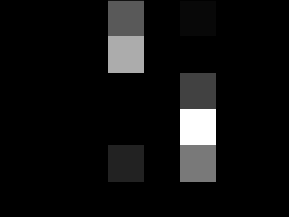}\\[30pt]
		\makecell{Fixed\\SLM\\(s)}
		& \includegraphics[width=0.16\linewidth,valign=m]{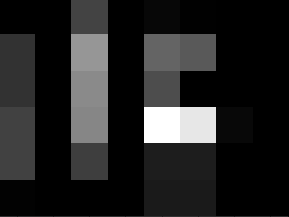}  
		& \includegraphics[width=0.16\linewidth,valign=m]{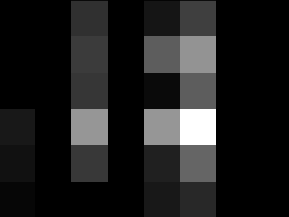}  &\includegraphics[width=0.16\linewidth,valign=m]{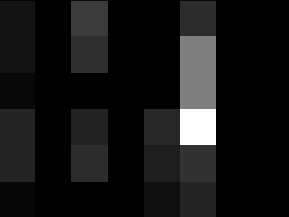} & \includegraphics[width=0.16\linewidth,valign=m]{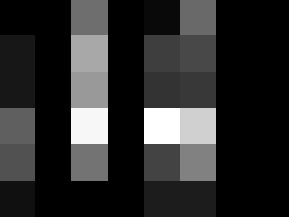} & \includegraphics[width=0.16\linewidth,valign=m]{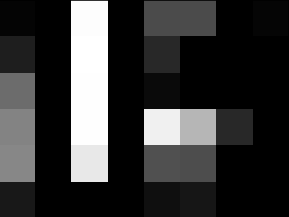}\\[30pt]
		\makecell{Learned\\SLM}
		& \includegraphics[width=0.16\linewidth,valign=m]{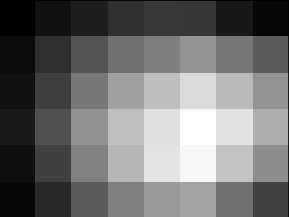}  
		& \includegraphics[width=0.16\linewidth,valign=m]{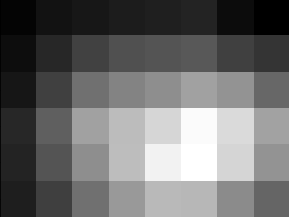}  &\includegraphics[width=0.16\linewidth,valign=m]{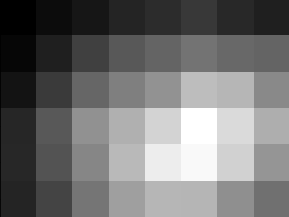} & \includegraphics[width=0.16\linewidth,valign=m]{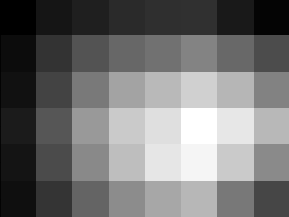} & \includegraphics[width=0.16\linewidth,valign=m]{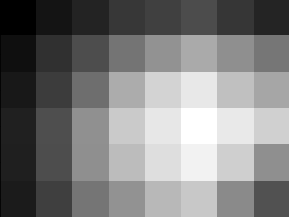}\\[30pt]
	\end{tabular}
	\endgroup
	\caption{Example reconstructions (except for \textit{Lens}) for an embedding dimension of $ 6\times 8  = 48$, which corresponds to a downsampling of around $ 507 $ along each dimension. Reconstructions are obtained using ADMM to solve \Cref{eq:lensless_inverse} for $ 100 $ iterations.
	}
	\label{fig:recon_lower}
\end{figure}

\Cref{fig:recon_lower} shows example reconstructions for an embedding dimension of $ 6 \times 8 = 48 $. For this sensor resolution, the measurements exhibit higher visual privacy as it is not possible to discern distinguishable features of digits from the images recovered by ADMM (nor from \textit{Lens}). Despite this inability to recover distinguishable features, all lensless approaches achieve above $ 90\%$ classification on the \textit{raw measurement}, see \Cref{tab:mnist_vary_embedding} for the two-layer neural network classifier.

\newpage

\newpage

\newpage

\subsection{Test accuracy curves for experiments on robustness to image transformations- \Cref{sec:robustness}}
\label{sec:test_acc_robust}

\begin{figure}[!htb]
	\centering
	\begin{subfigure}{.24\textwidth}
		\centering
		\includegraphics[width=0.99\linewidth]{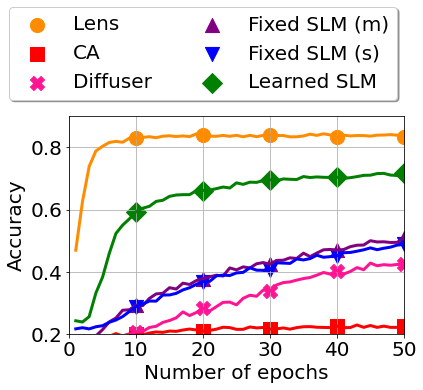}
		\caption{Shift.}
		\label{fig:transform_shift_768}
	\end{subfigure}
	\hfill
	\begin{subfigure}{.24\textwidth}
		\centering
		\includegraphics[width=0.99\linewidth]{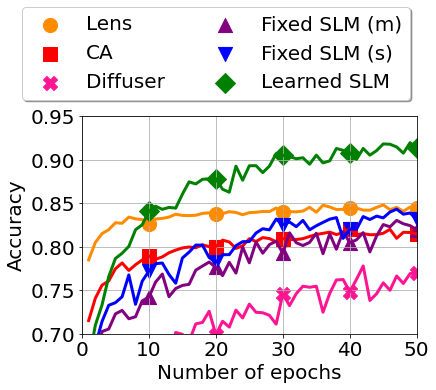}
		\caption{Rescale.}
		\label{fig:transform_rescale_768}
	\end{subfigure}
	\hfill
	\begin{subfigure}{.24\textwidth}
		\centering
		\includegraphics[width=0.99\linewidth]{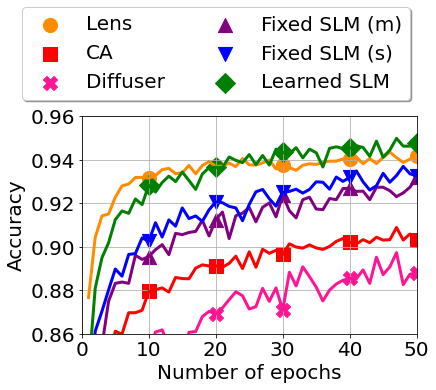}
		\caption{Rotate.}
		\label{fig:transform_rotate_768}
	\end{subfigure}
	\hfill
	\begin{subfigure}{.24\textwidth}
		\centering
		\includegraphics[width=0.99\linewidth]{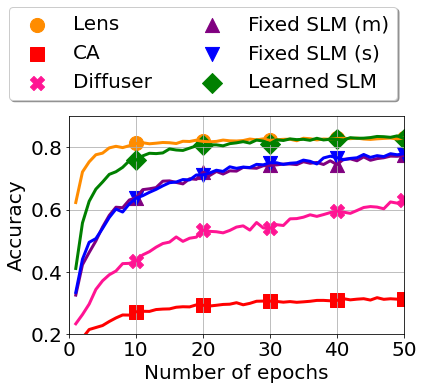}
		\caption{Perspective change.}
		\label{fig:transform_perspective_768}
	\end{subfigure}
	\\  
	\begin{subfigure}{.24\textwidth}
		\centering
		\includegraphics[width=0.99\linewidth]{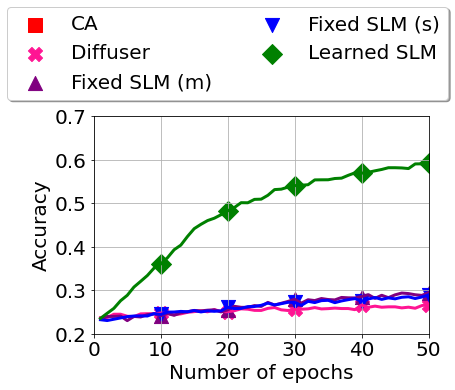}
		\caption{Shift.}
		\label{fig:transform_shift_48}
	\end{subfigure}
	\hfill
	\begin{subfigure}{.24\textwidth}
		\centering
		\includegraphics[width=0.99\linewidth]{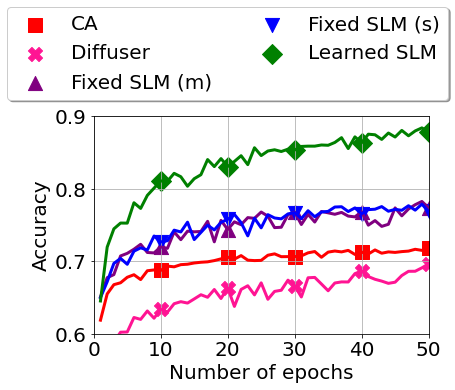}
		\caption{Rescale.}
		\label{fig:transform_rescale_48}
	\end{subfigure}
	\hfill
	\begin{subfigure}{.24\textwidth}
		\centering
		\includegraphics[width=0.99\linewidth]{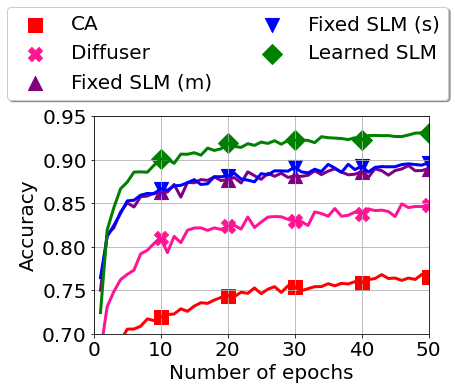}
		\caption{Rotate.}
		\label{fig:transform_rotate_48}
	\end{subfigure}
	\hfill
	\begin{subfigure}{.24\textwidth}
		\centering
		\includegraphics[width=0.99\linewidth]{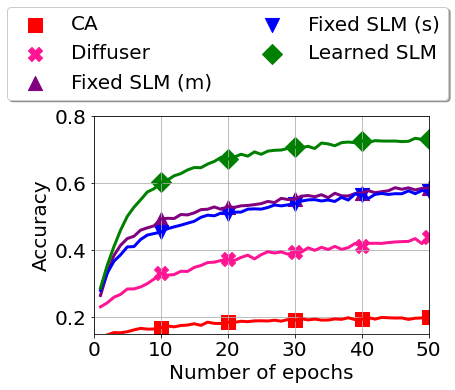}
		\caption{Perspective change.}
		\label{fig:transform_perspective_48}
	\end{subfigure}
	\caption{MNIST test accuracy curves for various image transformations. Top row is for an embedding dimension of 24$ \times $32 = 768, bottom row for an embedding dimension of 6$ \times $8 = 48. The classifier architecture is as described in \Cref{sec:nn}. The image transformation is indicated in the sub-figure caption.}
	\label{fig:mnist_transformation}
\end{figure}

\newpage

\subsection{PSFs of learned SLM masks}
\label{sec:learned_psfs}

The PSFs corresponding to the SLM masks determined from the end-to-end optimizations of \Cref{sec:vary_dimension} and \Cref{sec:robustness} can be found in this section.

\subsubsection{Varying embedding dimension experiment - \Cref{sec:vary_dimension}}
\label{sec:learned_psfs_vary}


\begin{figure}[h!]
	\centering
	\begin{subfigure}{.24\textwidth}
		\centering
		\includegraphics[width=0.99\linewidth]{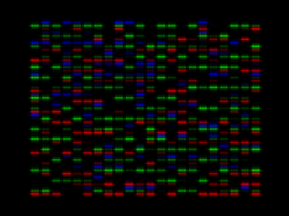}
		\caption{LR, 24$\times$32 = 768.}
		\label{fig:learned_psf_in768_logistic}
	\end{subfigure}
	\hfill
	\begin{subfigure}{.24\textwidth}
		\centering
		\includegraphics[width=0.99\linewidth]{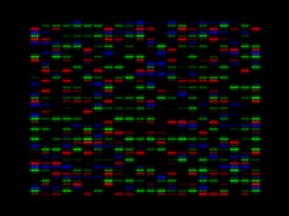}
		\caption{LR, 12$\times$16 = 192.}
		\label{fig:learned_psf_in192_logistic}
	\end{subfigure}
	\hfill
	\begin{subfigure}{.24\textwidth}
		\centering
		\includegraphics[width=0.99\linewidth]{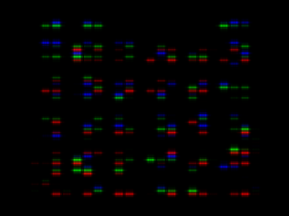}
		\caption{LR, 6$\times$8 = 48.}
		\label{fig:learned_psf_in48_logistic}
	\end{subfigure}
	\hfill
	\begin{subfigure}{.24\textwidth}
		\centering
		\includegraphics[width=0.99\linewidth]{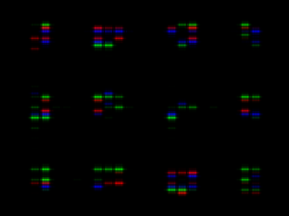}
		\caption{LR, 3$\times$4 = 12.}
		\label{fig:learned_psf_in12_logistic}
	\end{subfigure}
	\\ 
	\begin{subfigure}{.24\textwidth}
		\centering
		\includegraphics[width=0.99\linewidth]{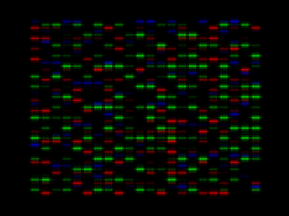}
		\caption{NN, 24$\times$32 = 768.}
		\label{fig:learned_psf_in768_singlehidden}
	\end{subfigure}
	\hfill
	\begin{subfigure}{.24\textwidth}
		\centering
		\includegraphics[width=0.99\linewidth]{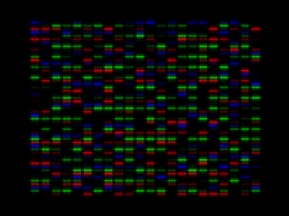}
		\caption{NN, 12$\times$16 = 192.}
		\label{fig:learned_psf_in192_singlehidden}
	\end{subfigure}
	\hfill
	\begin{subfigure}{.24\textwidth}
		\centering
		\includegraphics[width=0.99\linewidth]{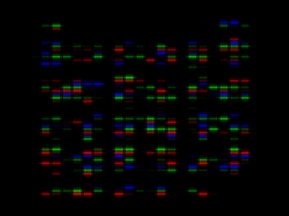}
		\caption{NN, 6$\times$8 = 48.}
		\label{fig:learned_psf_in48_singlehidden}
	\end{subfigure}
	\hfill
	\begin{subfigure}{.24\textwidth}
		\centering
		\includegraphics[width=0.99\linewidth]{figs/learned_psf_in12_singlehidden.png}
		\caption{NN, 3$\times$4 = 12.}
		\label{fig:learned_psf_in12_singlehidden}
	\end{subfigure}
	\caption{PSFs of learned masks. Top row is for a logistic regression (LR) classifier, bottom row for a fully connected neural network (NN) with a single hidden layer of 800 units. The input dimension is indicated in the sub-figure caption.}
	\label{fig:learned_masks_var}
\end{figure}

\subsubsection{Robustness to image transformations experiment - \Cref{sec:robustness}}

\begin{figure}[h!]
	\centering
	\begin{subfigure}{.24\textwidth}
		\centering
		\includegraphics[width=0.99\linewidth]{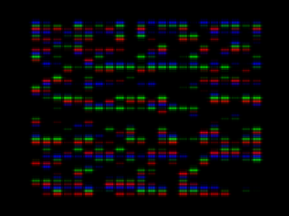}
		\caption{Shift, 24$\times$32.}
		\label{fig:learned_psf_768_shift}
	\end{subfigure}
	\hfill
	\begin{subfigure}{.24\textwidth}
		\centering
		\includegraphics[width=0.99\linewidth]{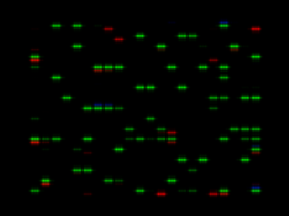}
		\caption{Rescale, 24$\times$32.}
		\label{fig:learned_psf_768_rescale}
	\end{subfigure}
	\hfill
	\begin{subfigure}{.24\textwidth}
		\centering
		\includegraphics[width=0.99\linewidth]{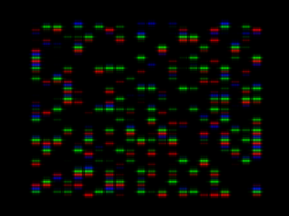}
		\caption{Rotate, 24$\times$32.}
		\label{fig:learned_psf_768_rotate}
	\end{subfigure}
	\hfill
	\begin{subfigure}{.24\textwidth}
		\centering
		\includegraphics[width=0.99\linewidth]{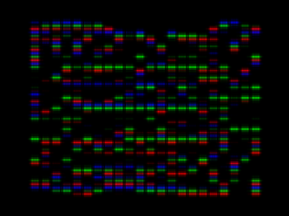}
		\caption{Perspective, 24$\times$32.}
		\label{fig:learned_psf_768_perspective}
	\end{subfigure}
	\\ 
	\begin{subfigure}{.24\textwidth}
		\centering
		\includegraphics[width=0.99\linewidth]{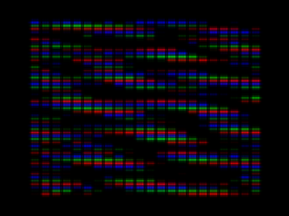}
		\caption{Shift, 6$\times$8.}
		\label{fig:learned_psf_48_shift}
	\end{subfigure}
	\hfill
	\begin{subfigure}{.24\textwidth}
		\centering
		\includegraphics[width=0.99\linewidth]{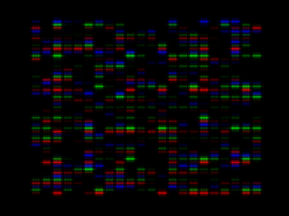}
		\caption{Rescale, 6$\times$8.}
		\label{fig:learned_psf_48_rescale}
	\end{subfigure}
	\hfill
	\begin{subfigure}{.24\textwidth}
		\centering
		\includegraphics[width=0.99\linewidth]{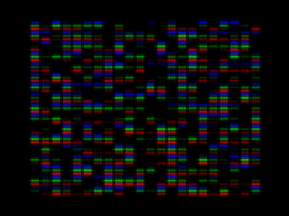}
		\caption{Rotate, 6$\times$8.}
		\label{fig:learned_psf_48_rotate}
	\end{subfigure}
	\hfill
	\begin{subfigure}{.24\textwidth}
		\centering
		\includegraphics[width=0.99\linewidth]{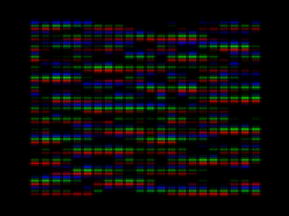}
		\caption{Perspective, 6$\times$8.}
		\label{fig:learned_psf_48_perspective}
	\end{subfigure}
	\caption{PSFs of learned masks for \Cref{sec:robustness} experiment. Top row is for an embedding dimension of 24$\times$32 = 768, while the bottom row if for an embedding dimension of 6$\times$8 = 48. The type of transformation is indicated in the sub-figure caption.}
	\label{fig:learned_masks_robust}
\end{figure}

\newpage

\subsection{Visualizing image transformation effects}
\label{sec:viz_transformation}

In this section, we visualize the various image transformations effects that are applied in the experiments of \Cref{sec:robustness}, namely the raw measurements and its corresponding reconstruction with \Cref{eq:lensless_inverse} for the lensless approaches. By comparing to \Cref{fig:recon}, we can see how the image transformations affect the capability of recovering the underlying image. Note that the raw measurements (and not the recovered images) are passed to classifier.

In what follows, we focus on the embeddings of \textit{Fixed SLM (m)} and \textit{Learned SLM}. For all transformations, we observe that it is difficult to recover discernible features from the recovered images of \textit{Learned SLM}'s measurements. Nonetheless, its raw measurements produce better classification results than the other (fixed) lensless encoders (see \Cref{tab:robustness}).

\subsubsection{Shift}
\label{sec:shift_effect}

\begin{figure}[h]
	\centering
	\begin{subfigure}{.19\textwidth}
		\centering
		\includegraphics[width=0.99\linewidth]{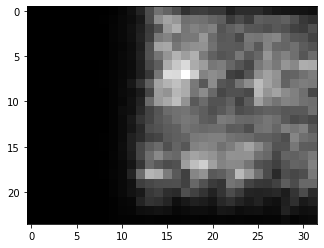}
		\caption{}
		\label{fig:ada0_768_shift}
	\end{subfigure}
	\hfill
	\begin{subfigure}{.19\textwidth}
		\centering
		\includegraphics[width=0.99\linewidth]{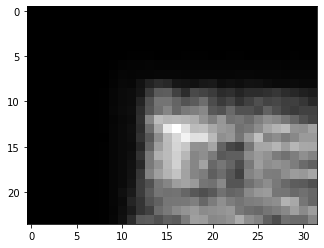}
		\caption{}
		\label{fig:ada1_768_shift}
	\end{subfigure}
	\hfill
	\begin{subfigure}{.19\textwidth}
		\centering
		\includegraphics[width=0.99\linewidth]{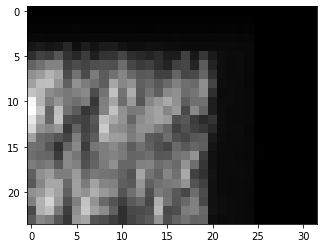}
		\caption{}
		\label{fig:ada2_768_shift}
	\end{subfigure}
	\hfill
	\begin{subfigure}{.19\textwidth}
		\centering
		\includegraphics[width=0.99\linewidth]{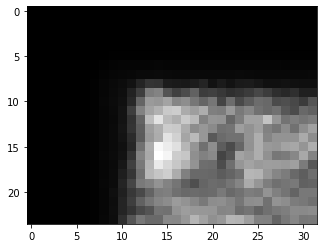}
		\caption{}
		\label{fig:ada3_768_shift}
	\end{subfigure}
	\hfill
	\begin{subfigure}{.19\textwidth}
		\centering
		\includegraphics[width=0.99\linewidth]{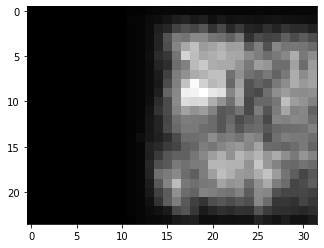}
		\caption{}
		\label{fig:ada4_768_shift}
	\end{subfigure}
	\begin{subfigure}{.19\textwidth}
		\centering
		\includegraphics[width=0.99\linewidth]{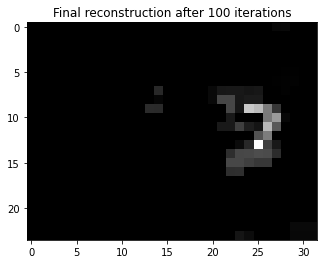}
		\caption{}
		\label{fig:ada0_768_shift_recon}
	\end{subfigure}
	\hfill
	\begin{subfigure}{.19\textwidth}
		\centering
		\includegraphics[width=0.99\linewidth]{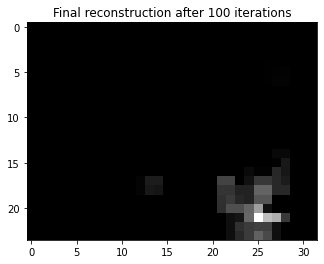}
		\caption{}
		\label{fig:ada1_768_shift_recon}
	\end{subfigure}
	\hfill
	\begin{subfigure}{.19\textwidth}
		\centering
		\includegraphics[width=0.99\linewidth]{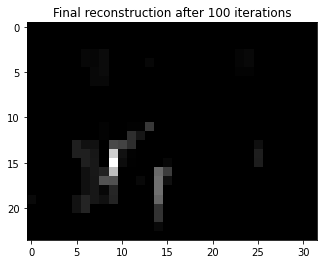}
		\caption{}
		\label{fig:ada2_768_shift_recon}
	\end{subfigure}
	\hfill
	\begin{subfigure}{.19\textwidth}
		\centering
		\includegraphics[width=0.99\linewidth]{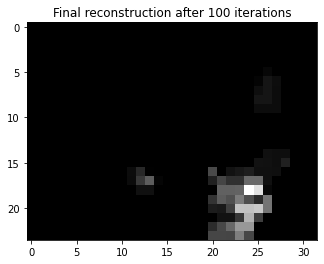}
		\caption{}
		\label{fig:ada3_768_shift_recon}
	\end{subfigure}
	\hfill
	\begin{subfigure}{.19\textwidth}
		\centering
		\includegraphics[width=0.99\linewidth]{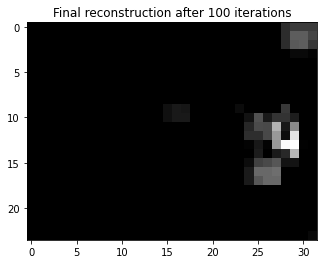}
		\caption{}
		\label{fig:ada4_768_shift_recon}
	\end{subfigure}
	\caption{Example reconstructions (100 iterations of ADMM to solve \Cref{eq:lensless_inverse}) for \textit{Fixed SLM (m)} in the presence of random shifts, for an embedding dimension of $ 24\times 32 $ (downsampling of around $ 126 $ along each dimension). (Top) raw measurements and (bottom) corresponding reconstruction.}
	\label{fig:ada_shift_768}
\end{figure}

\begin{figure}[h]
	\centering
	\begin{subfigure}{.19\textwidth}
		\centering
		\includegraphics[width=0.99\linewidth]{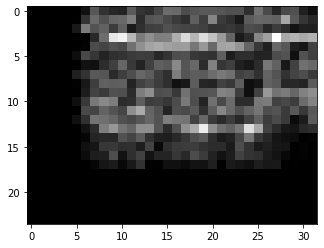}
		\caption{}
		\label{fig:learned0_768_shift}
	\end{subfigure}
	\hfill
	\begin{subfigure}{.19\textwidth}
		\centering
		\includegraphics[width=0.99\linewidth]{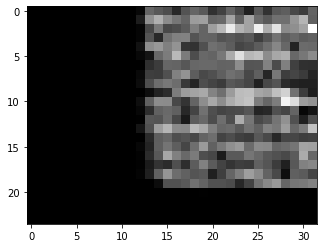}
		\caption{}
		\label{fig:learned1_768_shift}
	\end{subfigure}
	\hfill
	\begin{subfigure}{.19\textwidth}
		\centering
		\includegraphics[width=0.99\linewidth]{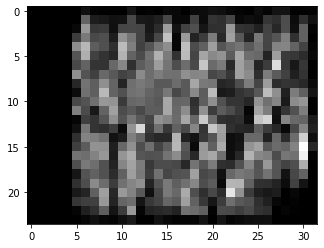}
		\caption{}
		\label{fig:learned2_768_shift}
	\end{subfigure}
	\hfill
	\begin{subfigure}{.19\textwidth}
		\centering
		\includegraphics[width=0.99\linewidth]{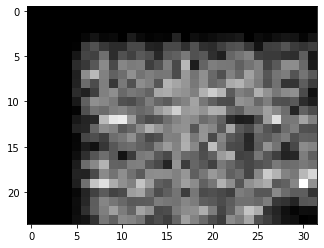}
		\caption{}
		\label{fig:learned3_768_shift}
	\end{subfigure}
	\hfill
	\begin{subfigure}{.19\textwidth}
		\centering
		\includegraphics[width=0.99\linewidth]{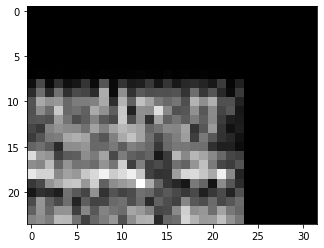}
		\caption{}
		\label{fig:learned4_768_shift}
	\end{subfigure}
	\begin{subfigure}{.19\textwidth}
		\centering
		\includegraphics[width=0.99\linewidth]{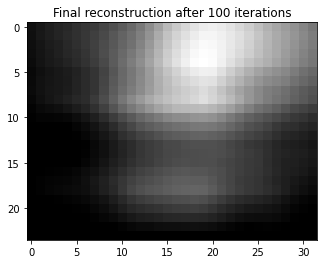}
		\caption{}
		\label{fig:learned0_768_shift_recon}
	\end{subfigure}
	\hfill
	\begin{subfigure}{.19\textwidth}
		\centering
		\includegraphics[width=0.99\linewidth]{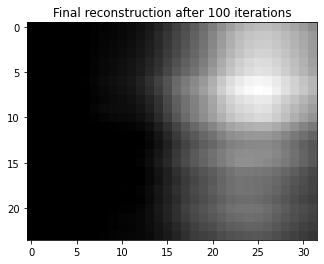}
		\caption{}
		\label{fig:learned1_768_shift_recon}
	\end{subfigure}
	\hfill
	\begin{subfigure}{.19\textwidth}
		\centering
		\includegraphics[width=0.99\linewidth]{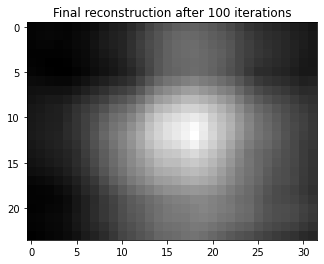}
		\caption{}
		\label{fig:learned2_768_shift_recon}
	\end{subfigure}
	\hfill
	\begin{subfigure}{.19\textwidth}
		\centering
		\includegraphics[width=0.99\linewidth]{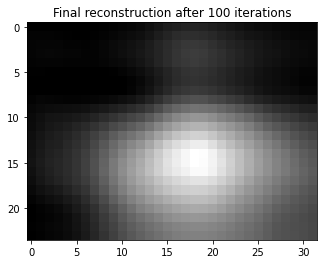}
		\caption{}
		\label{fig:learned3_768_shift_recon}
	\end{subfigure}
	\hfill
	\begin{subfigure}{.19\textwidth}
		\centering
		\includegraphics[width=0.99\linewidth]{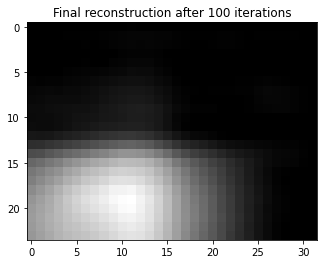}
		\caption{}
		\label{fig:learned4_768_shift_recon}
	\end{subfigure}
	\caption{Example reconstructions (100 iterations of ADMM to solve \Cref{eq:lensless_inverse}) for \textit{Learned SLM} in the presence of random shifts, for an embedding dimension of $ 24\times 32 $ (downsampling of around $ 126 $ along each dimension). (Top) raw measurements and (bottom) corresponding reconstruction.}
	\label{fig:learned_shift_768}
\end{figure}

\Cref{fig:ada_shift_768,fig:learned_shift_768} show the raw embeddings and reconstructed outputs for \textit{Fixed SLM (m)} and \textit{Learned SLM} under random shifts. We can see how shifting the object at the scene plane results in a shift in the raw measurement. As a result, the sensor loses multiplexed information with respect to objects that are centered, see \Cref{tab:mnist_examples}. This loss of multiplexed information may explain why there is a sudden drop in classification performance for \textit{Shift} in \Cref{tab:performance_drop_trans} for lensless approaches.

\begin{figure}[h]
	\centering
	\begin{subfigure}{.19\textwidth}
		\centering
		\includegraphics[width=0.99\linewidth]{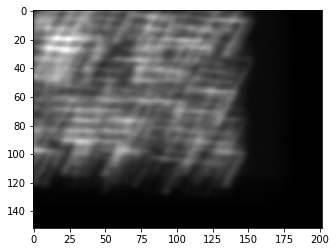}
		\caption{}
		\label{fig:ada0_down20_shift}
	\end{subfigure}
	\hfill
	\begin{subfigure}{.19\textwidth}
		\centering
		\includegraphics[width=0.99\linewidth]{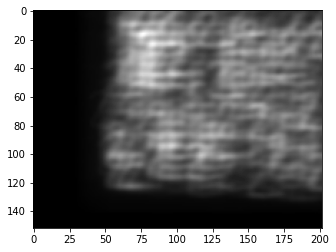}
		\caption{}
		\label{fig:ada1_down20_shift}
	\end{subfigure}
	\hfill
	\begin{subfigure}{.19\textwidth}
		\centering
		\includegraphics[width=0.99\linewidth]{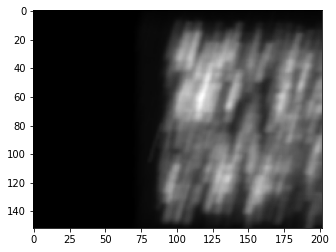}
		\caption{}
		\label{fig:ada2_down20_shift}
	\end{subfigure}
	\hfill
	\begin{subfigure}{.19\textwidth}
		\centering
		\includegraphics[width=0.99\linewidth]{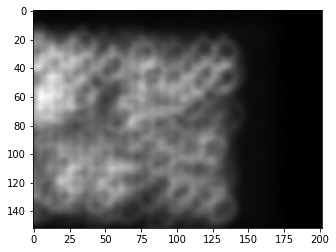}
		\caption{}
		\label{fig:ada3_down20_shift}
	\end{subfigure}
	\hfill
	\begin{subfigure}{.19\textwidth}
		\centering
		\includegraphics[width=0.99\linewidth]{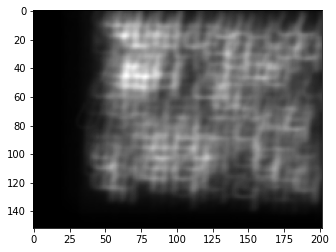}
		\caption{}
		\label{fig:ada4_down20_shift}
	\end{subfigure}
	\begin{subfigure}{.19\textwidth}
		\centering
		\includegraphics[width=0.99\linewidth]{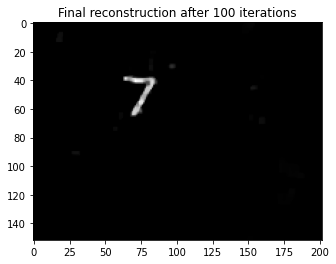}
		\caption{}
		\label{fig:ada0_down20_shift_recon}
	\end{subfigure}
	\hfill
	\begin{subfigure}{.19\textwidth}
		\centering
		\includegraphics[width=0.99\linewidth]{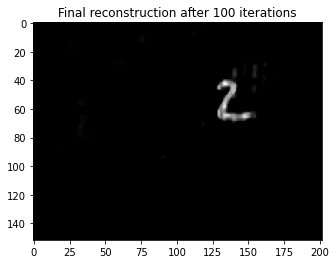}
		\caption{}
		\label{fig:ada1_down20_shift_recon}
	\end{subfigure}
	\hfill
	\begin{subfigure}{.19\textwidth}
		\centering
		\includegraphics[width=0.99\linewidth]{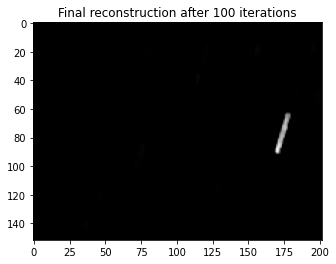}
		\caption{}
		\label{fig:ada2_down20_shift_recon}
	\end{subfigure}
	\hfill
	\begin{subfigure}{.19\textwidth}
		\centering
		\includegraphics[width=0.99\linewidth]{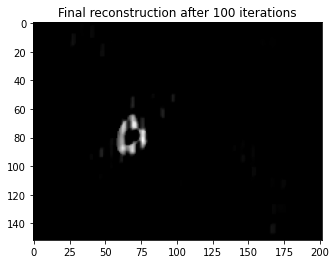}
		\caption{}
		\label{fig:ada3_down20_shift_recon}
	\end{subfigure}
	\hfill
	\begin{subfigure}{.19\textwidth}
		\centering
		\includegraphics[width=0.99\linewidth]{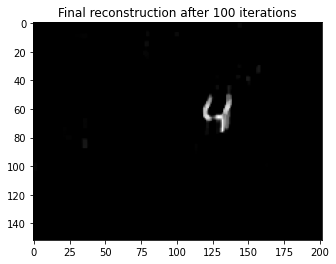}
		\caption{}
		\label{fig:ada4_down20_shift_recon}
	\end{subfigure}
	\caption{Example reconstructions (100 iterations of ADMM to solve \Cref{eq:lensless_inverse}) for \textit{Fixed SLM (m)} in the presence of random shifts, for an embedding dimension of $ 152\times 202 $ (downsampling of around $ 20 $ along each dimension). (Top) raw measurements and (bottom) corresponding reconstruction.}
	\label{fig:ada_shift_down20}
\end{figure}

\Cref{fig:ada_shift_down20} shows that for \textit{Fixed SLM (m)} with a larger sensor resolution, and hence more multiplexed information, a faithful image of the scene can be recovered under random shifts.

\newpage

\subsubsection{Rescale}

\begin{figure}[h]
	\centering
	\begin{subfigure}{.19\textwidth}
		\centering
		\includegraphics[width=0.99\linewidth]{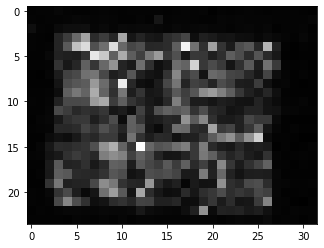}
		\caption{}
		\label{fig:ada0_768_2cm}
	\end{subfigure}
	\hfill
	\begin{subfigure}{.19\textwidth}
		\centering
		\includegraphics[width=0.99\linewidth]{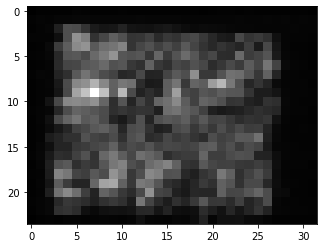}
		\caption{}
		\label{fig:ada0_768_6.5cm}
	\end{subfigure}
	\hfill
	\begin{subfigure}{.19\textwidth}
		\centering
		\includegraphics[width=0.99\linewidth]{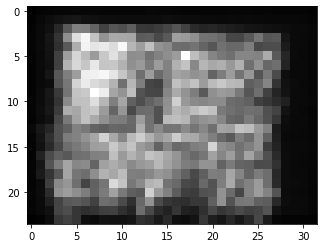}
		\caption{}
		\label{fig:ada0_768_11cm}
	\end{subfigure}
	\hfill
	\begin{subfigure}{.19\textwidth}
		\centering
		\includegraphics[width=0.99\linewidth]{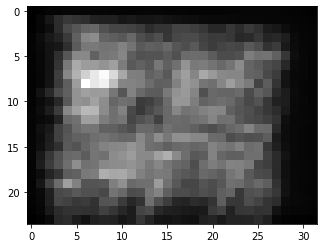}
		\caption{}
		\label{fig:ada0_768_15.5cm}
	\end{subfigure}
	\hfill
	\begin{subfigure}{.19\textwidth}
		\centering
		\includegraphics[width=0.99\linewidth]{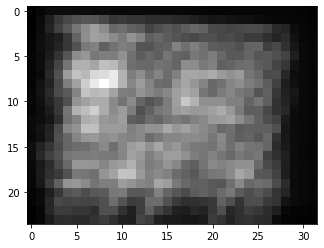}
		\caption{}
		\label{fig:ada0_768_20cm}
	\end{subfigure}
	\begin{subfigure}{.19\textwidth}
		\centering
		\includegraphics[width=0.99\linewidth]{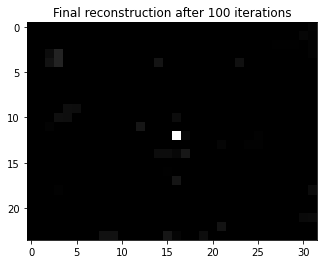}
		\caption{}
		\label{fig:ada0_768_2cm_recon}
	\end{subfigure}
	\hfill
	\begin{subfigure}{.19\textwidth}
		\centering
		\includegraphics[width=0.99\linewidth]{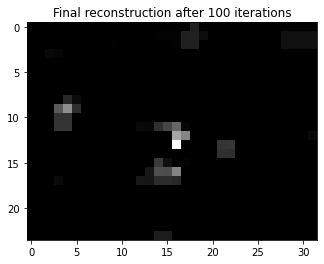}
		\caption{}
		\label{fig:ada0_768_6p5cm_recon}
	\end{subfigure}
	\hfill
	\begin{subfigure}{.19\textwidth}
		\centering
		\includegraphics[width=0.99\linewidth]{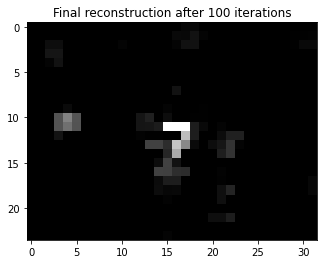}
		\caption{}
		\label{fig:ada0_768_11cm_recon}
	\end{subfigure}
	\hfill
	\begin{subfigure}{.19\textwidth}
		\centering
		\includegraphics[width=0.99\linewidth]{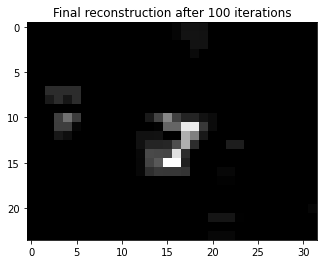}
		\caption{}
		\label{fig:ada0_768_15p5cm_recon}
	\end{subfigure}
	\hfill
	\begin{subfigure}{.19\textwidth}
		\centering
		\includegraphics[width=0.99\linewidth]{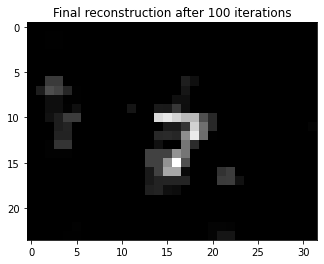}
		\caption{}
		\label{fig:ada0_768_20cm_recon}
	\end{subfigure}
	\caption{Example reconstructions (100 iterations of ADMM to solve \Cref{eq:lensless_inverse}) for \textit{Fixed SLM (m)}, for an embedding dimension of $ 24\times 32 $. The object height increases from left to right: \SI{2}{\centi\meter}, \SI{6.5}{\centi\meter}, \SI{11}{\centi\meter}, \SI{15.5}{\centi\meter}, \SI{20}{\centi\meter}. (Top) raw measurements and (bottom) corresponding reconstruction.}
	\label{fig:ada_rescale_768}
\end{figure}

\begin{figure}[h]
	\centering
	\begin{subfigure}{.19\textwidth}
		\centering
		\includegraphics[width=0.99\linewidth]{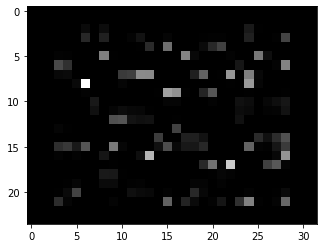}
		\caption{}
		\label{fig:learned0_768_2cm}
	\end{subfigure}
	\hfill
	\begin{subfigure}{.19\textwidth}
		\centering
		\includegraphics[width=0.99\linewidth]{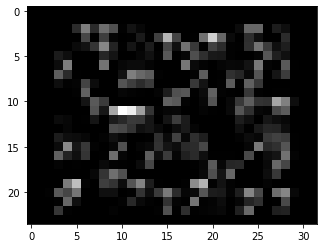}
		\caption{}
		\label{fig:learned0_768_6p5cm}
	\end{subfigure}
	\hfill
	\begin{subfigure}{.19\textwidth}
		\centering
		\includegraphics[width=0.99\linewidth]{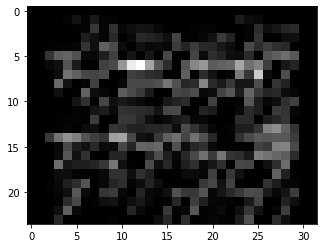}
		\caption{}
		\label{fig:learned0_768_11cm}
	\end{subfigure}
	\hfill
	\begin{subfigure}{.19\textwidth}
		\centering
		\includegraphics[width=0.99\linewidth]{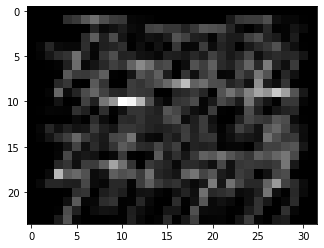}
		\caption{}
		\label{fig:learned0_768_15p5cm}
	\end{subfigure}
	\hfill
	\begin{subfigure}{.19\textwidth}
		\centering
		\includegraphics[width=0.99\linewidth]{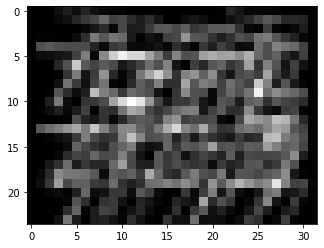}
		\caption{}
		\label{fig:learned0_768_20cm}
	\end{subfigure}
	\begin{subfigure}{.19\textwidth}
		\centering
		\includegraphics[width=0.99\linewidth]{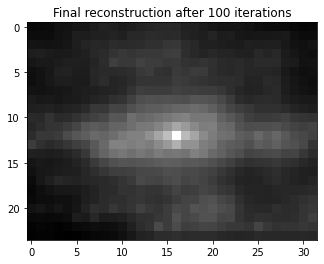}
		\caption{}
		\label{fig:learned0_768_2cm_recon}
	\end{subfigure}
	\hfill
	\begin{subfigure}{.19\textwidth}
		\centering
		\includegraphics[width=0.99\linewidth]{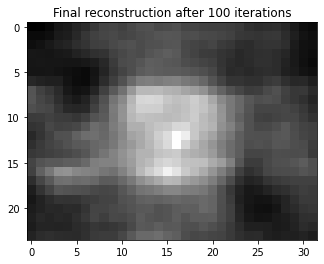}
		\caption{}
		\label{fig:learned0_768_6p5cm_recon}
	\end{subfigure}
	\hfill
	\begin{subfigure}{.19\textwidth}
		\centering
		\includegraphics[width=0.99\linewidth]{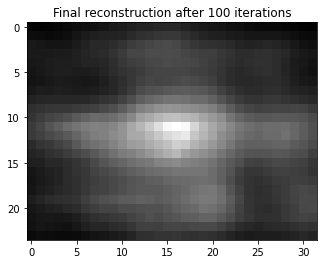}
		\caption{}
		\label{fig:learned0_768_11cm_recon}
	\end{subfigure}
	\hfill
	\begin{subfigure}{.19\textwidth}
		\centering
		\includegraphics[width=0.99\linewidth]{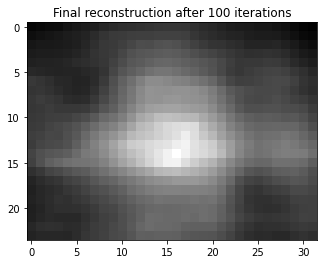}
		\caption{}
		\label{fig:learned0_768_15p5cm_recon}
	\end{subfigure}
	\hfill
	\begin{subfigure}{.19\textwidth}
		\centering
		\includegraphics[width=0.99\linewidth]{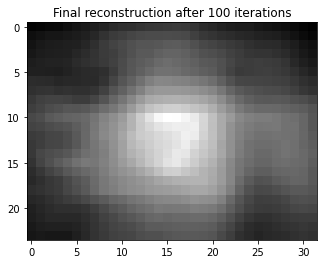}
		\caption{}
		\label{fig:learned0_768_20cm_recon}
	\end{subfigure}
	\caption{Example reconstructions (100 iterations of ADMM to solve \Cref{eq:lensless_inverse}) for \textit{Learned SLM}, for an embedding dimension of $ 24\times 32 $. The object height increases from left to right: \SI{2}{\centi\meter}, \SI{6.5}{\centi\meter}, \SI{11}{\centi\meter}, \SI{15.5}{\centi\meter}, \SI{20}{\centi\meter}. (Top) raw measurements and (bottom) corresponding reconstruction.}
	\label{fig:learned_rescale_768}
\end{figure}

\Cref{fig:ada_rescale_768,fig:learned_rescale_768} show the effect of rescaling on raw embeddings and reconstructed outputs of \textit{Fixed SLM (m)} and \textit{Learned SLM}.

\newpage

\subsubsection{Rotate}

\begin{figure}[h]
	\centering
	\begin{subfigure}{.19\textwidth}
		\centering
		\includegraphics[width=0.99\linewidth]{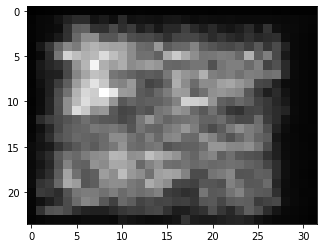}
		\caption{}
		\label{fig:ada0_768_rotate}
	\end{subfigure}
	\hfill
	\begin{subfigure}{.19\textwidth}
		\centering
		\includegraphics[width=0.99\linewidth]{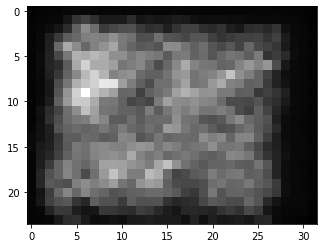}
		\caption{}
		\label{fig:ada1_768_rotate}
	\end{subfigure}
	\hfill
	\begin{subfigure}{.19\textwidth}
		\centering
		\includegraphics[width=0.99\linewidth]{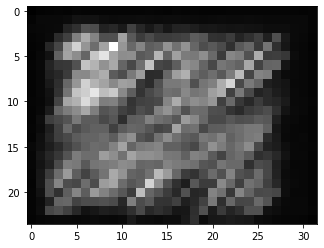}
		\caption{}
		\label{fig:ada2_768_rotate}
	\end{subfigure}
	\hfill
	\begin{subfigure}{.19\textwidth}
		\centering
		\includegraphics[width=0.99\linewidth]{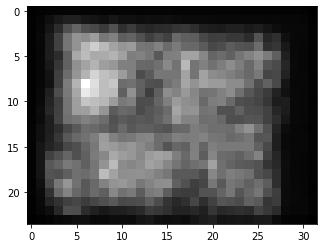}
		\caption{}
		\label{fig:ada3_768_rotate}
	\end{subfigure}
	\hfill
	\begin{subfigure}{.19\textwidth}
		\centering
		\includegraphics[width=0.99\linewidth]{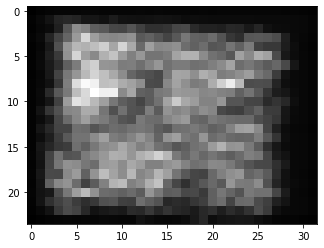}
		\caption{}
		\label{fig:ada4_768_rotate}
	\end{subfigure}
	\begin{subfigure}{.19\textwidth}
		\centering
		\includegraphics[width=0.99\linewidth]{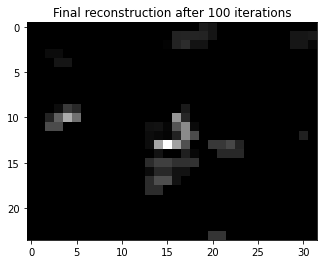}
		\caption{}
		\label{fig:ada0_768_rotate_recon}
	\end{subfigure}
	\hfill
	\begin{subfigure}{.19\textwidth}
		\centering
		\includegraphics[width=0.99\linewidth]{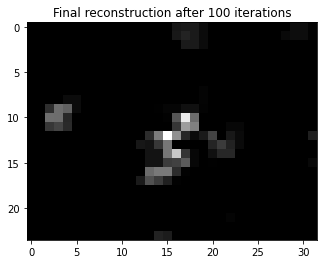}
		\caption{}
		\label{fig:ada1_768_rotate_recon}
	\end{subfigure}
	\hfill
	\begin{subfigure}{.19\textwidth}
		\centering
		\includegraphics[width=0.99\linewidth]{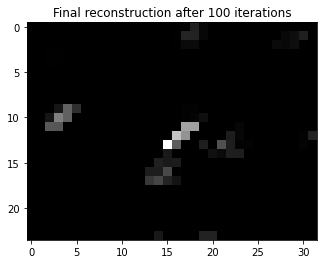}
		\caption{}
		\label{fig:ada2_768_rotate_recon}
	\end{subfigure}
	\hfill
	\begin{subfigure}{.19\textwidth}
		\centering
		\includegraphics[width=0.99\linewidth]{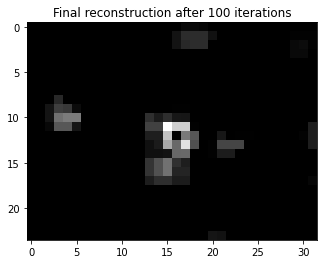}
		\caption{}
		\label{fig:ada3_768_rotate_recon}
	\end{subfigure}
	\hfill
	\begin{subfigure}{.19\textwidth}
		\centering
		\includegraphics[width=0.99\linewidth]{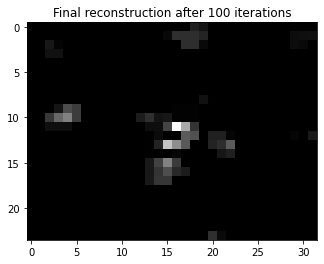}
		\caption{}
		\label{fig:ada4_768_rotate_recon}
	\end{subfigure}
	\caption{Example reconstructions (100 iterations of ADMM to solve \Cref{eq:lensless_inverse}) for \textit{Fixed SLM (m)} in the presence of random rotations, for an embedding dimension of $ 24\times 32 $. (Top) raw measurements and (bottom) corresponding reconstruction.}
	\label{fig:ada_rotate_768}
\end{figure}

\begin{figure}[h]
	\centering
	\begin{subfigure}{.19\textwidth}
		\centering
		\includegraphics[width=0.99\linewidth]{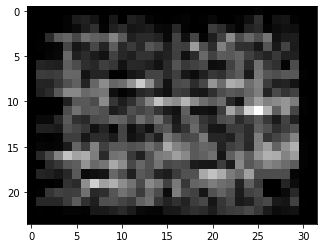}
		\caption{}
		\label{fig:learned0_768_rotate}
	\end{subfigure}
	\hfill
	\begin{subfigure}{.19\textwidth}
		\centering
		\includegraphics[width=0.99\linewidth]{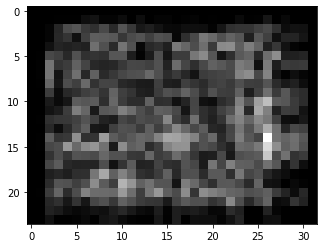}
		\caption{}
		\label{fig:learned1_768_rotate}
	\end{subfigure}
	\hfill
	\begin{subfigure}{.19\textwidth}
		\centering
		\includegraphics[width=0.99\linewidth]{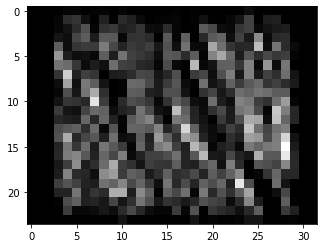}
		\caption{}
		\label{fig:learned2_768_rotate}
	\end{subfigure}
	\hfill
	\begin{subfigure}{.19\textwidth}
		\centering
		\includegraphics[width=0.99\linewidth]{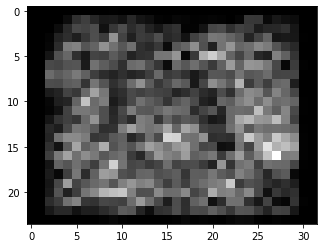}
		\caption{}
		\label{fig:learned3_768_rotate}
	\end{subfigure}
	\hfill
	\begin{subfigure}{.19\textwidth}
		\centering
		\includegraphics[width=0.99\linewidth]{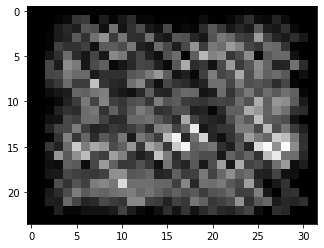}
		\caption{}
		\label{fig:learned4_768_rotate}
	\end{subfigure}
	\begin{subfigure}{.19\textwidth}
		\centering
		\includegraphics[width=0.99\linewidth]{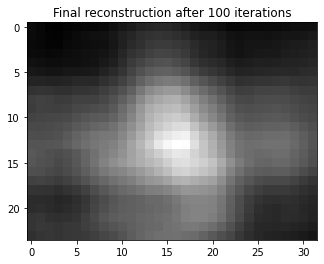}
		\caption{}
		\label{fig:learned0_768_rotate_recon}
	\end{subfigure}
	\hfill
	\begin{subfigure}{.19\textwidth}
		\centering
		\includegraphics[width=0.99\linewidth]{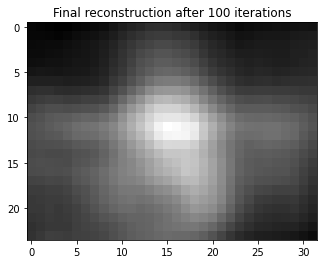}
		\caption{}
		\label{fig:learned1_768_rotate_recon}
	\end{subfigure}
	\hfill
	\begin{subfigure}{.19\textwidth}
		\centering
		\includegraphics[width=0.99\linewidth]{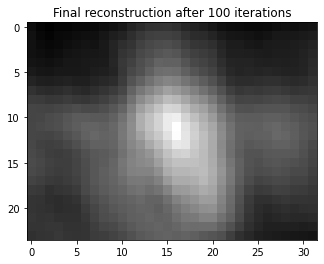}
		\caption{}
		\label{fig:learned2_768_rotate_recon}
	\end{subfigure}
	\hfill
	\begin{subfigure}{.19\textwidth}
		\centering
		\includegraphics[width=0.99\linewidth]{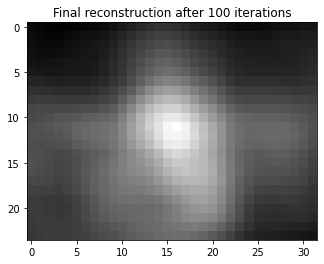}
		\caption{}
		\label{fig:learned3_768_rotate_recon}
	\end{subfigure}
	\hfill
	\begin{subfigure}{.19\textwidth}
		\centering
		\includegraphics[width=0.99\linewidth]{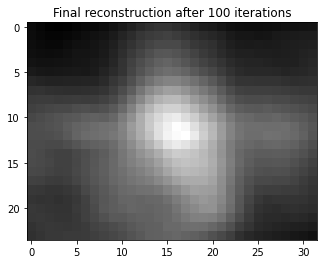}
		\caption{}
		\label{fig:learned4_768_rotate_recon}
	\end{subfigure}
	\caption{Example reconstructions (100 iterations of ADMM to solve \Cref{eq:lensless_inverse}) for \textit{Learned SLM} in the presence of random rotations, for an embedding dimension of $ 24\times 32 $. (Top) raw measurements and (bottom) corresponding reconstruction.}
	\label{fig:learned_rotate_768}
\end{figure}

\Cref{fig:ada_rotate_768,fig:learned_rotate_768} show the effect of rotation on raw embeddings and reconstructed outputs of \textit{Fixed SLM (m)} and \textit{Learned SLM}.

\newpage

\subsubsection{Perspective}

\begin{figure}[h]
	\centering
	\begin{subfigure}{.19\textwidth}
		\centering
		\includegraphics[width=0.99\linewidth]{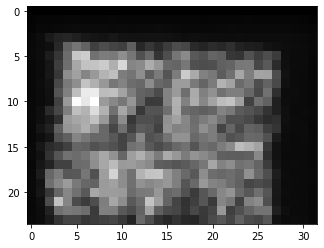}
		\caption{}
		\label{fig:ada0_768_pers}
	\end{subfigure}
	\hfill
	\begin{subfigure}{.19\textwidth}
		\centering
		\includegraphics[width=0.99\linewidth]{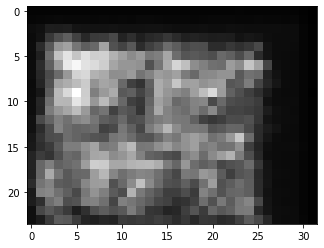}
		\caption{}
		\label{fig:ada1_768_pers}
	\end{subfigure}
	\hfill
	\begin{subfigure}{.19\textwidth}
		\centering
		\includegraphics[width=0.99\linewidth]{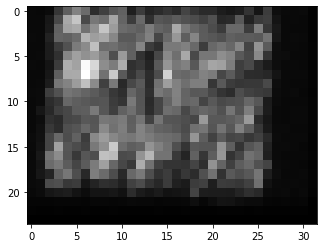}
		\caption{}
		\label{fig:ada2_768_pers}
	\end{subfigure}
	\hfill
	\begin{subfigure}{.19\textwidth}
		\centering
		\includegraphics[width=0.99\linewidth]{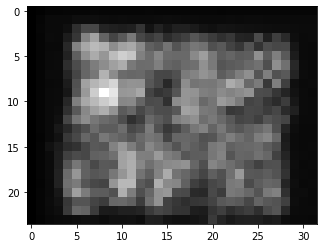}
		\caption{}
		\label{fig:ada3_768_pers}
	\end{subfigure}
	\hfill
	\begin{subfigure}{.19\textwidth}
		\centering
		\includegraphics[width=0.99\linewidth]{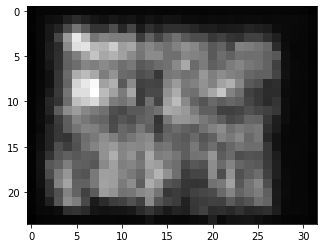}
		\caption{}
		\label{fig:ada_768_pers}
	\end{subfigure}
	\begin{subfigure}{.19\textwidth}
		\centering
		\includegraphics[width=0.99\linewidth]{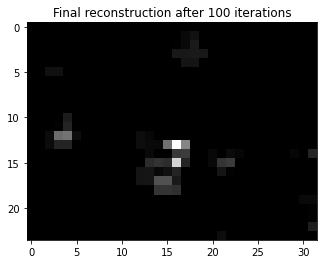}
		\caption{}
		\label{fig:ada0_768_pers_recon}
	\end{subfigure}
	\hfill
	\begin{subfigure}{.19\textwidth}
		\centering
		\includegraphics[width=0.99\linewidth]{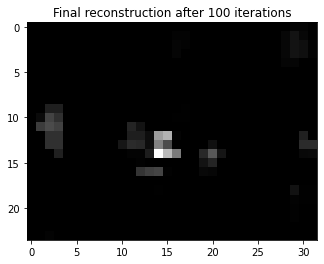}
		\caption{}
		\label{fig:ada1_768_pers_recon}
	\end{subfigure}
	\hfill
	\begin{subfigure}{.19\textwidth}
		\centering
		\includegraphics[width=0.99\linewidth]{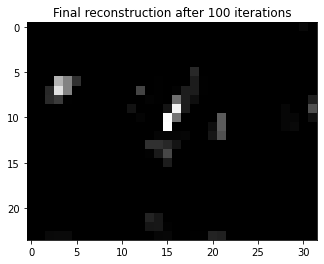}
		\caption{}
		\label{fig:ada2_768_pers_recon}
	\end{subfigure}
	\hfill
	\begin{subfigure}{.19\textwidth}
		\centering
		\includegraphics[width=0.99\linewidth]{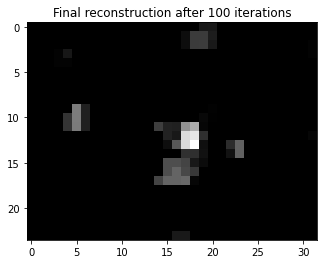}
		\caption{}
		\label{fig:ada3_768_pers_recon}
	\end{subfigure}
	\hfill
	\begin{subfigure}{.19\textwidth}
		\centering
		\includegraphics[width=0.99\linewidth]{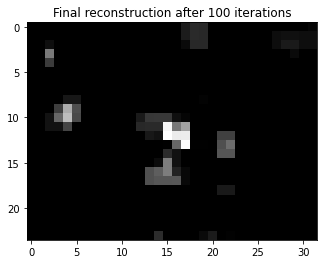}
		\caption{}
		\label{fig:ada4_768_pers_recon}
	\end{subfigure}
	\caption{Example reconstructions (100 iterations of ADMM to solve \Cref{eq:lensless_inverse}) for \textit{Fixed SLM (m)} in the presence of random perspective changes, for an embedding dimension of $ 24\times 32 $. (Top) raw measurements and (bottom) corresponding reconstruction.}
	\label{fig:ada_pers_768}
\end{figure}

\begin{figure}[h]
	\centering
	\begin{subfigure}{.19\textwidth}
		\centering
		\includegraphics[width=0.99\linewidth]{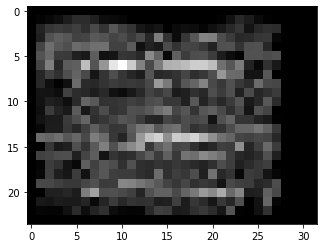}
		\caption{}
		\label{fig:learned0_768_pers}
	\end{subfigure}
	\hfill
	\begin{subfigure}{.19\textwidth}
		\centering
		\includegraphics[width=0.99\linewidth]{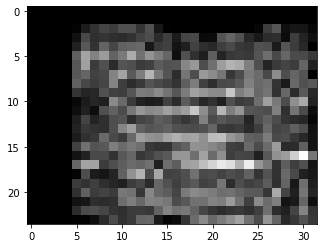}
		\caption{}
		\label{fig:learned1_768_pers}
	\end{subfigure}
	\hfill
	\begin{subfigure}{.19\textwidth}
		\centering
		\includegraphics[width=0.99\linewidth]{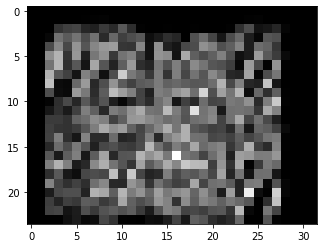}
		\caption{}
		\label{fig:learned2_768_pers}
	\end{subfigure}
	\hfill
	\begin{subfigure}{.19\textwidth}
		\centering
		\includegraphics[width=0.99\linewidth]{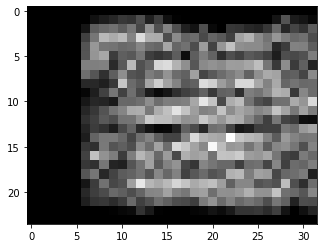}
		\caption{}
		\label{fig:learned3_768_pers}
	\end{subfigure}
	\hfill
	\begin{subfigure}{.19\textwidth}
		\centering
		\includegraphics[width=0.99\linewidth]{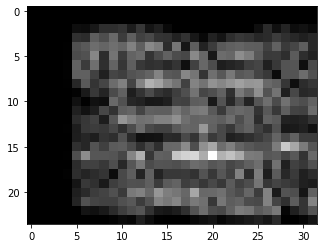}
		\caption{}
		\label{fig:learned4_768_pers}
	\end{subfigure}
	\begin{subfigure}{.19\textwidth}
		\centering
		\includegraphics[width=0.99\linewidth]{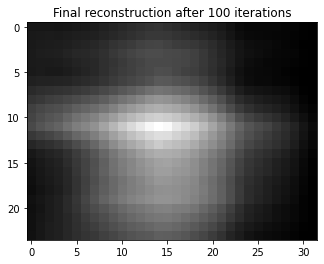}
		\caption{}
		\label{fig:learned0_768_pers_recon}
	\end{subfigure}
	\hfill
	\begin{subfigure}{.19\textwidth}
		\centering
		\includegraphics[width=0.99\linewidth]{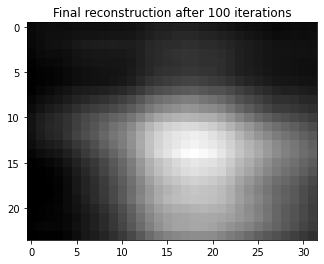}
		\caption{}
		\label{fig:learned1_768_pers_recon}
	\end{subfigure}
	\hfill
	\begin{subfigure}{.19\textwidth}
		\centering
		\includegraphics[width=0.99\linewidth]{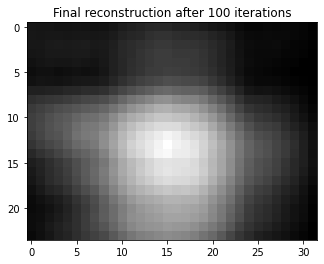}
		\caption{}
		\label{fig:learned2_768_pers_recon}
	\end{subfigure}
	\hfill
	\begin{subfigure}{.19\textwidth}
		\centering
		\includegraphics[width=0.99\linewidth]{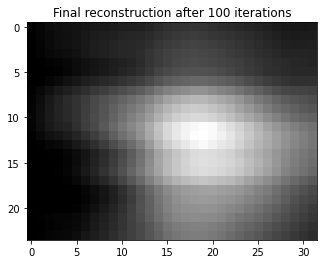}
		\caption{}
		\label{fig:learned3_768_pers_recon}
	\end{subfigure}
	\hfill
	\begin{subfigure}{.19\textwidth}
		\centering
		\includegraphics[width=0.99\linewidth]{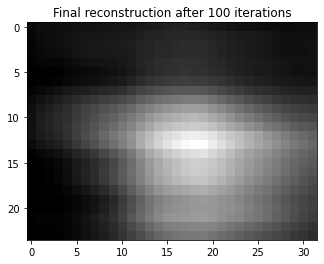}
		\caption{}
		\label{fig:learned4_768_pers_recon}
	\end{subfigure}
	\caption{Example reconstructions (100 iterations of ADMM to solve \Cref{eq:lensless_inverse}) for \textit{Learned SLM} in the presence of random perspective changes, for an embedding dimension of $ 24\times 32 $. (Top) raw measurements and (bottom) corresponding reconstruction.}
	\label{fig:learned_pers_768}
\end{figure}

\Cref{fig:ada_pers_768,fig:learned_pers_768} show the effect of perspective changes on raw embeddings and reconstructed outputs of \textit{Fixed SLM (m)} and \textit{Learned SLM}.

\end{document}